\documentclass[11pt]{article}
\usepackage{graphicx}
\usepackage[margin=1in]{geometry}
\usepackage{booktabs}
\usepackage{xcolor}
\usepackage{amsmath,amsfonts,amssymb,amsthm}
\usepackage[colorlinks=true,linkcolor=blue,citecolor=blue]{hyperref}
\usepackage{natbib}
\usepackage{soul}
\usepackage{multirow}
\usepackage{pifont}
\usepackage{authblk}
\usepackage{float}
\usepackage[ruled,vlined,boxed]{algorithm2e}
\usepackage{subcaption}

\newcommand{\rev}[1]{\textcolor{black}{#1}}

\graphicspath{{./imgs/}} 

\title{Rule-based and reinforcement-learning policies for resource allocation in a pediatric echo lab: a simulation study}
\author[1]{Bozhi Sun}
\author[2]{Seda Tierney}
\author[2,3]{Jeffrey A. Feinstein}
\author[2]{Frederick Damen}
\author[2,3]{Alison L. Marsden}
\author[1]{Daniele E. Schiavazzi\thanks{Corresponding author - \emph{dschiavazzi@nd.edu}}}
\affil[1]{Department of Applied and Computational Mathematics and Statistics, University of Notre Dame, Notre Dame, IN, USA.}
\affil[2]{Department of Pediatrics (Cardiology), Stanford University, CA 94305, US.}
\affil[3]{Department of Bioengineering, Stanford University, CA 94305, US.}
\date{}

\begin{document}

\maketitle

\begin{abstract}
\noindent Scheduling echocardiographic examinations in a hospital setting presents significant challenges due to non-deterministic factors such as patient no-shows, variable arrival times, and heterogeneous exam durations, as well as asymmetric resource constraints between fetal and non-fetal patient streams. \rev{Fetal cases involve specialized ultrasound examinations performed during pregnancy, whereas non-fetal cases correspond to general diagnostic echocardiography services.}
Using one week (184 patients) of operational data from the Echo Laboratory at Stanford University's Lucile Packard Children's Hospital, we estimate patient no-show probabilities and derive empirical distributions of arrival times and exam durations, and use them to build a discrete-event stochastic simulation model using \texttt{SimPy}, exposed through an open-source Gymnasium interface (\url{https://github.com/desResLab/echo_rl}).
As a baseline for policy optimization, we systematically compare two scheduling paradigms: on-the-fly allocation, in which patients are served according to their arrival time without prior reservations, and reservation-based allocation, in which varying proportions of resources are preassigned.
We consider a hospital with seven ultrasound rooms (one equipped for both fetal and non-fetal exams, six for non-fetal exams only) and six sonographers (four qualified for both exam types, two for non-fetal exams only). In this setting, on-the-fly allocation markedly reduces the average patient waiting time compared with every reservation-based configuration tested.
Building on this foundation, we train a Dueling Double Deep Q-Network (DQN) with prioritized experience replay to approximate an optimal dynamic on-the-fly allocation policy. The learned RL policy is comparable to, but does not outperform, the best on-the-fly rule-based policy. 
\end{abstract}


\section{Introduction}
Echocardiography, or \emph{echo}, is a noninvasive imaging technique widely used to assess cardiac structure and function. It plays a vital role in diagnosing and managing a range of cardiovascular conditions.
However, in most healthcare settings, the number of available echo rooms and trained sonographers is limited, creating a bottleneck in service delivery. Delays are further exacerbated by patient-related uncertainties, such as late arrivals or no-shows, which can disrupt tightly scheduled workflows.
Together, these two factors -- resource scarcity and patient unpredictability -- contribute to prolonged patient waiting times.
\rev{Since hospital visits often involve multiple tests scheduled in close succession rather than a single echocardiography exam,} extended delays in echo testing can trigger a ripple effect that disrupts all subsequent appointments. These disruptions are not merely operational; they can postpone critical diagnoses and delay timely treatment, particularly for patients requiring urgent cardiac evaluation.

This type of problem, also known as the hospital staff and resource allocation problem, is addressed by determining an efficient resource allocation policy.
This problem is a specific instance of the broader outpatient-appointment-scheduling literature under no-shows and unpunctuality with heterogeneous, partially flexible servers~\citep{ahmadijavid2017outpatient, cayirli2003outpatient}, and is most closely related to diagnostic-resource scheduling with dynamic priority rules~\citep{green2006managing, patrick2008dynamic}. We estimate the no-show rate from our data (Section~\ref{sec:nalysis of patient's no-show rate}) and treat it as i.i.d.\ rather than modeling lead time or prior no-show history as predictors; we also perform a sensitivity analysis (Section~\ref{sec:sensitivity}) where we perturb such a rate from 10\% to 40\% to assess robustness to this assumption. 
Whether flexible servers should be pooled across patient classes or dedicated to individual classes is a well known trade-off in queueing theory, and documented counterexamples show that pooling is not always beneficial~\citep{joustra2010pool}. For a broader review of resource allocation under uncertainty, see~\cite{cardoen2010operating}.
Resource allocation has traditionally relied on heuristic rules and staff judgment, but simulation and machine learning methods are now increasingly used to improve operational efficiency.
Hejazi~\cite{HEJAZI2021107502} used simulation to show that a state-dependent queuing policy reduces waiting time while increasing throughput, and Mizan and Taghipour~\cite{MIZAN2022102430} combine machine learning with optimization to reduce radiological workload and patient waiting time.
Reinforcement learning has also shown promise in this domain~\citep{wu2025reinforcement}: Lazebnik~\cite{LAZEBNIK2023106783} develops a multi-agent simulation with a DRL-based global decision-maker, and, closer in formulation to the present work, Lee and Lee~\cite{lee2020improving} train a DQN over an MDP that minimizes weighted patient waiting time with acuity-based priority classes in an emergency department.
This latter represents the nearest published analogue to Section~\ref{sec:mdp} in this study.

In this study, we address the challenges of optimizing echo resource allocation in healthcare settings by incorporating non-deterministic factors into a stochastic modeling framework. 
We compare rule-based on-the-fly and reservation-based policies, and evaluate whether a reinforcement-learning policy can improve on the best rule-based baseline.
To achieve this, we propose a structured framework for resource allocation.
Our key contributions include:

\begin{enumerate}
\item \textbf{Systematic analysis of predefined (or rule-based) policies:} We have implemented and analyzed four on-the-fly policies and two parametric hybrid policies (where patients are allocated partially on-the-fly and partially based on reservations), demonstrating that on-the-fly allocation outperforms the reservation-based configurations we tested in accommodating patients.
These findings are validated through simulations conducted under specific hospital settings, over 365 independent replications of a representative day (Section~\ref{sec:predefined_policies}).
\item \textbf{A critical evaluation of reinforcement learning for the hospital allocation problem:} We train an RL agent (Dueling Double DQN with prioritized experience replay) and benchmark it against multiple rule-based policies.
Under our reward design, the RL policy matches but does not improve over the strongest rule-based policy, underperforming on end-of-day overruns; we discuss why, and what changes (reward/state redesign, multi-seed training, additional baselines) would be needed to determine whether this is a property of the problem or of the specific algorithm used (Section~\ref{sec:dif} and Discussion).
\end{enumerate}

Our contributions lay the groundwork for advancing resource allocation methodologies in healthcare, combining simulation-based analysis with practical implications, to enhance patient accommodation and operational efficiency.

This paper is organized as follows: we first analyze available data in Section~\ref{sec:data}, and provide a detailed description of the proposed hospital stochastic model in Section~\ref{sec:model}.
Patients' \emph{weighted} waiting times (i.e., doubled waiting time for patients with \emph{on-time} arrivals) are then analyzed for six families of policy (four fixed policies, plus two parametric families spanning ten configurations each) in Section~\ref{sec:predefined_policies}.
The problem is then formalized as a Markov Decision Process (MDP) in Section~\ref{sec:mdp}, where we also provide details for the design of states, actions, rewards and the implementation of a \rev{Dueling Double Q-learning} algorithm to learn the optimal policy.
Results under this policy are discussed in Section~\ref{sec:lp}.
In Section~\ref{sec:sensitivity}, we perform sensitivity analysis for both rule-based and RL policies, considering the main environmental parameters.
In Section~\ref{sec:dif}, we discuss differences between rule-based and RL policies, under an increase and a reduction in the available resources (number of sonographers, echo rooms). 
Finally, Sections~\ref{sec:discussion} and~\ref{sec:conclusions} present a discussion of the results and our concluding remarks, respectively.

The source code for this project is available at \url{https://github.com/desResLab/echo_rl}, archived at the version used for this manuscript (DOI to be assigned upon acceptance). 
The custom environment we designed is compatible with the \texttt{gymnasium} framework.

\section{Dataset}\label{sec:data}

The dataset used in this study, available at \url{https://github.com/desResLab/echo_rl/blob/main/Whiteboard%20Data-Training.xlsx}, was collected by monitoring echocardiography room activity continuously from Monday to Friday during a full workweek.

The observed quantities were summarized in a spreadsheet which included: five patient-related features (\emph{patient ID}, \emph{subspecialty}, \emph{chemotherapy}, \emph{fetal}, \emph{research}), three features quantifying the expected/actual arrival time for each patient (\emph{expected check-in/arrival time}, \emph{actual check-in/arrival time}, \emph{pt$>$10 min late}, defined as a delay greater than 10 minutes), and the corresponding late-arrival flag.
The next set of fields in the dataset focuses on the measurement of vitals using four records (\emph{vitals in}, \emph{vitals out}, \emph{vitals duration}, \emph{met vitals cycle time}).
Eight records were then related to the echocardiographic exam (\emph{echo scheduled time}, \emph{echo time in}, \emph{echo time out}, \emph{echo duration}, \emph{met echo cycle time}, \emph{echo start late $>$10 min}, \emph{sonographer echo room number},
\emph{echo teaching case}).
The next three records kept track of the schedule and duration of the \rev{electrocardiogram (or EKG)} examination (\emph{EKG in}, \emph{EKG out}, \emph{EKG duration}), while the last three records were related to the clinical visits (\emph{clinic scheduled}, \emph{patient in the clinic room}, \emph{late to clinic appointment $>$10 min}).
A summary of the fields in the dataset and their types is reported in Table~\ref{tab:dataset}. 
The fields used in this study are: fetal (yes/no), echo scheduled time, echo time in, and echo time out; the vitals, EKG, and clinic-visit fields listed above are recorded but not used in our model (see Discussion, Section~\ref{sec:discussion}).
%
The number of patients in the dataset is equal to $\{38,34,41,40,31\}$ for Monday, Tuesday, Wednesday, Thursday and Friday, respectively, for a total of 184 patient records across the week. 
As is typical of manually logged, whiteboard-style scheduling data, many fields in the raw dataset are missing or incomplete: front-desk and clinical staff did not consistently record every timestamp for every patient, so fields such as vitals, EKG, and echo in/out times are frequently left blank when a step was skipped, delayed in charting, or not applicable to a given visit. 
Because of this, the samples used in the analyses of Sections~\ref{sec:Analysis of patient's delay time}--\ref{Analysis of patient's echo duration} and in Figures~\ref{fig:arrival_time}/\ref{fig:echo_duration} are smaller than the full 184-record total. 
Each analysis retains only the subset of patients with complete, valid entries for the specific fields it requires (e.g., both an expected and an actual arrival timestamp for the delay analysis, or a recorded echo duration for the duration analysis), and drops records with missing or clearly invalid entries for those fields. 
We consider this field-by-field completeness filtering a reasonable and standard way of extracting a usable analysis set from an imperfect, manually collected clinical dataset.

Moreover, the dataset also quantified available resources in terms of echo rooms and sonographers. 
The facility has six regular echo rooms and one fetal echo room. A total of six sonographers are dedicated to the outpatient heart center (only four of these sonographers can do fetal scans), two are deployed to inpatient activities, and one to two are deployed to three satellite facilities. 

Each sonographer is supposed to take a 30-minute lunch break and two 15-minute breaks per day.
Our model focuses on outpatient activities, and for the sake of allocation smoothness, we omit the lunch break period to avoid sudden interruptions in the allocation process. In practice, lunch break timing tends to be flexible, and a sonographer who is in the middle of an active case does not necessarily step away at a fixed time; this makes it difficult to faithfully model staggered lunch breaks in the simulation. 
We therefore omit the lunch break from the model, and do not expect this simplification to materially affect our main conclusions.

\begin{table}[!ht]
\centering
\caption{List of features in the dataset.}\label{tab:dataset}
\begin{tabular}{c c c c}
\toprule
{\bf QTY} & {\bf TYPE} & {\bf QTY} & {\bf TYPE}\\
\midrule
patient ID & \texttt{int} & echo time in & \texttt{time}\\
subspecialty & \texttt{string} & echo time out & \texttt{time}\\
chemotherapy & \texttt{yes/no} & echo duration & \texttt{time}\\
fetal & \texttt{yes/no} & met echo cycle time & \texttt{yes/no}\\
research & \texttt{yes/no} & echo start late $>$10 min & \texttt{yes/no}\\
expected check-in/arrival time & \texttt{time} & sonographer echo room number & \texttt{int}\\
actual check in/arrival time & \texttt{time} & echo teaching case & \texttt{yes/no}\\
pt$>$10 min late & \texttt{yes/no} & ekg in & \texttt{time}\\
vitals in & \texttt{time} & ekg out & \texttt{time}\\
vitals out & \texttt{time} & ekg duration & \texttt{time}\\
vitals duration & \texttt{time} & clinic scheduled & \texttt{time}\\
met vitals cycle time & \texttt{yes/no} & patient in clinic room & \texttt{time}\\
echo scheduled time & \texttt{time} & late to clinic appt. $>$10 min & \texttt{yes/no}\\
\bottomrule
\end{tabular}
\end{table}

\section{Model design}\label{sec:model}

We note that our model, like any simulation, is necessarily a simplification of reality: given the limitations of the underlying dataset (a single week of manually logged, partially incomplete records) and the practical need for a tractable simulation, it cannot reproduce every operational detail of the real system. 
Instead, we designed the model to capture what we judged to be the dominant drivers of patient waiting time (resource scarcity, no-shows, and variability in arrival times and exam durations), and to approximate the real operating environment as closely as possible within that scope. 
The resulting model design was reviewed and endorsed by our clinical collaborators at the Echo Laboratory.

\subsection{Hours of operation and visit schedule}\label{sec:hours_visit_schedule}

To evaluate the effect of different resource allocation on total patient waiting time, we designed a simulated hospital environment, trying to make it as realistic as possible.
All routines are implemented in \texttt{SimPy}~\cite{simpyteam} and, during the simulations, time is measured in minutes.

In our simulation, the hospital operates from 8:00 AM to 5:00 PM. Patients arriving before 8:00~AM wait until 8:00~AM to enter; patients arriving after 5:00~PM are unable to physically access the facility. However, patients that are already in the facility will still receive service.
We track the dynamics of patients and echo resources until 7:00 PM. When patients remain in the echo lab past 7:00 PM, we impose a penalty. The specific definition and implementation of this penalty are described later in the paper.
While all sonographers and echo rooms can conduct non-fetal tests, not all are equipped to handle fetal exams.

\begin{figure}[ht!]
\centering
\begin{subfigure}[b]{0.48\textwidth}
    \centering
    \includegraphics[width=\textwidth]{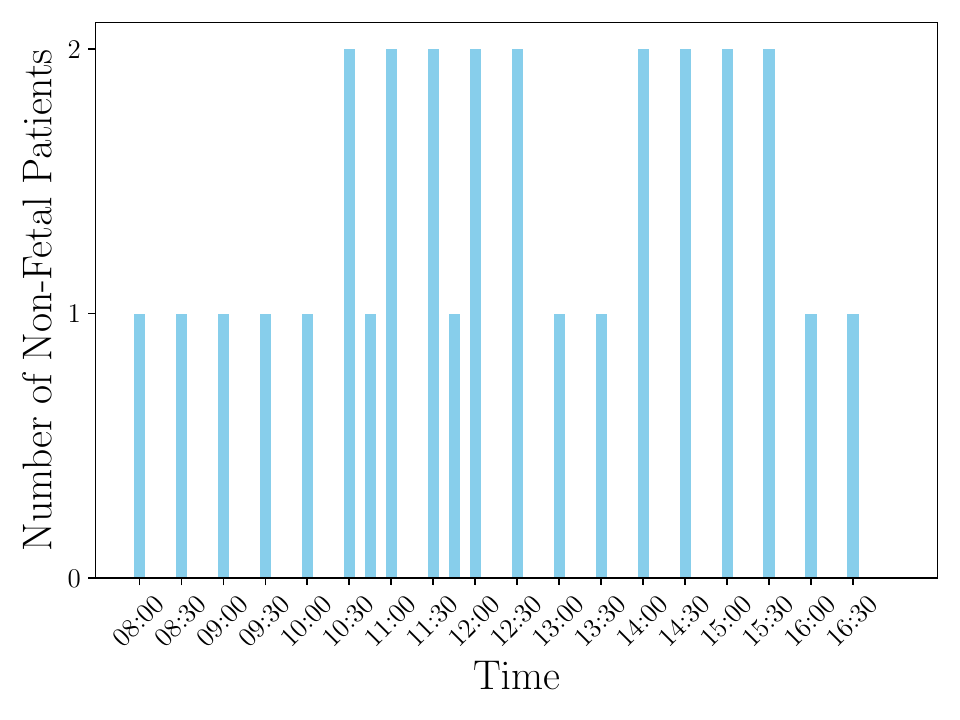}
    \caption{Scheduled non-fetal visits during a typical day of operation.}
    \label{fig:nonfetal}
\end{subfigure}
\hfill
\begin{subfigure}[b]{0.48\textwidth}
    \centering
    \includegraphics[width=\textwidth]{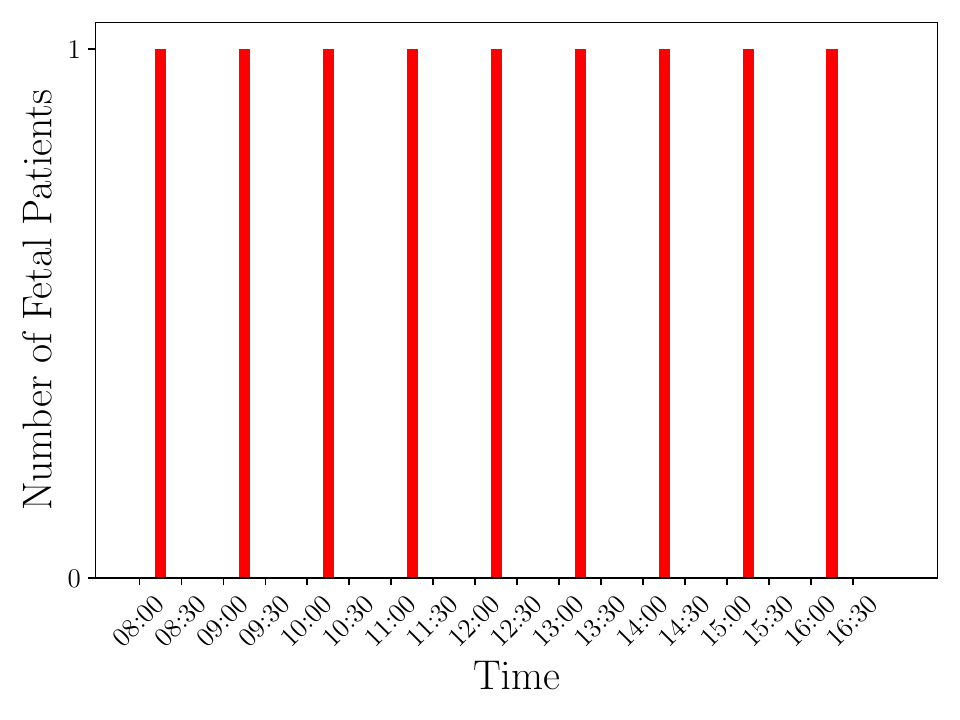}
    \caption{Scheduled fetal visits during a typical day of operation.}
    \label{fig:fetal}
\end{subfigure}
\caption{Echo visit schedules for both fetal and non-fetal patients.}
\label{fig:patients_schedules}
\end{figure}

Visits in the simulation are scheduled at predetermined time points throughout the day.
For fetal patients, visits begin at 8:15 AM, with subsequent appointments scheduled every hour until 4:15 PM. A schematic diagram is shown in Figure~\ref{fig:fetal}.
For non-fetal patients, visits start at 8:00 AM, with appointments scheduled every half hour until 4:30 PM.
Some time slots are busier, with two non-fetal patients scheduled simultaneously. Additionally, some appointments are scheduled at fifteen or forty-five minutes past the hour, see Figure~\ref{fig:nonfetal}. There are 38 patients in total in our simulated environment. It is important to note that this stochastic model is designed to mimic routine operations in an echocardiography facility and is not intended for emergency care scenarios. 
Patients whose simulated arrival time falls after 5:00~PM are handled by the same mechanism as no-shows: in both cases the patient's status never transitions out of ``incoming,'' so no waiting penalty is incurred and the current implementation does not distinguish or separately label the two cases.

\subsection{Arrival times}

The simulation assumes a 10\% chance that a patient will not show up for their appointment. \rev{We define an \emph{absence} in our dataset as a case in which a patient has a scheduled echo appointment but no corresponding recorded check-in time. For additional details on how these absences were identified, please refer to Section~\ref{sec:nalysis of patient's no-show rate}}.
We define \emph{early arrival} as a patient arriving more than 10 minutes before the scheduled time, and \emph{on-time arrival} as a patient arriving between 10 minutes before and 10 minutes after the scheduled time.
A \emph{late arrival} occurs when a patient arrives more than 10 minutes after the scheduled time.

\begin{figure}[ht!]
\centering
\begin{subfigure}[b]{0.48\textwidth}
    \centering
    \includegraphics[width=\textwidth]{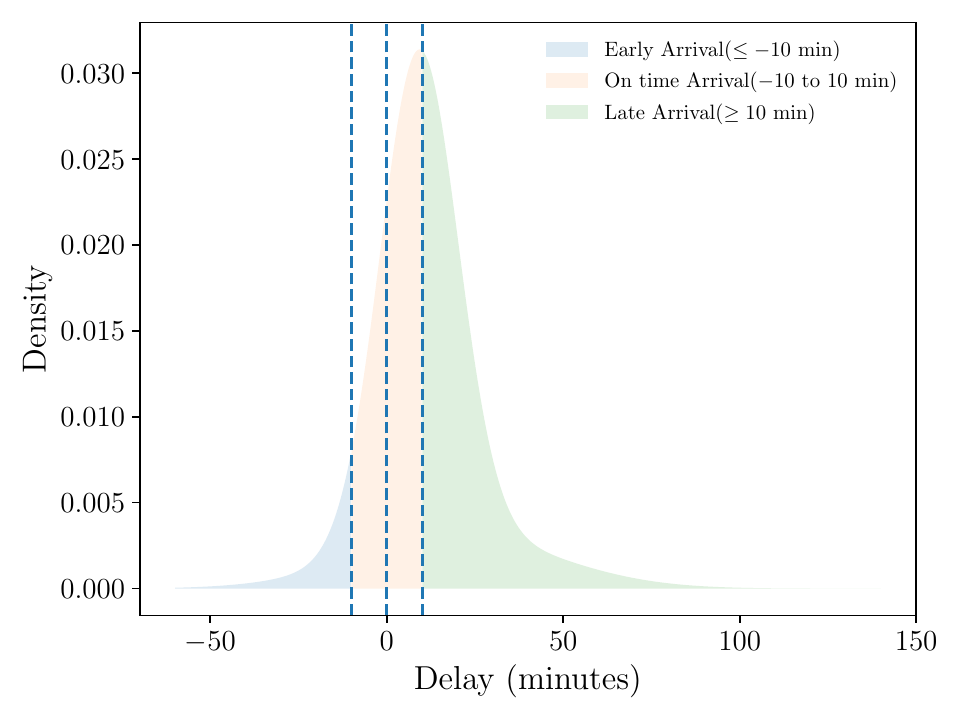}
    \caption{Distribution of difference between arrival and scheduled time.}
    \label{fig:time_diff_distribution}
\end{subfigure}
\hfill
\begin{subfigure}[b]{0.48\textwidth}
    \centering
    \includegraphics[width=\textwidth]{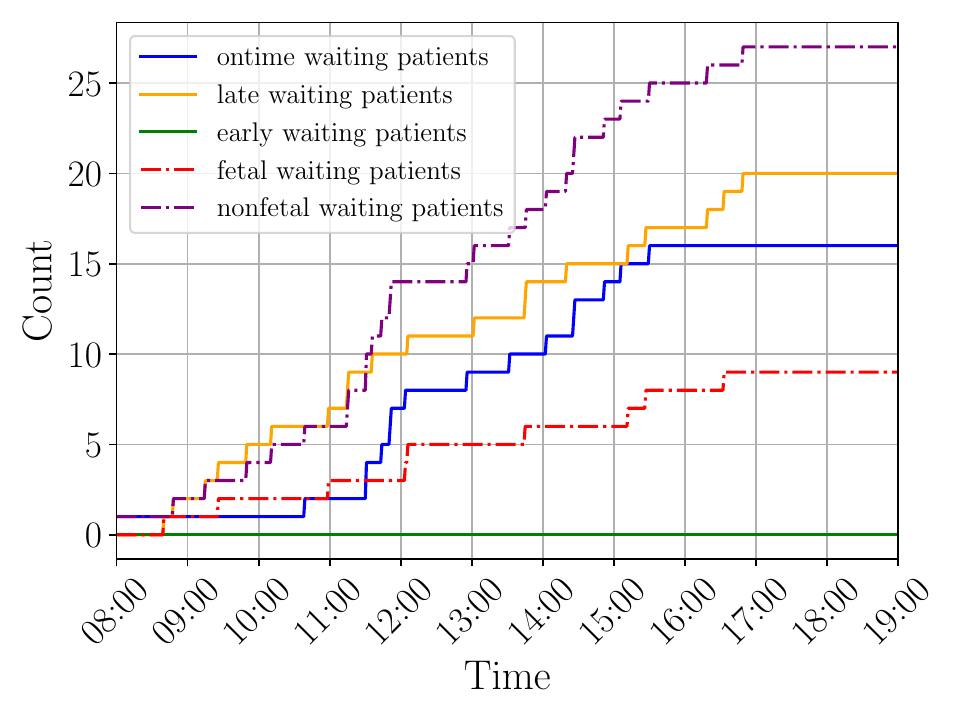}
    \caption{An example of the dynamics of patients arriving at the facility.}
    \label{fig:arrival_dyn}
\end{subfigure}
\caption{Fitted distribution of arrival delays (left) and example of arrival dynamics (right).}
\label{fig:arrival_figure}
\end{figure}

For situations where patients arrive before 8:00 AM, their arrival time is recorded as 8:00 AM, the time they are allowed to enter the facility.
As previously discussed, patients arriving after 5:00 PM will not have access to the facility.
\rev{We assume that the difference between a patient’s actual arrival time and their scheduled appointment time follows a mixture of two Gaussian distributions. The first component represents the majority of patients and has a mean of $\mu_1 = 9.12$ minutes with a standard deviation of $\sigma_1 = 11.02$ minutes, contributing $79\%$ of the total probability mass. The second component, representing a smaller subgroup with more variable arrival patterns, has a mean of $\mu_2 = 20.14$ minutes and a standard deviation of $\sigma_2 = 27.98$ minutes, contributing the remaining $21\%$.
For additional details on the rationale behind the selection of a mixture model formulation, please see Section~\ref{sec:Analysis of patient's delay time}.
A mixture model captures both the predominant group of patients who arrive close to their scheduled times and the smaller fraction who arrive substantially later.
This step is necessary to establish the allocation priority order (see Figure~\ref{fig:time_diff_distribution} for the corresponding early, on-time, and late thresholds).
Since the Gaussian mixture is continuous, we round the values to minutes using the \emph{ceiling function} ($\lceil \cdot \rceil$)}, which introduces a systematic sub-minute lateness bias relative to rounding to the nearest minute. An example of the patient arrival dynamics is shown in Figure \ref{fig:arrival_dyn}.

\subsection{Echo test resources and duration}

\begin{figure}[ht!]
\centering
\begin{subfigure}[b]{0.48\textwidth}
    \centering
    \includegraphics[width=\textwidth]{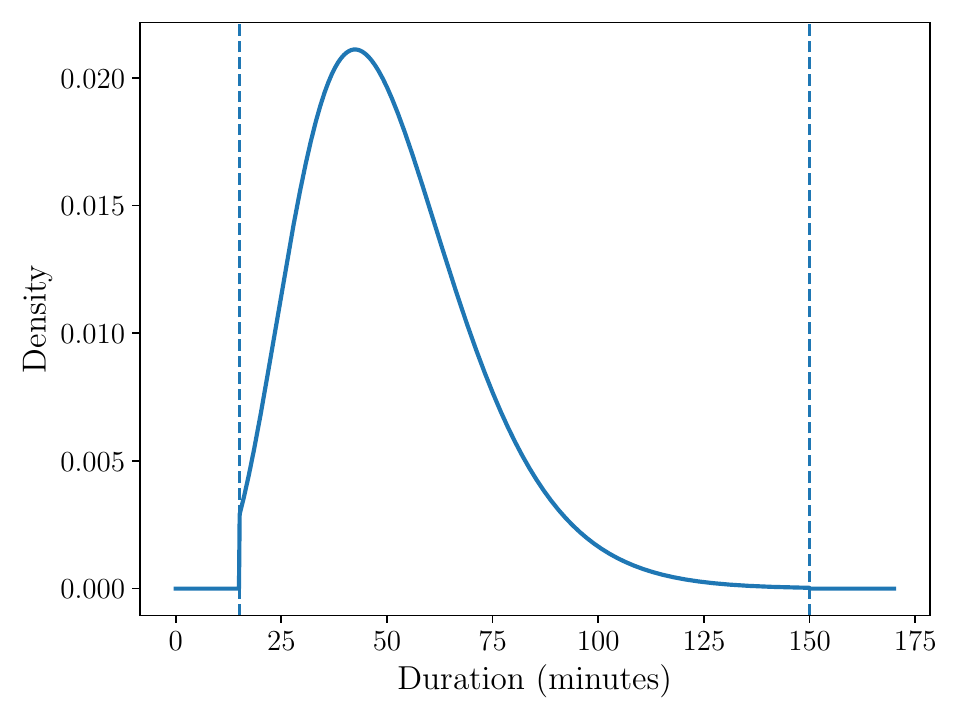}
    \caption{Fitted distribution for the echo test duration.}
    \label{fig:distr_echo_test}
\end{subfigure}
\hfill
\begin{subfigure}[b]{0.48\textwidth}
    \centering
    \includegraphics[width=\textwidth]{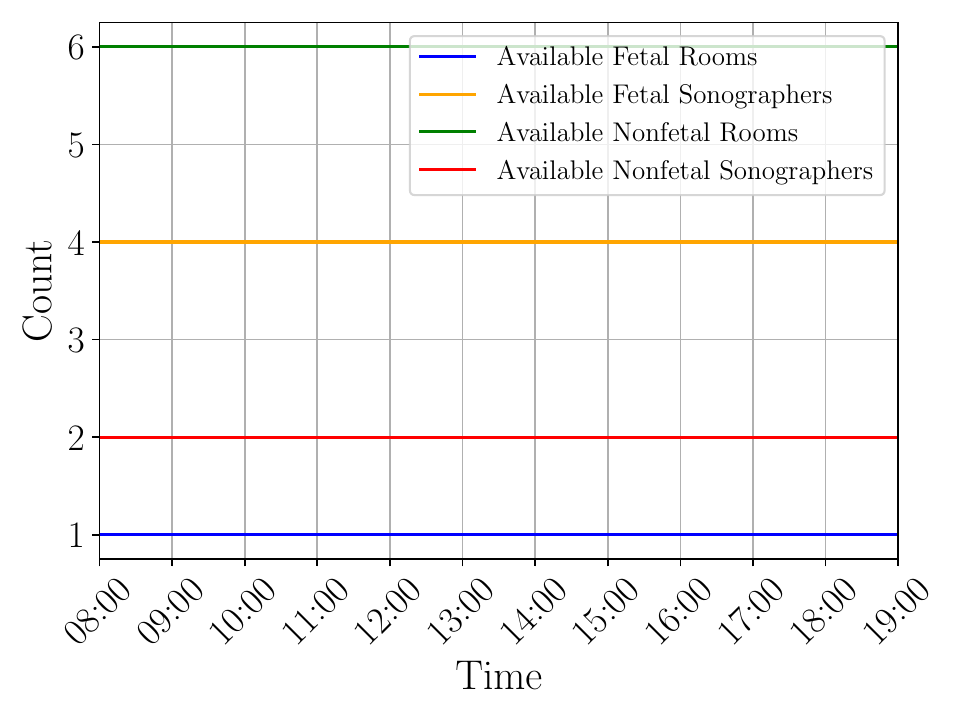}
    \caption{Plot of all the echo resources in the hospital.}
    \label{fig:avail_resources}
\end{subfigure}
\caption{Fitted distribution for the echo test duration and the number of echo resources.}
\label{c}
\end{figure}

The duration of an echo test is assumed to follow a truncated Gamma distribution, with parameters $k= 6.1151$ and $\theta = 8.2967$. The sampled duration is then truncated to the interval $[15, 150]$~minutes, which bounds the values generated by the simulator; no observation in our data falls outside this range (see Table~\ref{tab:echo_desc_stats}).
Additional details on this assumption can be found in Section~\ref{Analysis of patient's echo duration}.
As a result, the assumed distribution for the echo test duration is shown in Figure~\ref{fig:distr_echo_test}, whereas the resources considered in the simulation in terms of fetal/non-fetal rooms and sonographers are illustrated in Figure~\ref{fig:avail_resources}.

\subsection{Sonographer availability and break policy}

In our simulations, we ensure that at least two sonographers are scheduled each day -- one capable of conducting fetal echo exams and one restricted to non-fetal exams. This assumption is made to avoid extreme edge cases where accommodating patients becomes trivially infeasible.

For the remaining sonographers, there is a 10\% chance that each sonographer will be on leave and absent for the entire day.
Each sonographer is allotted two breaks after completing a test (breaks do not interrupt an in-progress exam), each with a 15-minute duration.
The break-taking behavior after each completed test depends on the sonographer's remaining break allotment for the day, as summarized in Table~\ref{tab:break_policy}.
\begin{table}[!ht] 
\centering 
\caption{Sonographer break-taking probabilities after completing a test, conditional on the number of breaks remaining that day.}\label{tab:break_policy} 
\begin{tabular}{c l c} 
\toprule 
Breaks remaining & Outcome & Probability\\ 
\midrule 2 & Take both breaks together (30 min) & 20\%\\
2 & Take one break (15 min) & 30\%\\ 
2 & No break taken & 50\%\\ 
1 & Take the remaining break (15 min) & 30\%\\ 
1 & No break taken & 70\%\\ 
0 & No break taken (none remaining) & --- \\ 
\bottomrule 
\end{tabular}
\end{table}
These probabilities could not be inferred from the dataset and are therefore assumed.
Furthermore, additional policies may be introduced \rev{in future work} to ensure compliance with state-specific regulations concerning break periods.

\section{Predefined policies for echo resource allocation}\label{sec:predefined_policies}

To compare the effectiveness of various rule-based policies under different scenarios, we observe their average effect over 365 independent replications of a representative day.
Additionally, we quantify the average patient waiting time and the sonographer \emph{quota}, measured as the average number of patients seen by each sonographer on a given day. 
At each decision step, we classify the waiting patients into three groups: the first group consists of \emph{waiting patients who arrive on time} and are assigned priority to access the echo resources. 
These patients are denoted as $P_{t}$. The second and third group $P_{l}$, $P_{e}$ consist of \emph{waiting patients who arrived late and early}, respectively.
As time advances, patients who arrive early and are still waiting will be reclassified as on-time arrivals when the difference between their scheduled time and the current time is less than 10 minutes, thereby gaining priority over echo resources.
Echo resources include both rooms and sonographers, both of which are required for patient examinations: there is one echo room capable of performing both fetal and non-fetal echo tests ($R_{tot,b}=1$), alongside six echo rooms dedicated solely to non-fetal echo testing ($R_{tot,n}=6$).
For sonographers, we have four who can perform both fetal and non-fetal testing ($S_{tot,b}=4$), and two who are only qualified to perform non-fetal echo testing ($S_{tot,n}=2$).

\subsection{Policy 1 - Accommodating on-time and late arrivals}

\begin{figure}[ht!]
\centering
\includegraphics[width=0.325\textwidth]{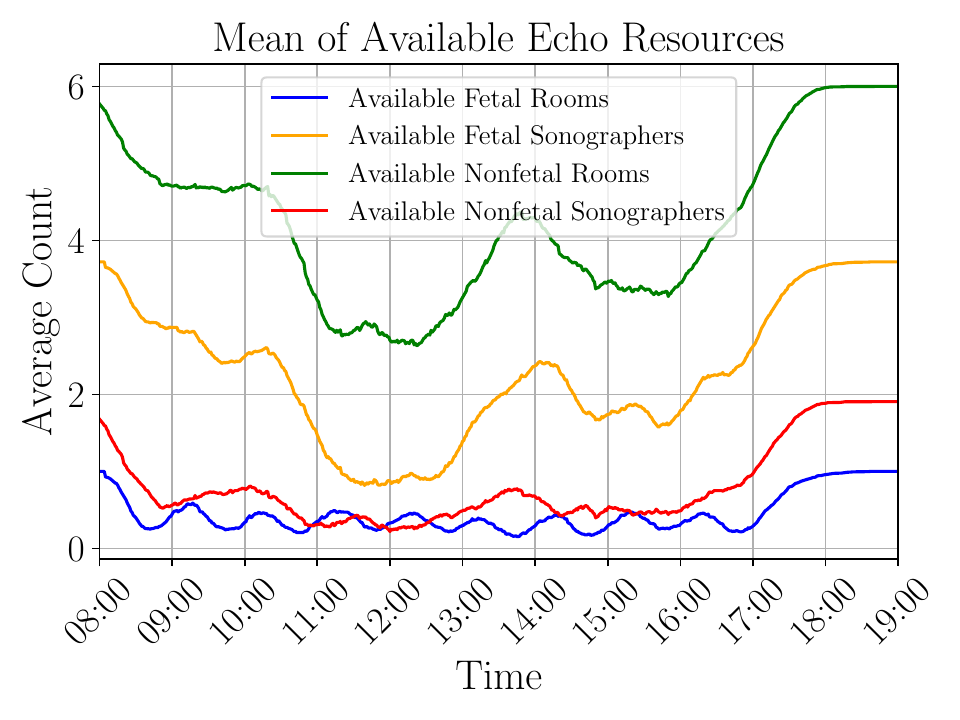}
\hfill
\includegraphics[width=0.325\textwidth]{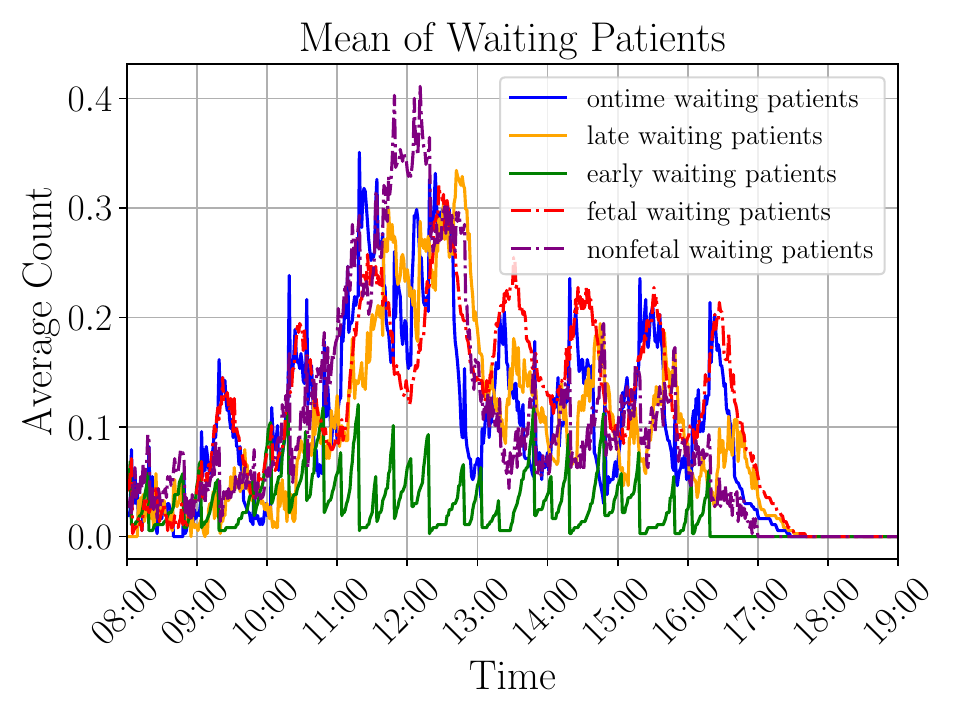}
\hfill
\includegraphics[width=0.325\textwidth]{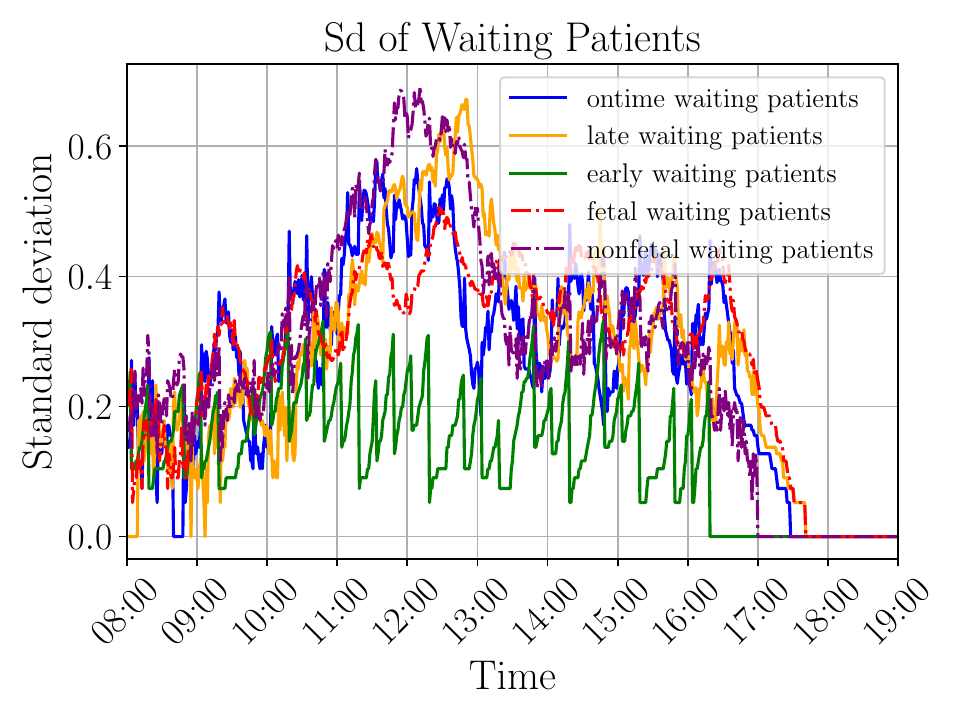}
\caption{Average results in terms of echo resources and waiting patients over 365 independent replications of a representative day for the stochastic hospital process enforcing policy 1.}\label{fig:res_Policy1}
\end{figure}

In policy 1, we prioritize accommodating patients who arrived on time before those who arrived late. At each time step, we first check if any patients in $P_{e}$ can be moved to $P_{t}$ if their scheduled visit is less than 10 minutes away.
We then allocate echo resources to patients in $P_{t}$, with non-fetal patients using non-fetal echo resources with priority (a tie-breaking rule we apply throughout this section and refer back to below). To do this, we gather all patients in $P_{t}$ and calculate their waiting times: for patients who arrived after their scheduled time, the waiting time is the difference between the current time and their arrival time; for patients who arrived before their scheduled time, the waiting time is the difference between the current time and their scheduled time (negative waiting time if the current time precedes the scheduled visit time).
We then sort these patients by waiting time in descending order and allocate available echo resources (both rooms and sonographers) to the patient with the longest waiting time. If no resources are available for the patient with the longest waiting time, we skip to the next patient and see if resources are available.
This process continues until all patients in $P_{t}$ are accommodated or no on-time-arrival resources remain. Following this, we allocate echo resources to patients in $P_{l}$ using the same tie-breaking rule.
We sort these late-arriving patients by their waiting times in descending order and allocate available resources accordingly, following the same approach used for on-time patients.
Average results across 365 independent replications following policy 1 are shown in Figure~\ref{fig:res_Policy1}.

\subsection{Policy 2 - Accommodating on time, late and early arrivals}

\begin{figure}[ht!]
\centering
\includegraphics[width=0.32\textwidth]{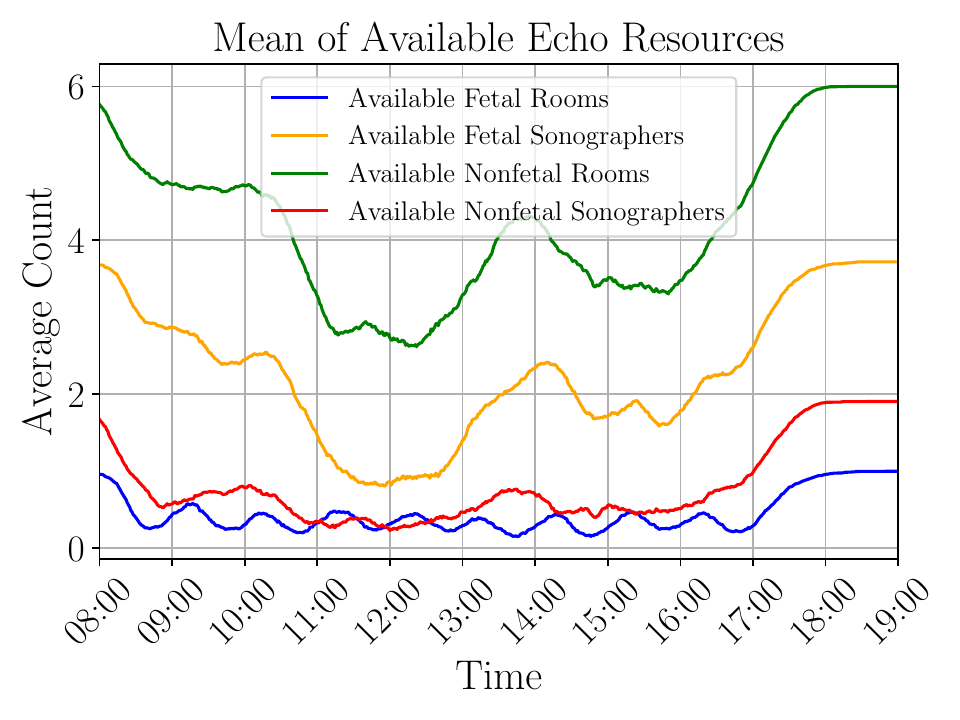}
\hfill
\includegraphics[width=0.32\textwidth]{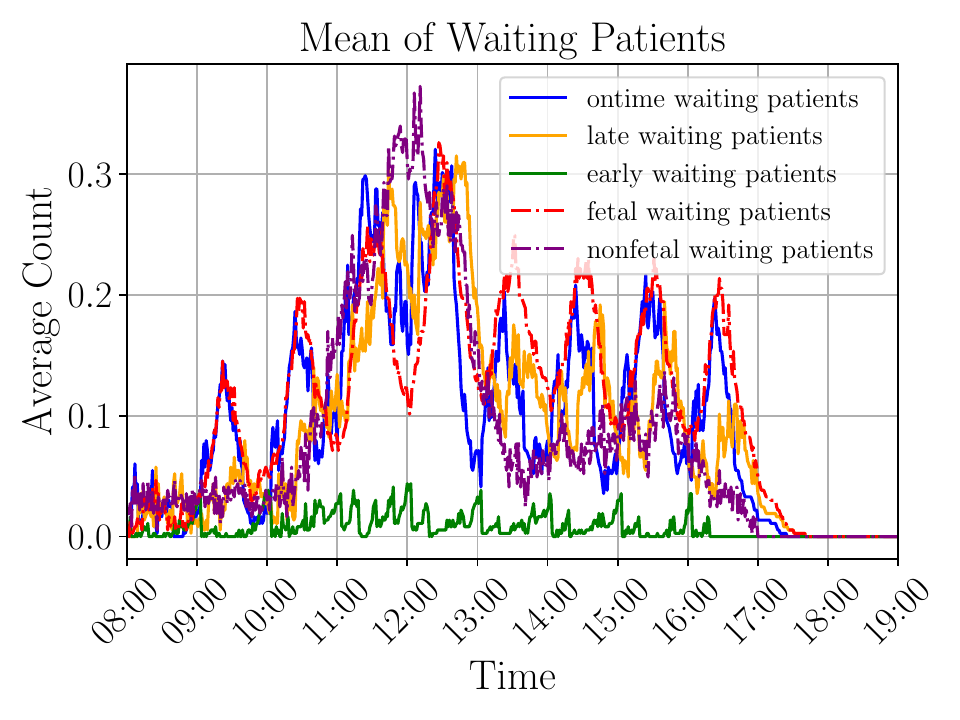}
\hfill
\includegraphics[width=0.32\textwidth]{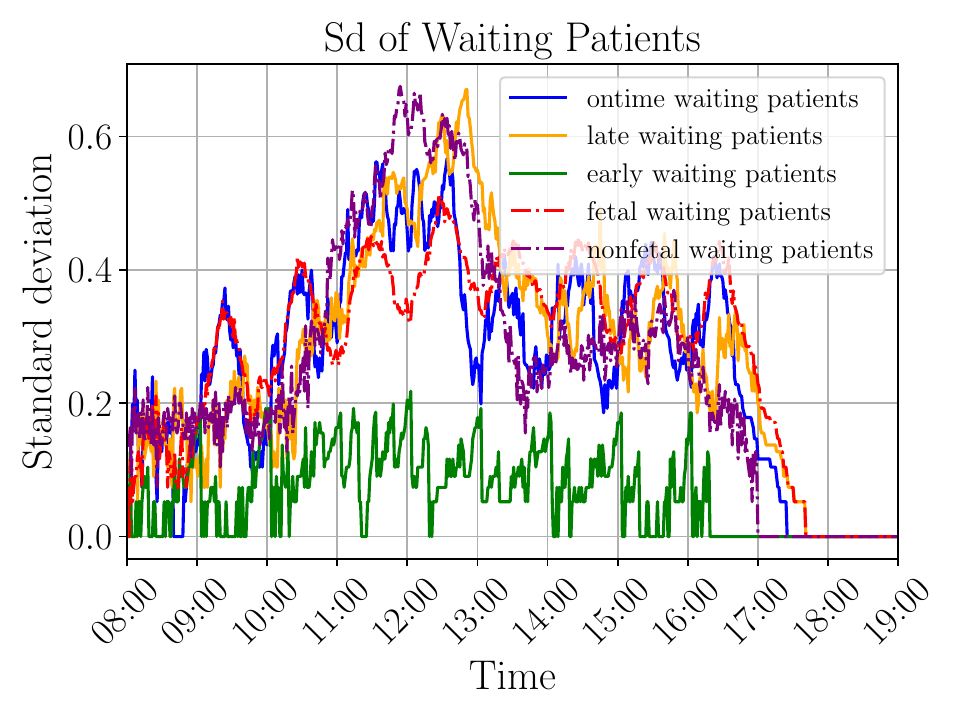}
\caption{Average results in terms of echo resources and waiting patients over 365 independent replications of a representative day for the stochastic hospital process enforcing predefined policy 2.}
\label{fig:res_Policy2}
\end{figure}

Policy 2 essentially follows the same procedure as policy 1, with an additional step at the end: accommodating patients who arrived early. At each time step, we first implement the allocation logic of policy 1, giving priority to patients who arrived on time or late. After that, any remaining available resources are used to serve early-arriving patients.
This process includes checking for patients who arrive early, sorting them in descending order of their waiting times (calculated as the difference between the current time and their scheduled time), and then allocating any available echo resources to these patients, applying the same non-fetal-priority tie-breaking rule defined above.
Average results across 365 independent replications following policy 2 are shown in Figure~\ref{fig:res_Policy2}. As expected, the average number of waiting patients arriving early is smaller than for policy 1.

\subsection{Policy 3 - Dedicating one fetal room and one fetal sonographer for fetal patients only}

\begin{figure}[ht!]
\centering
\includegraphics[width=0.32\textwidth]{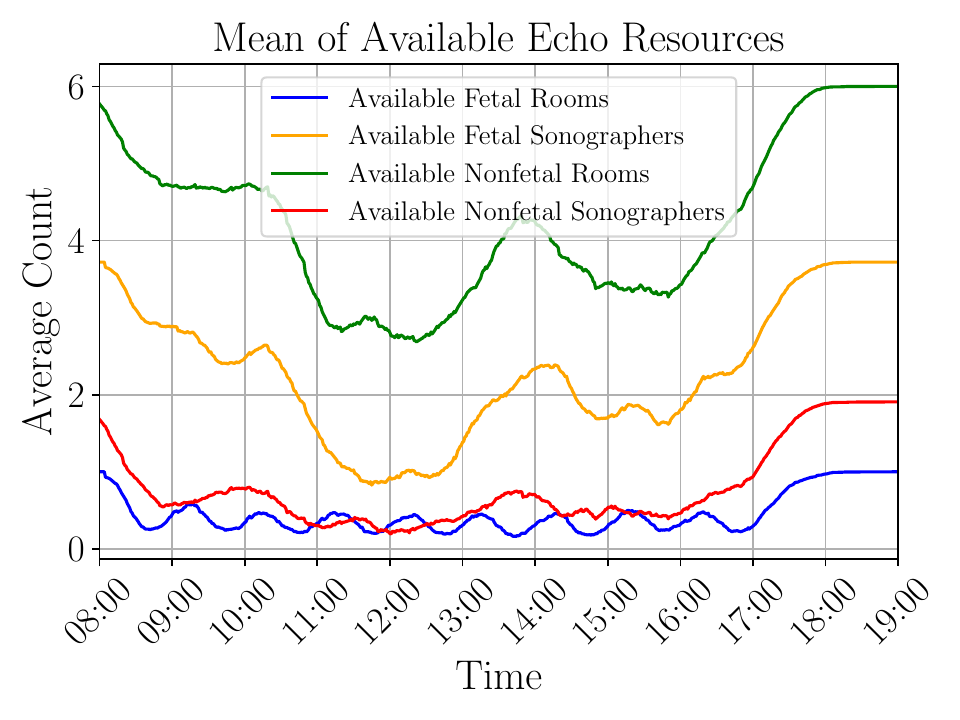}
\hfill
\includegraphics[width=0.32\textwidth]{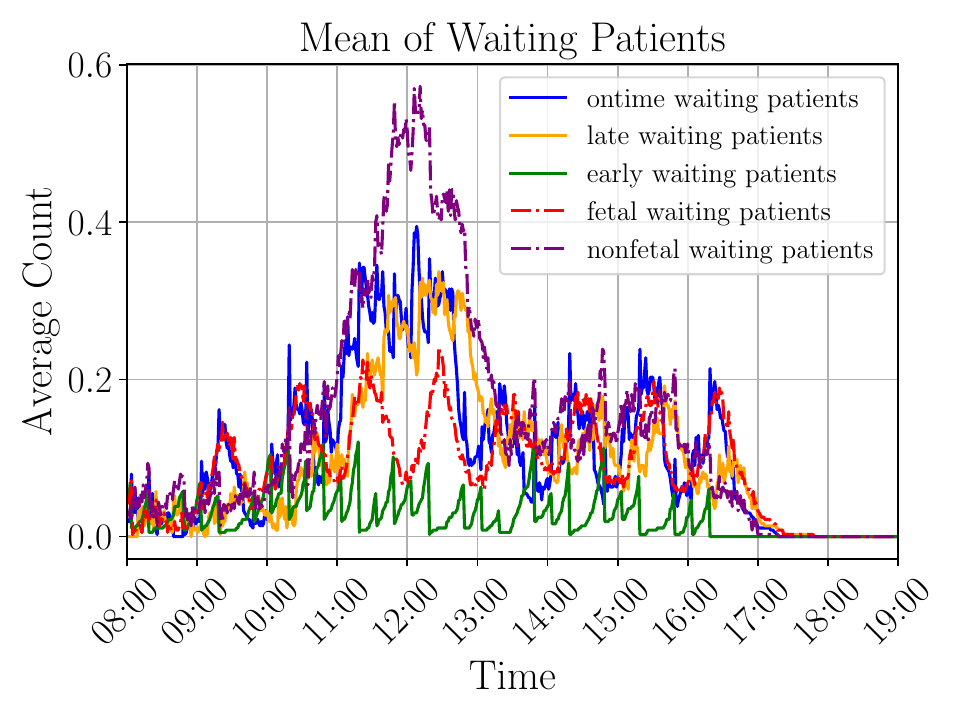}
\hfill
\includegraphics[width=0.32\textwidth]{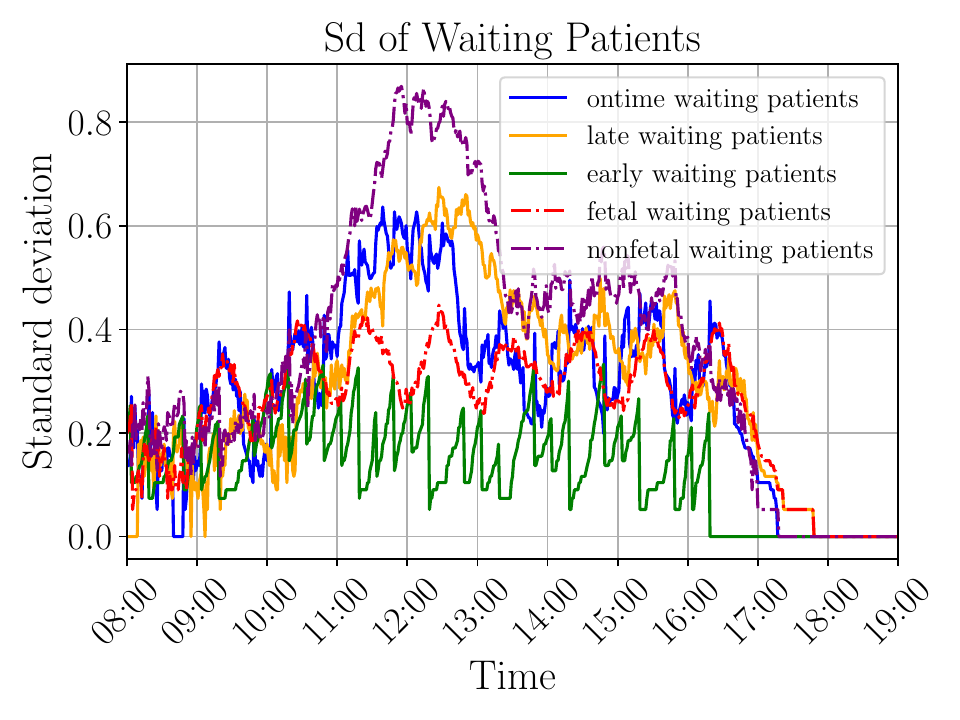}
\caption{Average results in terms of echo resources and waiting patients (mean count and standard deviation) over 365 independent replications of a representative day, for the stochastic hospital process enforcing policy 3.}
\label{fig:res_Policy3}
\end{figure}

Policy 3 introduces dedicated resources for fetal patients: one fetal echo room and one fetal sonographer are reserved exclusively for them, except once all fetal patients scheduled that day have already been accommodated, at which point the reserved resources become available for general use.
Patients arriving on time are prioritized, followed by late arrivals. For fetal patients, we first use the reserved fetal resources; if occupied, we fall back on general-duty fetal echo resources that can serve both fetal and non-fetal patients.
The allocation logic largely mirrors policy 1, with added priority handling for fetal cases. If the longest-waiting patient is fetal, we first check for exclusive fetal resources. If only one is available, we complement it with a general-duty counterpart. If neither is available, the patient continues waiting.
When the longest-waiting patient is non-fetal, we follow policy 1’s procedures but exclude the exclusively reserved fetal room and sonographer.
As shown in Figure~\ref{fig:res_Policy3}, this policy reduces the number of waiting fetal patients, albeit at the cost of increased wait times for non-fetal patients.

\subsection{Policy 4 - Dedicating fetal resources while accommodating on-time, late, and early arrivals}

\begin{figure}[ht!]
\centering
\includegraphics[width=0.32\textwidth]{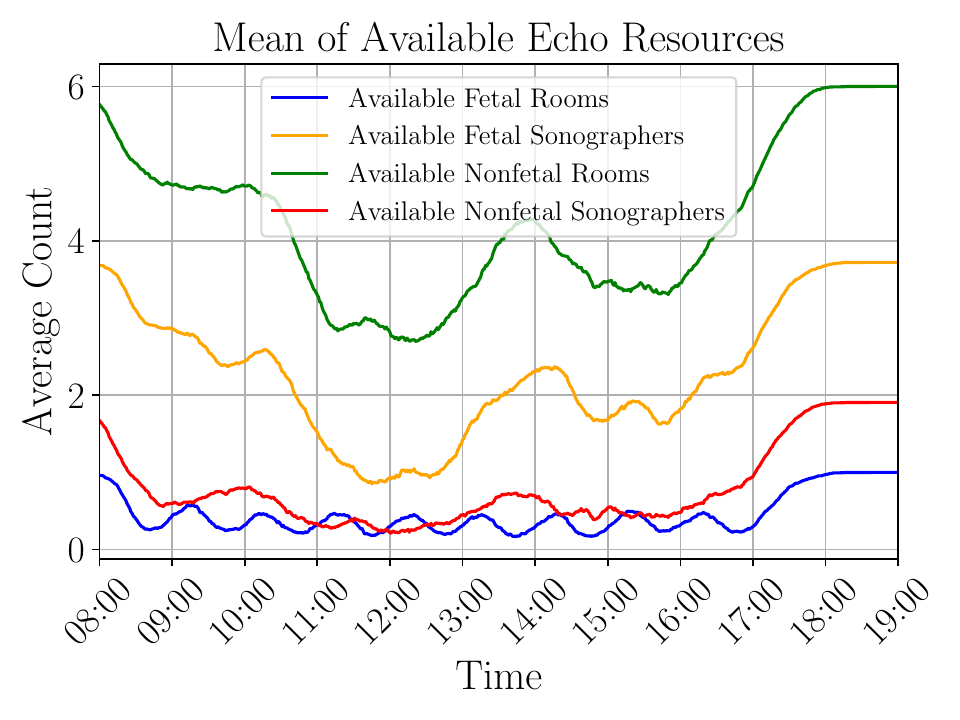}
\hfill
\includegraphics[width=0.32\textwidth]{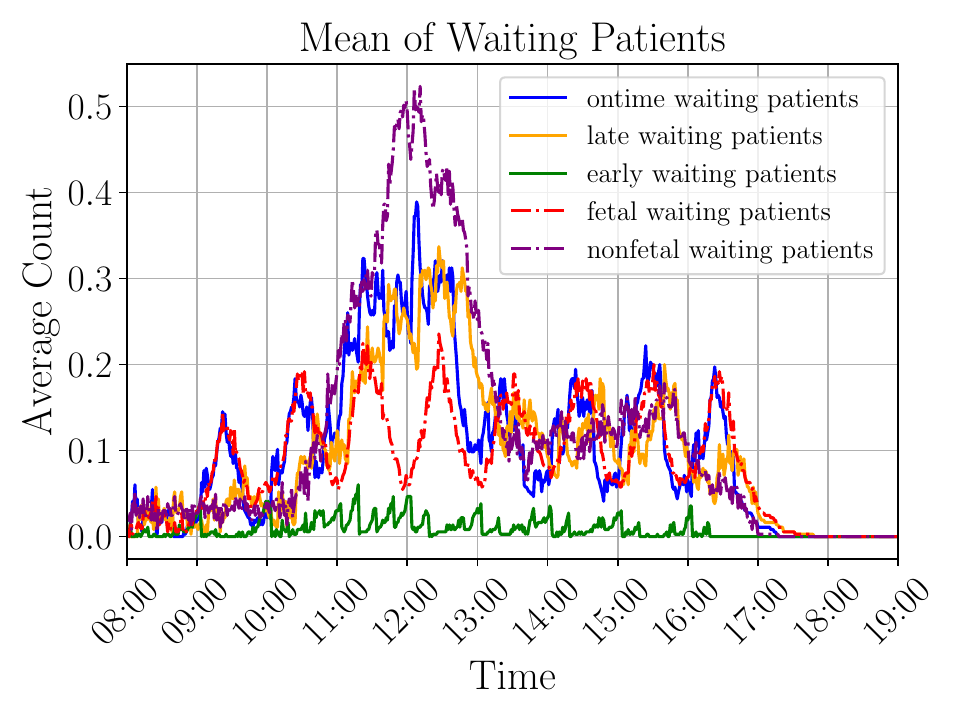}
\hfill
\includegraphics[width=0.32\textwidth]{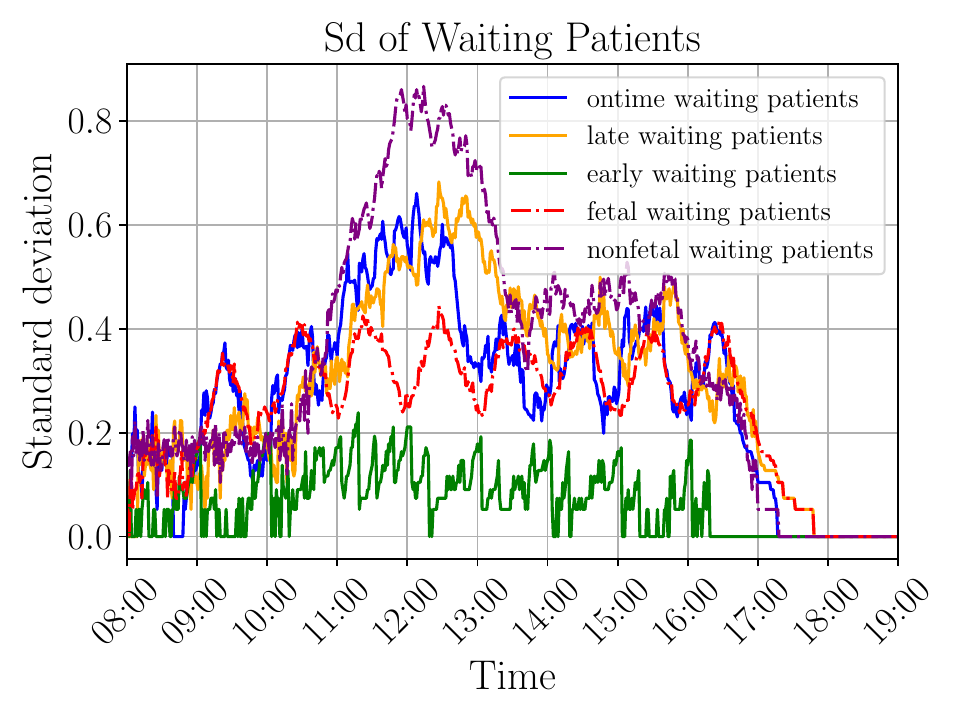}
\caption{Average results in terms of echo resources and waiting patients (average counts and standard deviation) over 365 independent replications of a representative day for the stochastic hospital process enforcing policy 4.}
\label{fig:res_Policy4}
\end{figure}

In policy 4, we basically do the same as policy 3, but also accommodate, at the end, patients who arrived early.
Results are shown in Figure~\ref{fig:res_Policy4}. As expected, the average number of waiting patients arriving early is smaller than policy 3.

\subsection{Policy 5 ($\alpha, \beta$) - Reserve echo resources in advance while accommodating patients arriving on-time without reservation and late arrivals}

In policy 5, we implement a reservation system. We reserve some echo pairs (a sonographer with an echo room forms an echo pair) for some patients, in advance, at the beginning of the day.
For sonographers who are able to do both fetal and non-fetal tests, we assign the echo room that can do both fetal and non-fetal tests with priority, to form a fetal echo pair.
We make reservations for fetal patients with fetal pairs and for non-fetal patients with non-fetal pairs, ensuring that the time gap between consecutive reservations using the same echo pair is greater than 40 minutes. 
This 40-minute criterion is a \emph{scheduling-grid} spacing rule, not a claim about exam duration: patients are selected for the same reserved echo pair only if their \emph{scheduled} times are at least 40 minutes apart, evaluated on the laboratory's discrete grid (non-fetal appointments every 15--30 minutes, fetal every 60 minutes, see Section~\ref{sec:hours_visit_schedule}).
Because of this grid, the smallest gap that actually satisfies the criterion is, for most appointments, a full 60 minutes rather than 40, so consecutive reservations on the same pair are, in practice, spaced about an hour apart. 
Separately, we ask how often an exam's \emph{actual duration} overruns this 60-minute spacing and delays the next reserved patient: under the fitted echo-duration distribution (Section~\ref{Analysis of patient's echo duration}), the probability that an exam exceeds 60 minutes is $\approx$29\%, versus $\approx$67\% for a literal 40-minute interval, so cascading delay on reserved pairs is possible but limited in practice -- consistent with the finding in Sections~\ref{sec:wait_quotas_rules}--~\ref{sec:rules_comments} that reservation-based policies increase waiting times for late-arriving patients relative to on-the-fly allocation.
If a reserved patient does not arrive within 10 minutes of their scheduled time, the reservation is forfeited and the patient is thereafter treated as a late arrival.
To accommodate additional patients and improve system efficiency, we identify echo resources available for general use at each time step. These include (1) unreserved resources and (2) reserved resources that meet two criteria.
First, all patients with earlier appointments must have been tested or missed their appointments, and second, no incoming patients should be expected within the next 40 minutes. This means that either the next scheduled appointment is at least 40 minutes later than the current time, or all patients scheduled between the current time and 40 minutes later have already been tested.
If a patient who is not late arrives with a reservation for an echo pair, and both reserved resources are available, we must check whether all patients with earlier appointments in the same echo room, with the corresponding sonographer, have completed their tests or missed their appointments.
If that is the case, the patient can utilize the reserved resources.

After accommodating patients who aren't late and have a reservation for an echo pair, we handle other patients similarly to policy 1. We first allocate resources to patients who arrived on time without reservations, followed by those who arrived late (patients without reservations who arrived late, and patients whose reservations were canceled due to late arrival), based on their waiting times.

Since there is only one fetal echo room and at least one sonographer who can perform both fetal and non-fetal tests, there is always one fetal echo pair in the system. We use $\alpha$ to denote the percentage of fetal echo pairs with reservations, which can be either 0\% or 100\% given the available resources.
We use $\beta$ to denote the percentage of non-fetal echo pairs with reservation, and evaluate the performance of policy 5 for $\beta=\{0\%, 25\%, 50\%, 75\%, 100\%\}$. We refer to these cases as case 1 to 10, as shown in Table~\ref{tab:p5_cases}.
Note that the choice $\alpha = 0\%, \beta = 0\%$ leads to a policy that coincides with policy 1. The results for cases 1 to 10 are reported in Figure~\ref{fig:res_Policy5 0 0} to Figure~\ref{fig:res_Policy5 100 100}, respectively in the appendix. 
Here we only show case 10 for brevity.
\begin{table}[ht!]
\centering
\begin{tabular}{c c c | c c c}
\toprule
{\bf Case} & $\boldsymbol{\alpha}$ & $\boldsymbol{\beta}$ & {\bf Case} & $\boldsymbol{\alpha}$ & $\boldsymbol{\beta}$\\
\midrule
Case 1 & 0\% & 0\% & Case 6 & 100\% & 0\%\\
Case 2 & 0\% & 25\% & Case 7 & 100\% & 25\%\\
Case 3 & 0\% & 50\% & Case 8 & 100\% & 50\%\\
Case 4 & 0\% & 75\% & Case 9 & 100\% & 75\%\\
Case 5 & 0\% & 100\% & Case 10 & 100\% & 100\%\\
\bottomrule
\end{tabular}
\caption{List of analyzed parameters for policy 5.}\label{tab:p5_cases}
\label{table:2}
\end{table}

\begin{figure}[ht!]
\centering
\includegraphics[width=0.32\textwidth]{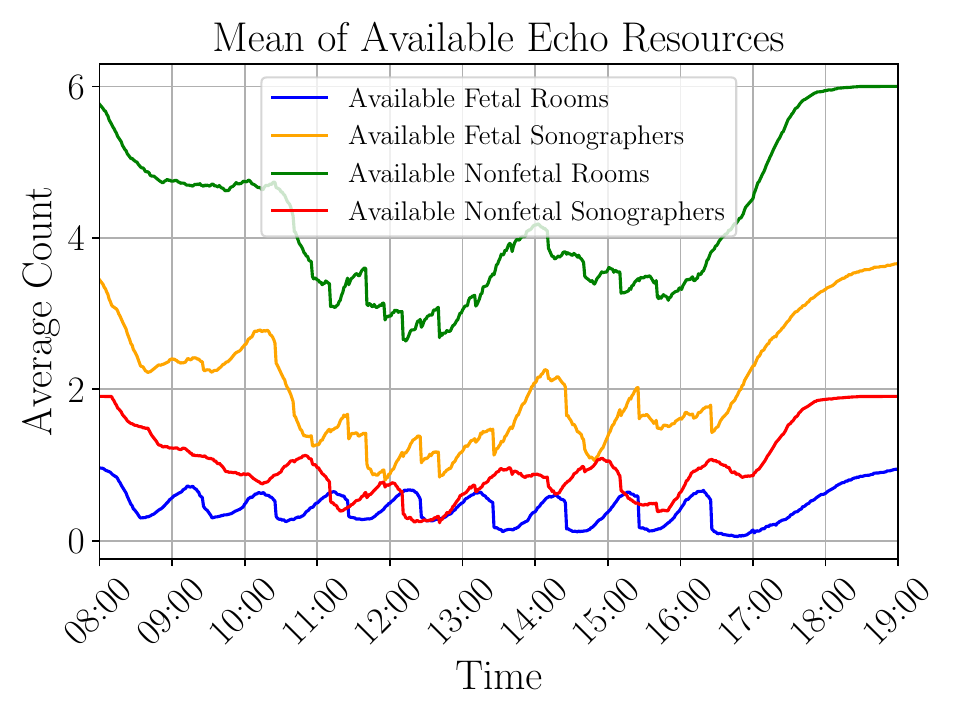}
\hfill
\includegraphics[width=0.32\textwidth]{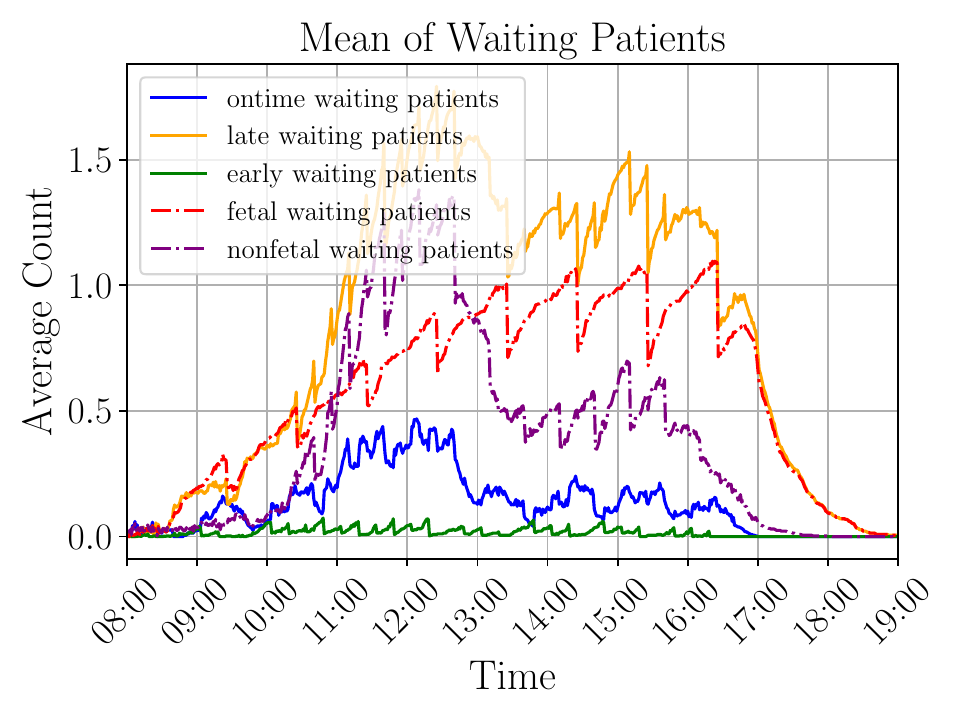}
\hfill
\includegraphics[width=0.32\textwidth]{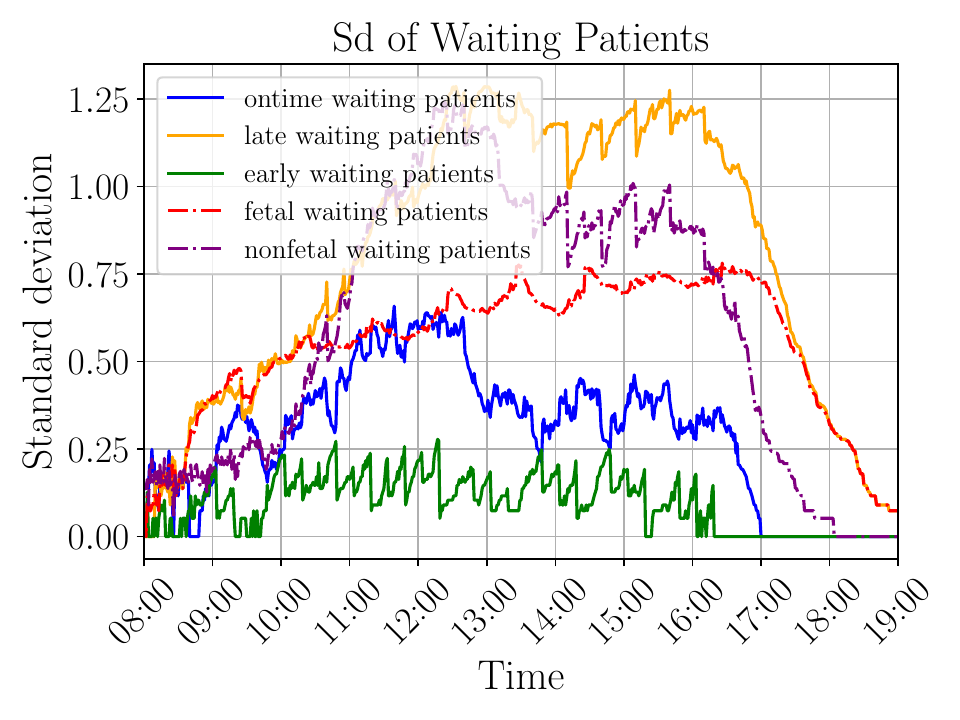}
\caption{Average results in terms of echo resources and waiting patients over 365 independent replications of a representative day for the stochastic hospital process enforcing policy 5 with $\alpha = 100\%$ and $\beta = 100\%$. Additional results are reported in the Appendix~\ref{policy 5}.}\label{fig:res_Policy5 100 100_main}
\end{figure}

\subsection{Policy 6 ($\alpha, \beta$) - Reserve echo resources in advance while accommodating patients arriving on-time without reservation, then late and early arrivals without reservation}

In policy 6 ($\alpha, \beta$), we essentially follow the same procedure as in policy 5 ($\alpha, \beta$), with an additional step at the end: accommodating patients who arrived early and without reservations. At each time step, we first implement policy 5 ($\alpha, \beta$), which involves prioritizing patients who arrived on time and late.
Following this, we allocate any remaining resources to patients who arrived early and without reservations. This process includes checking for patients who arrived early, sorting them in descending order of their waiting times (calculated as the difference between the current time and their scheduled time), and then allocating any available echo resources for general duties to these patients, applying the same non-fetal-priority tie-breaking rule defined in Section~\ref{sec:predefined_policies}.
\rev{Note that policy 6 ($\alpha=100\%, \beta=100\%$) is the closest to the current policy implemented in the hospital. Additionally, the choice $\alpha = 0\%, \beta = 0\%$, produces a policy which is identical to policy 2.}
The results for cases 1 to 10 are reported in Figure~\ref{fig:res_Policy6 0 0} to Figure~\ref{fig:res_Policy6 1 1}, respectively in the Appendix. Here we only show case 10 for brevity.

\begin{figure}[ht!]
\centering
\includegraphics[width=0.32\textwidth]{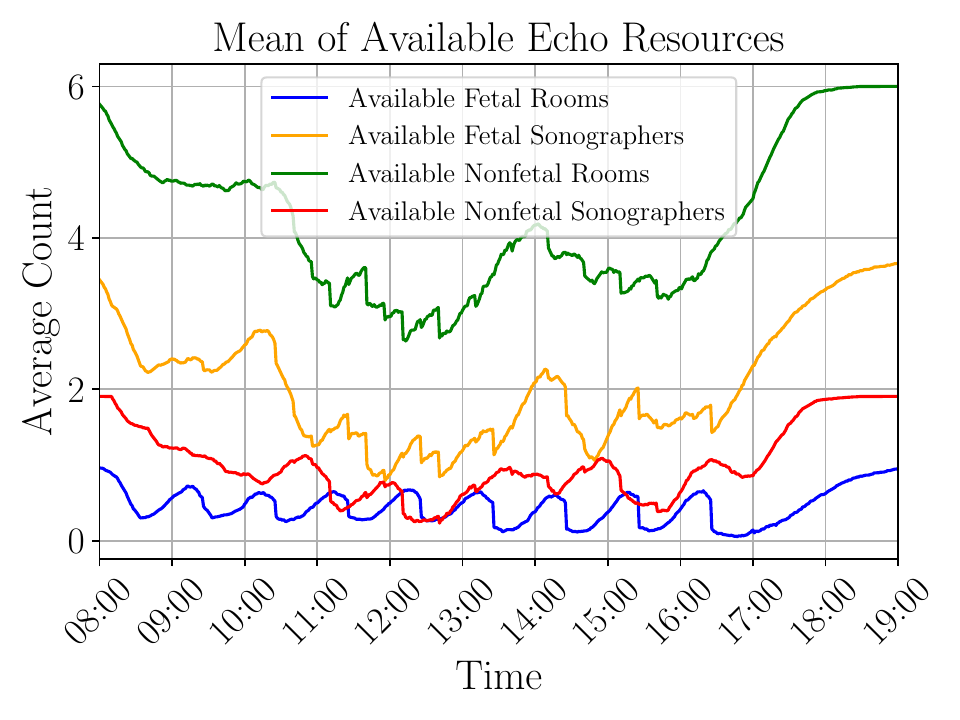}
\hfill
\includegraphics[width=0.32\textwidth]{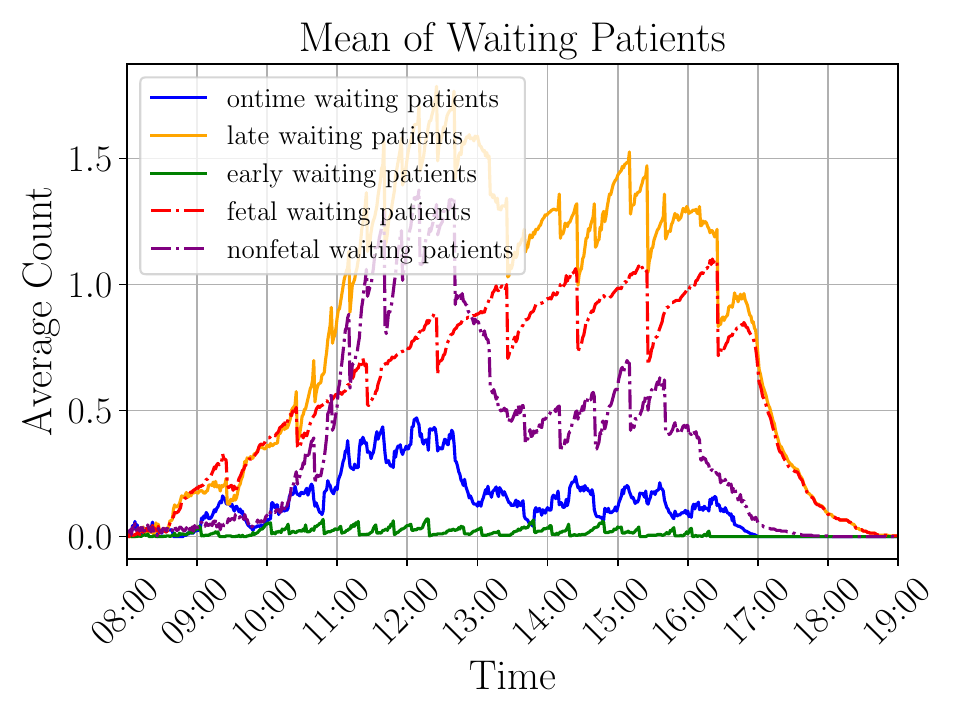}
\hfill
\includegraphics[width=0.32\textwidth]{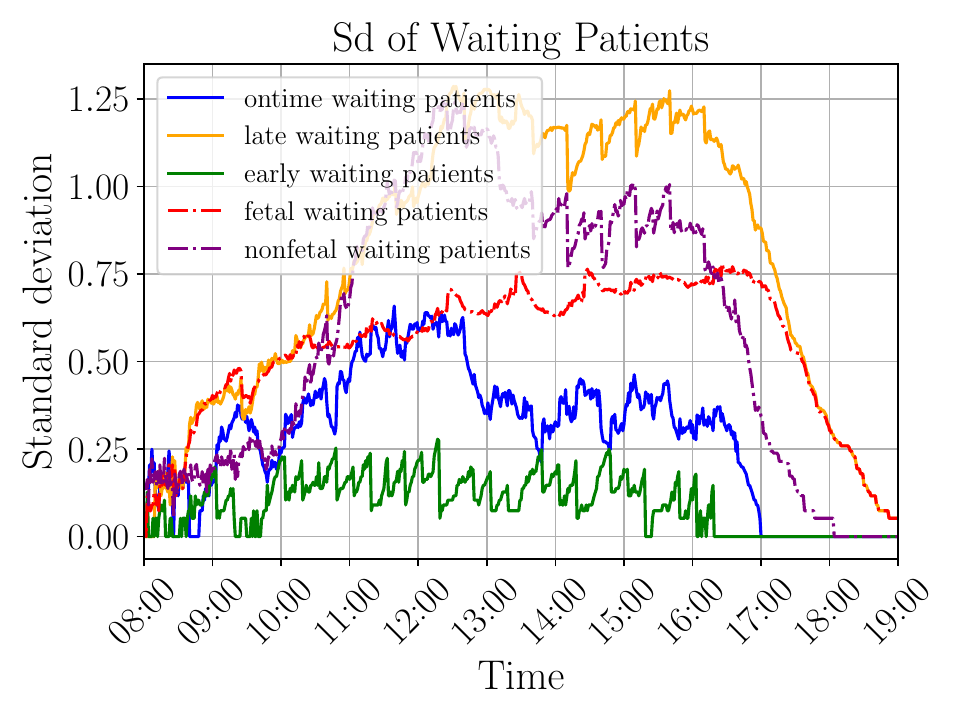}
\caption{Average results in terms of echo resources and waiting patients over 365 independent replications of a representative day, for the stochastic hospital process enforcing policy 6 with $\alpha = 100\%$ and $\beta = 100\%$. Additional results are reported in Section~\ref{policy 6}.}\label{fig:res_Policy6 1 1_main}
\end{figure}

%
\subsection{Average waiting times and sonographer quotas for the rule-based policies over a year of operation}\label{sec:wait_quotas_rules}
\begin{figure}[ht!]
\centering
\includegraphics[width=0.8\textwidth]{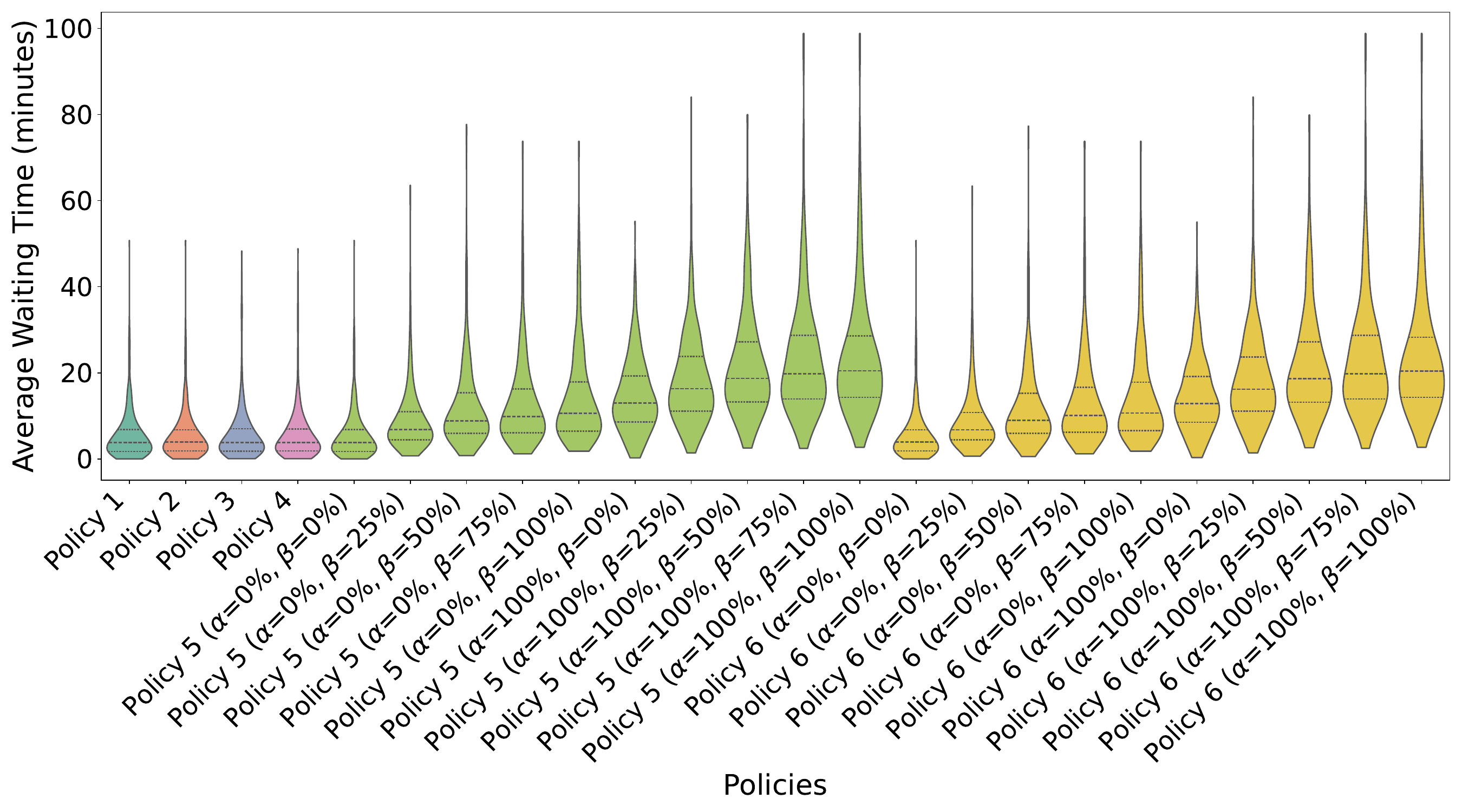}
\caption{Violin plot of average patient waiting times.}
\label{fig:waiting_time}
\end{figure}

\begin{figure}[ht!]
\centering
\includegraphics[width=0.8\textwidth]{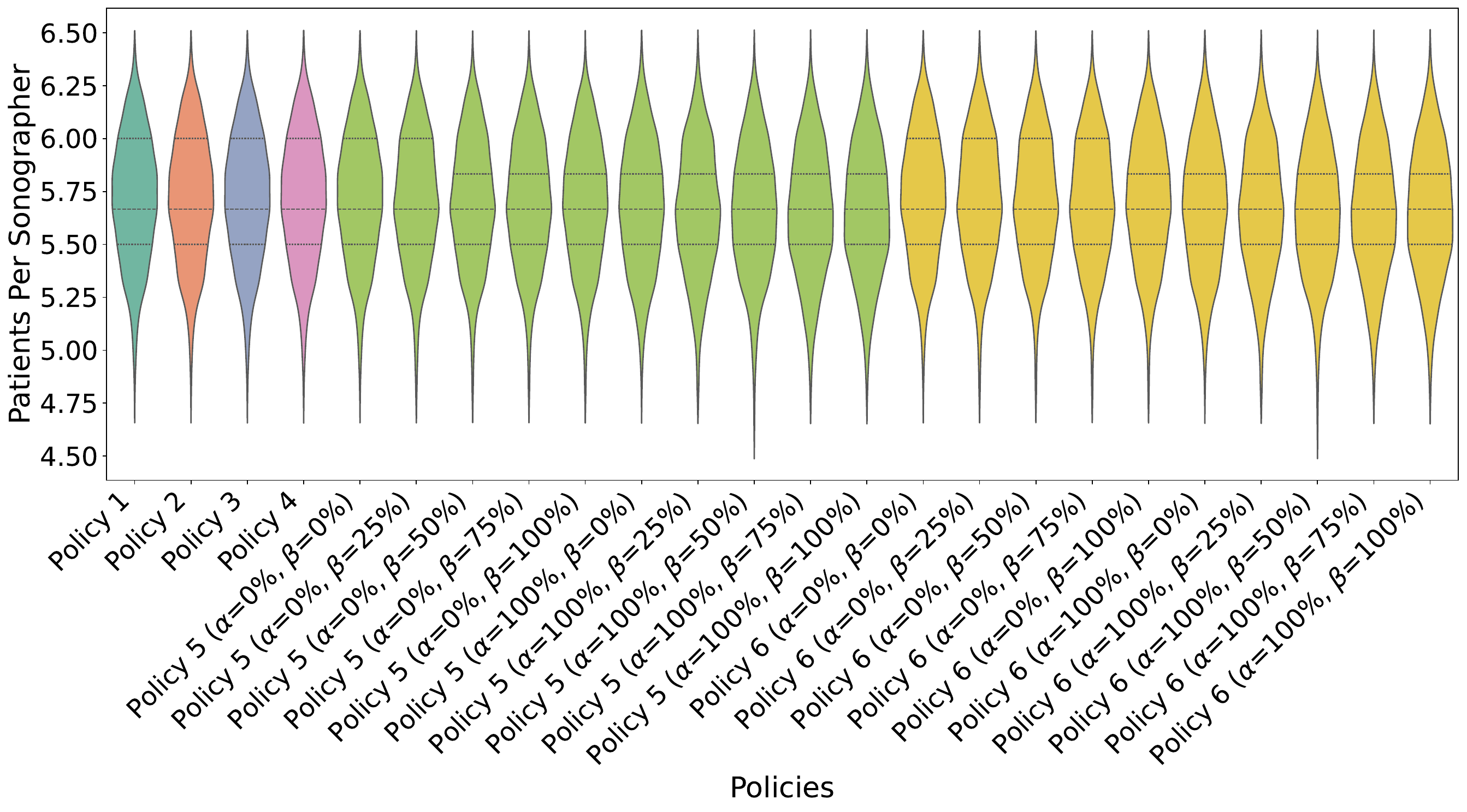}
\caption{Violin plot of average sonographer quotas.}
\label{fig:workload}
\end{figure}

We present the distributions of average patient waiting times and sonographer quotas using violin plots, based on 365 independent replications of a representative day, in Figures~\ref{fig:waiting_time} and~\ref{fig:workload}, respectively. 
Figure~\ref{fig:waiting_time} shows that policies~1--4 clearly outperform the remaining policies in terms of reducing average patient waiting time. 
Because all policies share common random number streams (identical arrival draws, durations, no-shows, and leave events), this gap reflects the policies themselves rather than simulation noise: it is large relative to the spread of each policy's own distribution and holds consistently across all 365 replications, so the ranking is not an artifact of sampling variability. 
This result demonstrates that scheduling strategies that prioritize patients based on their actual arrival times lead to more efficient service and shorter waiting times.
From Figure~\ref{fig:workload}, we observe that the number of patients examined by each sonographer is similar across all policies, and the total number of tests performed is close to the total number of patients scheduled, suggesting that the workload is comparable. 
Concretely: six sonographers seeing $\approx$5.8 patients/day each gives $\approx$34.8 tests, against 38 scheduled patients of whom $\approx$3.8 are expected no-shows ($\approx$34.2 expected attendees).
This indicates that although average patient waiting time varies across different policies, most patients are still accommodated by the end of the day under all implemented policies.
Because all policies share common random number streams (identical arrival draws, durations, no-shows, and leave events), this gap reflects the policies themselves rather than simulation noise: it is large relative to the spread of each policy's own distribution and holds consistently across all 365 replications, so the ranking is not an artifact of sampling variability. 
This result demonstrates that scheduling strategies that prioritize patients based on their actual arrival times lead to more efficient service and shorter waiting times. These same five days of real operational data (recording \emph{echo scheduled time}, \emph{echo time in}, and \emph{echo time out}) were used to calibrate the simulator's inputs; what has not been checked here is whether the simulator's aggregate output-level behavior reproduces the corresponding real-world outputs. 
Day-to-day variability in patient volume and timing over these five days is itself substantial -- comparable in magnitude to the differences reported between policies -- so a rigorous comparison at this sample size would largely reflect idiosyncratic day-to-day variation in the observed process rather than a genuine discrepancy between the simulator and reality. 
A quantitative validation subsection comparing, at minimum, the mean/distribution of patient wait, mean exams per sonographer per day, and end-of-day completion time (simulated versus observed) against a larger real-world dataset spanning many weeks is left as a direction for future work; Policy~6 (100\%, 100\%), which is designed to mirror current scheduling practice most closely, is intended to provide a qualitative point of reference in the interim.

\subsection{Comments}\label{sec:rules_comments}

\rev{Our preliminary analysis of a family of rule-based policies leads to two key observations.}
First, accommodating patients who arrive early can enhance the utilization of echo resources. However, the reduction in available resources resulting from accommodating early arrivals, may lead to longer waiting times for patients arriving late.
Second, a reservation-based policy can be effective in reducing the waiting time for patients arriving early. However, this strategy does not seem effective in reducing the waiting time for patients arriving on-time, and may lead to a significant backlog for patients arriving late.
This is consistently observed when reservations are made either for fetal or non-fetal echo pairs, becoming more severe as $\alpha$ and $\beta$ increase.
Among them, policies 1 and 2 are particularly effective in minimizing both the number of waiting patients and their average waiting times.
Given these findings, we choose to focus on on-the-fly resource allocation, and investigate whether an optimal on-the-fly policy can be found through a Markov Decision Process (MDP) that further improves upon these baselines.
We then compare the optimal MDP policy with policy 2.

\section{Markov Decision Process for echo resource allocation}\label{sec:mdp}

In this section, we formalize the echo resource allocation problem as a MDP.
This follows the framework where resources are allocated on-the-fly.

We denote the number of patients currently waiting by $W$, where the subscripts $f$ and $n$ are used to indicate \emph{fetal} and \emph{non-fetal} patients, $b$ as resources for \emph{both} patient types, and subscripts $\{e, t, l\}$ instead represent patients who arrived \emph{early}, \emph{on time}, or \emph{late} for their scheduled visit, respectively.
Additionally, $R$ is used to quantify the number of available echo rooms, $S$ the number of available sonographers and $L$ the number of sonographers currently on leave.

\subsection{States}

We write the elapsed simulation time in minutes since 8:00~AM opening as $t$; this is the same quantity referred to as the ``decision horizon'' in the Discussion (Section~\ref{sec:discussion}). Decisions are made at a fixed one-minute granularity: the simulated clock advances one minute between successive decision epochs, so each unit of $t$ corresponds to one decision step, and the effective planning horizon under $\gamma=0.99$ spans on the order of several hundred minutes.

The state of our echo lab environment is represented as a vector containing the following elements:
\[
\boldsymbol{\mathcal{S}} = [W_{f,e}, W_{f,t}, W_{f,l}, W_{n,e}, W_{n,t}, W_{n,l}, R_n, R_b, S_n, S_b, L_n, L_b, t]^{T}.
\]
where $W_{f,e}$, $W_{f,t}$, $W_{f,l}$ is the number of waiting fetal patients who arrived early, on time, and late, respectively, whereas $W_{n,e}$, $W_{n,t}$, $W_{n,l}$ are similarly used for non-fetal patients.
The number of echo rooms and sonographers available for non-fetal and both fetal and non-fetal patients is denoted by $R_n$, $R_b$ and $S_n$, $S_b$.
Finally, the number of non-fetal and both fetal and non-fetal sonographers currently on leave is quantified by $L_n$, $L_b$, and $t$ indicates the current time.

\subsection{Actions}

Based on these states, we seek to find the optimal action that maximizes the reward (or minimizes the penalty).
An action consists of assigning a pair of resources (room, sonographer) to a waiting patient who arrived either early, on time, or late.
Therefore, an action is represented by the vector:
\[
\boldsymbol{\mathcal{A}} = [A_{f,e}, A_{f,t}, A_{f,l}, A_{n,e}, A_{n,t}, A_{n,l}]^{T}.
\]
(reordered here to match the $(e,t,l)$ ordering used for the state vector $\boldsymbol{\mathcal{S}}$ above; this is a notational change only) where the subscripts have the same meaning as those used for $W$.
Each action can take a value in $\{0,1,2\}$, representing the number of patients assigned to an appropriate pair of resources (echo room and sonographer). 
The cap of 2 per component was chosen based on our earlier analysis of the rule-based heuristic policies (Section~\ref{sec:predefined_policies}): under those policies, it was rare for more than two patients per category to require simultaneous allocation within a single one-minute decision step. 
Capping the action space at 2 per component keeps the total number of discrete actions manageable ($3^6=729$ rather than growing combinatorially with a higher cap), a deliberate trade-off between the network's expressiveness and the size of the action space.
The maximum number of accommodated patients at each component of the action array, is constrained by the number of waiting patients and the availability of echo resources.
To improve the flexibility of the allocation, we allocate non-fetal echo resources to non-fetal patients with priority.

\subsection{Rewards}

Negative rewards (penalties) are defined as follows.
For late arrivals (more than 10 minutes past their scheduled time), each waiting patient incurs a penalty of 2.
For patients who arrive on-time (within 10 minutes of their scheduled time), each waiting patient results in a penalty of 4.
For each pair of idle resources (one sonographer and one echo room), a penalty of 10 is assigned.
We now state the reward function explicitly. Let $\mathcal{W}_{ontime}(s)$ and $\mathcal{W}_{late}(s)$ denote the sets of on-time and late waiting patients in state $s$, and let $f_p \in \{0,1\}$ indicate whether patient $p$ requires a fetal exam. 
Let $I_{pair}(s), I_{pair}^{fetal}(s) \in \{0,1\}$ indicate, respectively, whether at least one available non-fetal-restricted sonographer-room pair, and at least one available fetal-qualified sonographer-room pair, exist in state $s$. 
The penalty at each one-minute step $t$ is 
\begin{equation} 
\begin{split} 
R(s) = \sum_{p \in \mathcal{W}_{ontime}(s)} 4 + \sum_{p \in \mathcal{W}_{late}(s)} 2 + \sum_{p \in \mathcal{W}_{ontime}(s) \cup \mathcal{W}_{late}(s)} 10\big[(1-f_p)I_{pair}(s) + f_p I_{pair}^{fetal}(s)\big] \\ + \mathbf{1}[t \geq t_{5pm}]\sum_{p:\,\text{waiting}(p)} 10 + \mathbf{1}[t \geq t_{7pm}]\sum_{p:\,\text{waiting}(p)\,\lor\,\text{testing}(p)} 50.
\end{split} 
\end{equation} 
The 5:00~PM and 7:00~PM terms are charged per qualifying patient at every one-minute step for as long as the condition holds, not once. 
Because the idle-pair term is gated on $\mathcal{W}_{ontime}(s) \cup \mathcal{W}_{late}(s) \neq \emptyset$, an idle room--sonographer pair contributes no penalty unless a compatible patient is actually waiting.
We impose a penalty of 10 if any patient is still waiting after 5:00 PM.
Furthermore, if any patient remains in the hospital after 7:00 PM, we apply an additional penalty of 50. We assign large penalties for cases in which patients are unable to complete all examinations before 5:00 PM and 7:00 PM to encourage the agent to schedule patients so that all work is finished before these two cutoff times.
The reward is defined over the number of waiting patients per class, not over individual waiting time; every downstream evaluation in this paper, however, is reported in weighted waiting time in minutes (e.g., Figures~\ref{fig:waiting_time_comparison}, \ref{fig:boundary}, and the sensitivity-analysis figures in Section~\ref{sec:sensitivity}). 
These are related but not equivalent objectives, and the RL agent is therefore evaluated on a quantity distinct from the one it was trained to optimize; this is a limitation we return to in the Discussion (Section~\ref{sec:discussion}).
We deliberately keep the reward as a per-step, per-patient count rather than switching to accumulated waiting time: waiting time depends on patient arrivals, sonographer breaks, and absences in a highly stochastic, hard-to-model fashion, and using it directly as the training signal made the reward substantially harder for the network to fit. The per-step count is therefore a deliberate simplification of the training objective relative to the reported evaluation metric.
Fetal and non-fetal waiting patients currently carry identical per-class weights (2 late / 4 on-time). Figure~\ref{fig:boundary} nonetheless shows a persistent multi-minute gap between fetal and non-fetal mean waiting time under every policy, including policies 3--4, which dedicate fetal capacity (see Section~\ref{sec:lp} below). 
We do not give fetal exams special weight in the reward: fetal and non-fetal waiting patients are assigned identical per-class penalties, and we do not report a separate fetal-weighted reward experiment in this study. Fetal prioritization is instead handled through the resource-compatibility constraints of the MDP rather than the reward weights; reward-level fetal weighting is a direction we leave for future work.
The relative weights were chosen to reflect operational priorities observed during policy training, rather than derived from a formal clinical cost model, and we clarify the rationale here. 
On-time patients are weighted more heavily than late patients (4 vs.\ 2) because on-time arrival reflects compliance with the assigned schedule; a larger penalty for keeping on-time patients waiting discourages a policy that defers them in favor of already-late arrivals, so that patients who arrive as scheduled are not systematically disadvantaged. 
The idle-pair penalty (10) is set substantially larger than the per-patient waiting penalties to prevent a degenerate policy, observed early in training, in which the agent leaves compatible room--sonographer pairs unused even while a matching patient is waiting; the large weight strongly discourages such under-utilization of available capacity. 
Finally, the 5:00~PM and 7:00~PM cutoff penalties (10, escalating to 50) encode a strong preference for completing all scheduled examinations within the operating day: a moderate penalty at 5:00~PM signals that operations should be concluding, while the much larger penalty at 7:00~PM reflects the substantially higher operational cost of a test still incomplete two hours after closing. 
The specific numeric ratios were tuned informally rather than derived from a clinical cost model; a systematic sensitivity analysis over these weights is left for future work.
%
\subsection{Overview of the algorithm}

Even with a custom-built environment, the decision problem entails an overwhelming number of state–action pairs, rendering exact solution methods impractical.
We therefore frame the task as a reinforcement-learning problem and train an agent from scratch.
We employ a \emph{Dueling Deep Q-Network} (Dueling DQN)~\cite{10.5555/3045390.3045601} architecture combined with Double Q-Learning~\cite{vanhasselt2010doubleq,10.5555/3016100.3016191} to determine an optimal policy for the MDP discussed in the previous section. 
This combination of components, together with prioritized replay described below, follows the general recipe of combining DQN extensions studied systematically in Rainbow~\citep{hessel2018rainbow}.
These components follow standard practice in the DQN literature; we do not conduct ablations isolating each component's individual contribution.
In our setting, characterized by a large number of possible states, it is essential to use neural networks as function approximators for $Q(s,a)$, to determine an optimal action for each state.
However, traditional Q-learning is known to suffer from overestimation, as first investigated by Thrun and Schwartz~\cite{Thrun1999IssuesIU}.
Overestimation can bias action selection toward overestimated actions. To address this problem, Double Q-learning uses two neural networks, one for action selection and the second for action evaluation~\cite{10.5555/3016100.3016191}.
This algorithm has demonstrated superior performance on Atari games compared to using a single neural network for both action selection and evaluation.
Beyond its use in video games, Double Q-learning has been applied to hospital bed--allocation policy design as a proof-of-concept using real clinical data~\cite{9388626}.

In this study, we employ a Dueling DQN architecture that decomposes the Q-value function into two streams: a value stream $V(s)$ that estimates the state value, and an advantage stream $A(s,a)$ that estimates the advantage of each action. 
The Q-value is then computed as:
\begin{equation}
    Q(s,a) = V(s) + \left(A(s,a) - \frac{1}{|\mathcal{A}|}\sum_{a'\in\mathcal{A}} A(s,a')\right),
\end{equation}
where the advantage values are centered by subtracting their mean, taken over the valid actions for the current state (see the discussion of action masking below).
This architecture allows the network to learn which states are valuable independently of the action taken, leading to more stable learning~\cite{10.5555/3045390.3045601}.
Our implementation consists of two networks: an \emph{Online Network} (policy network) responsible for determining actions, and a \emph{Target Network} used to compute target values for updating Q-values during training.
The network architecture comprises a shared feature extractor with two fully connected layers (13 $\rightarrow$ 128 $\rightarrow$ 128) using Layer Normalization~\citep{ba2016layernorm} and dropout (0.2), followed by separate value and advantage streams. 
Dropout is applied within the shared feature extractor, where it regularizes the online network's own Q-value estimates during training -- a standard mitigation for overfitting given the large discrete action space ($3^6=729$). 
The value used in the bootstrapped regression target, however, is computed exclusively by the target network, which is kept in evaluation mode (dropout disabled) throughout training, so the target values themselves are deterministic and unaffected by dropout. 
The action chosen for the Double DQN target, however, is selected by the online network while it remains in training mode, so this action-selection step is itself subject to dropout noise even though the resulting Q-value is evaluated deterministically by the target network.
The value stream maps the shared features to a single scalar through a 128 $\rightarrow$ 64 $\rightarrow$ 1 pathway, while the advantage stream maps to all possible actions (729 = $3^6$) through a 128 $\rightarrow$ 64 $\rightarrow$ 729 pathway.
Unlike conventional approaches that periodically copy parameters from the online to the target network, we employ \emph{soft updates} with a rate $\tau = 0.005$, updating the target network parameters $\theta'$ after each optimization step as:
\begin{equation}
    \theta' \leftarrow \tau \theta + (1-\tau)\theta',
\end{equation}
where $\theta$ are the online network parameters. This gradual update strategy has been shown to improve training stability~\cite{Lillicrap2015ContinuousCW}.

The loss function is expressed as a smooth $\ell_1$ loss (Huber loss):
\begin{equation}
    l =
    \begin{cases}
        \frac{1}{2} \delta^2, & \text{if } |\delta| < 1 \\
        |\delta| - \frac{1}{2}, & \text{otherwise}
    \end{cases}
    \quad s,s'\in\boldsymbol{\mathcal{S}},\, a,a'\in\boldsymbol{\mathcal{A}}.
\end{equation}
Because our architecture combines the dueling network above with Double Q-learning, the TD error $\delta$ is computed with the \emph{Double DQN} target -- using the online network to \emph{select} the greedy next action and the target network only to \emph{evaluate} it, which is precisely the decoupling that motivates Double Q-learning~\citep{vanhasselt2010doubleq,10.5555/3016100.3016191}:
\begin{equation}
    \delta = r + \gamma\,(1-d)\, Q'\bigl(s', \arg\max_{a'} Q(s', a'; \theta); \theta'\bigr) - Q(s, a; \theta),
\end{equation}
where $d \in \{0,1\}$ is a terminal-state indicator (masking the bootstrap term at the end of an episode). Moreover, $Q(s, a)$ is the Q-value for the current state $s$ and action $a$, $r$ is the reward received after taking action $a$, $\gamma$ is a discount factor determining the importance of future rewards (set to 0.99), and $Q'(s', \cdot; \theta')$ is the target network's value of the next state $s'$, evaluated at the action chosen by the \emph{online} network $Q(s',\cdot;\theta)$.

When selecting actions, we must consider the available options, which are constrained by the number of waiting patients in each category and the availability of echo resources.
To do so, we implement \emph{full action masking} both during action selection and network training. During action selection, we generate only the valid actions based on the current state constraints, and during training, we mask invalid actions by setting their Q-values to negative infinity before computing the loss, ensuring the network never learns to select infeasible actions.
When we choose actions during the learning process, we employ an \textit{epsilon-greedy} exploration strategy~\cite[][Chapter 2]{Sutton1998} with a decaying exploration rate ($\epsilon_{\text{max}} = 1.0$, $\epsilon_{\text{min}} = 0.01$, decay = 0.9995 per episode, reaching the floor at episode $\approx$9{,}200 -- roughly one-third of the way through the 30{,}000-episode training budget) to balance exploration and exploitation.
Specifically, we generate a number $r\sim\mathcal{U}[0,1]$; if $r < \epsilon$ (the exploration rate), we select a random action from the set of valid actions.
Otherwise, we choose the action with the highest Q-value among valid actions as predicted by the online network.

A critical component of our approach is the use of \emph{prioritized experience replay}~\cite{schaul2015prioritized}, which extends the (uniform) experience-replay mechanism originally proposed by Lin~\cite{lin1992selfimproving} and later applied with deep networks to train agents to play Atari games~\cite{article,mnih2015humanlevel}.
Rather than sampling uniformly from past transitions, prioritized replay samples transitions with probability proportional to their temporal-difference (TD) error raised to a priority exponent $\alpha_{\text{PER}} = 0.6$. 
To correct for the bias introduced by non-uniform sampling, we apply importance sampling weights with an exponent $\beta_{\text{PER}}$ that anneals from 0.4 to 1.0 during training (we use the subscript $\alpha_{\text{PER}}, \beta_{\text{PER}}$ throughout to avoid clashing with the reservation fractions $\alpha,\beta$ of Section~\ref{sec:predefined_policies}). 
We confirm this matches the intended schedule: $\beta_{\text{PER}}$ reaches its ceiling of 1.0 after 6{,}000 gradient steps (under 9 of the 30{,}000 training episodes), so importance-sampling bias correction is effectively full for essentially all of training.
This allows the agent to learn more efficiently from surprising or informative transitions.
This combination of dueling architecture, soft target updates, full action masking, and prioritized experience replay follows standard DQN practice; we did not ablate the individual contribution of each component.
Additional details on the training procedure are reported in Appendix~\ref{sec:Policy-training}.

\section{Results for RL policies}\label{sec:lp}

\subsection{Learning curve}

To illustrate the agent learning dynamics, we analyze the temporal evolution of the penalty during training (one episode corresponds to one simulated day). High-frequency fluctuations are mitigated by applying a moving average with a uniform kernel and a window size of 200 episodes.
The trace plot in Figure~\ref{fig:learning_curve} provides a cleaner representation of how the agent's total penalty evolves over time. 
The trend is a rapid decrease during roughly the first 1{,}000--2{,}000 episodes, after which the smoothed penalty flattens into a noisy plateau for the remaining training; there is no evidence of a continued systematic decrease by episode 30{,}000.

\subsection{Policy comparison: Policies 1,2,3,4 and RL policy}

\begin{figure}[ht!]
\centering
\begin{minipage}[t]{0.48\textwidth}
\centering
\includegraphics[width=\textwidth]{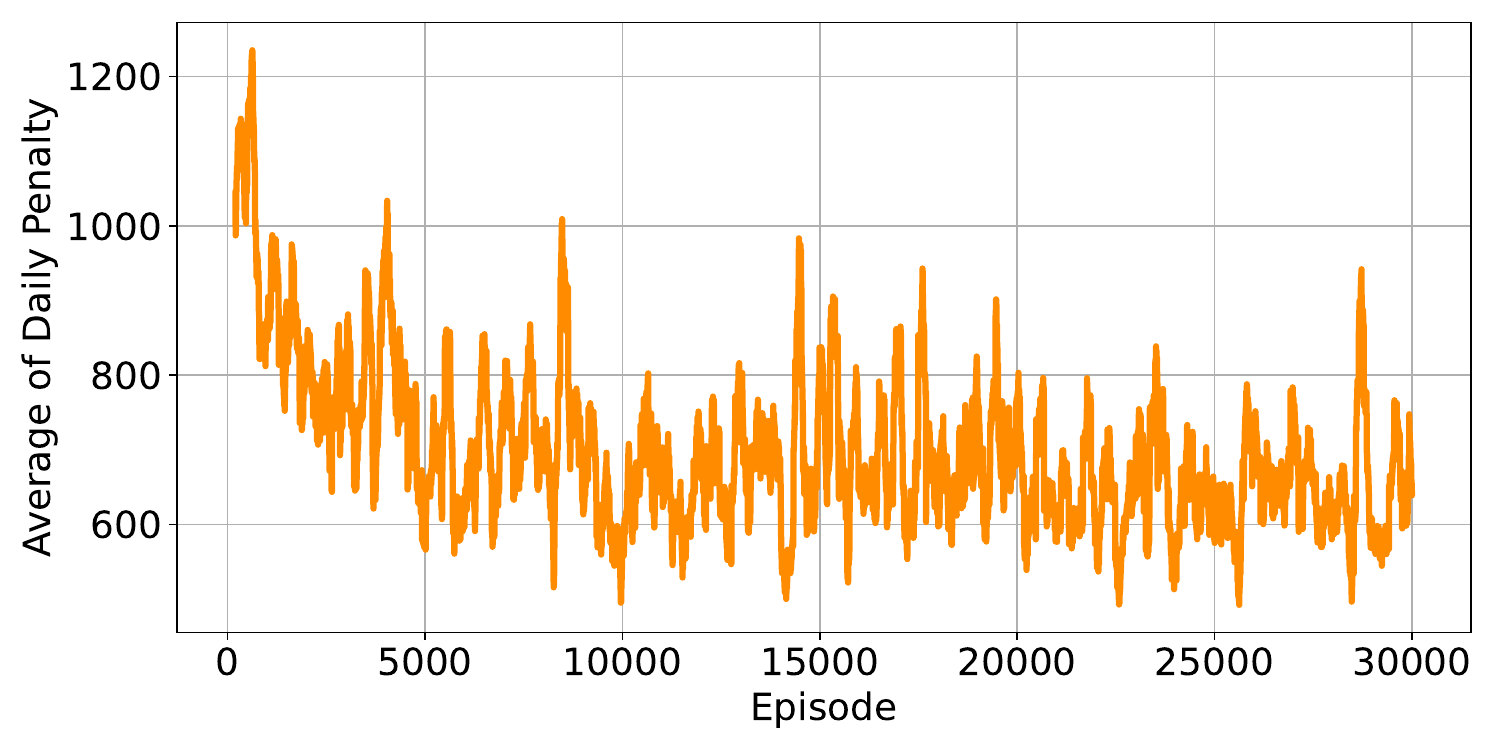}
\caption{Running average (200-day window) of daily penalty.}
\label{fig:learning_curve}
\end{minipage}
\hfill
\begin{minipage}[t]{0.48\textwidth}
\centering
\includegraphics[width=\textwidth]{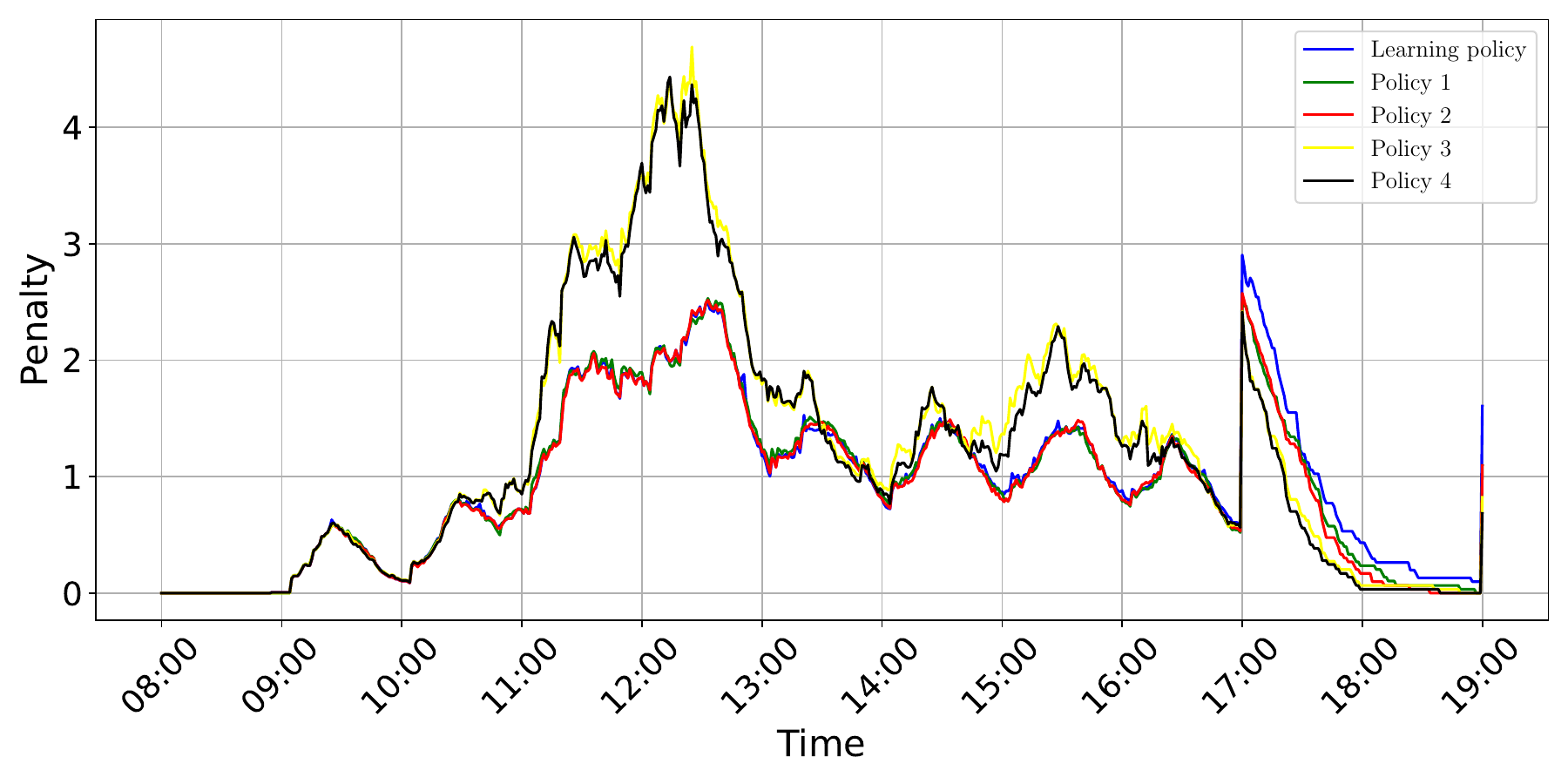}
\caption{Average penalty over time for different policies, based on a one-year simulation.}
\label{fig:avg_daily_penalty_comparison}
\end{minipage}
\end{figure}

To evaluate the dynamics of the system under different policies, we conducted a simulation over 365 independent replications of a representative day, implementing policy 1, policy 2, policy 3, policy 4 and the RL policy. 
Figure~\ref{fig:avg_daily_penalty_comparison} presents the average penalty accrued over time for each policy.
From the plot, we observe that policies 3 and 4 incur substantially higher penalties during regular operating hours (10:30 AM to 5:00 PM). However, after 5:00 PM, these same policies accumulate noticeably lower penalties compared with the others.
This pattern indicates that policies 3 and 4 are less effective at managing patient flow before the 5:00~PM closing time, but are more successful at clearing the remaining backlog afterwards, thereby avoiding large late-day penalties.
The RL policy exhibits a penalty trajectory similar to policies 1 and 2 before 5:00 PM but with higher penalty after 5:00 PM.

\subsection{Policy comparison: waiting times and sonographer quotas}

\begin{figure}[ht!]
\centering
\begin{minipage}[t]{0.48\textwidth}
\centering
\includegraphics[width=\textwidth]{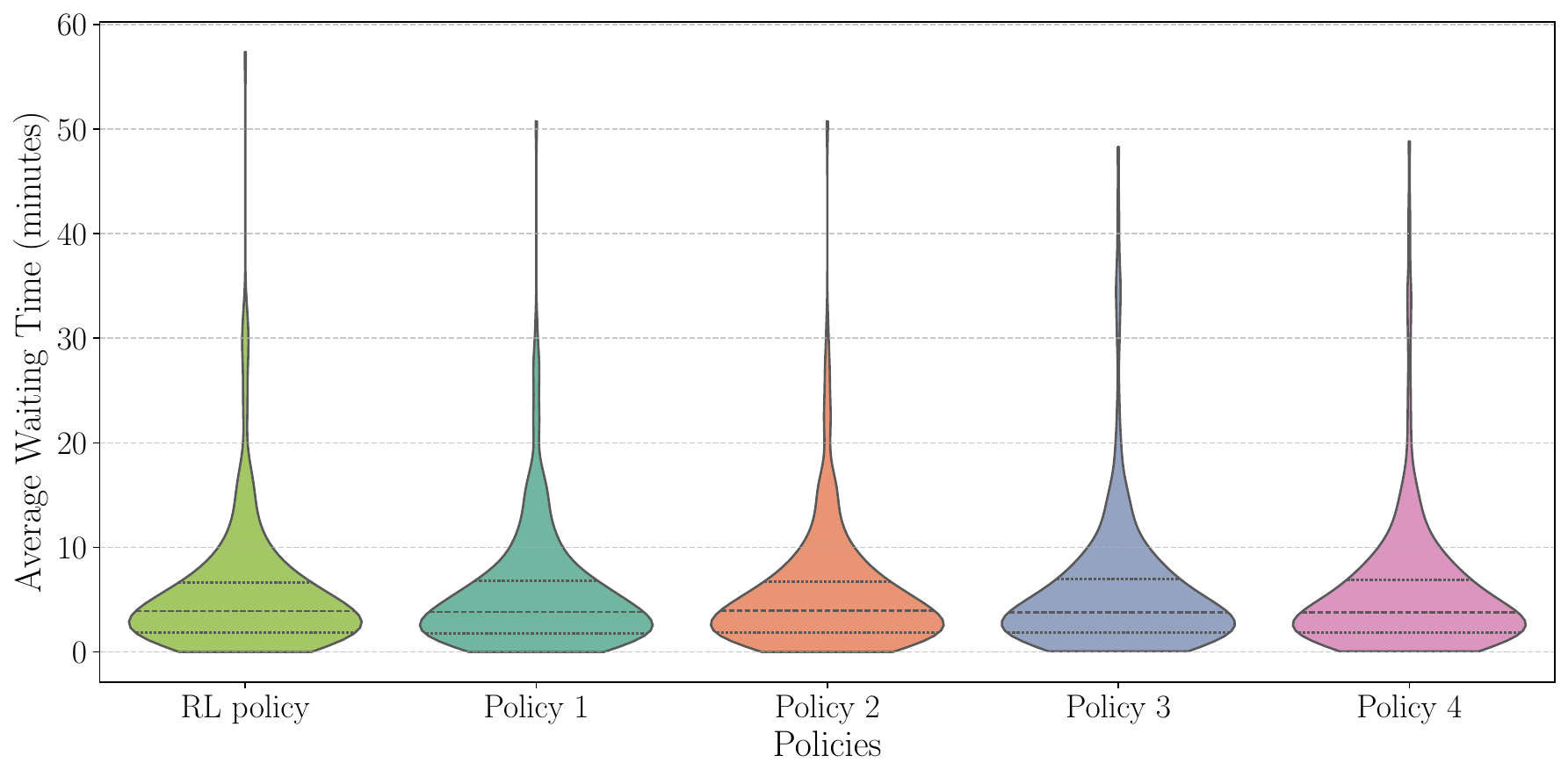}
\caption{Violin plot of average daily waiting time under different policies.}
\label{fig:waiting_time_comparison}
\end{minipage}
\hfill
\begin{minipage}[t]{0.48\textwidth}
\centering
\includegraphics[width=\textwidth]{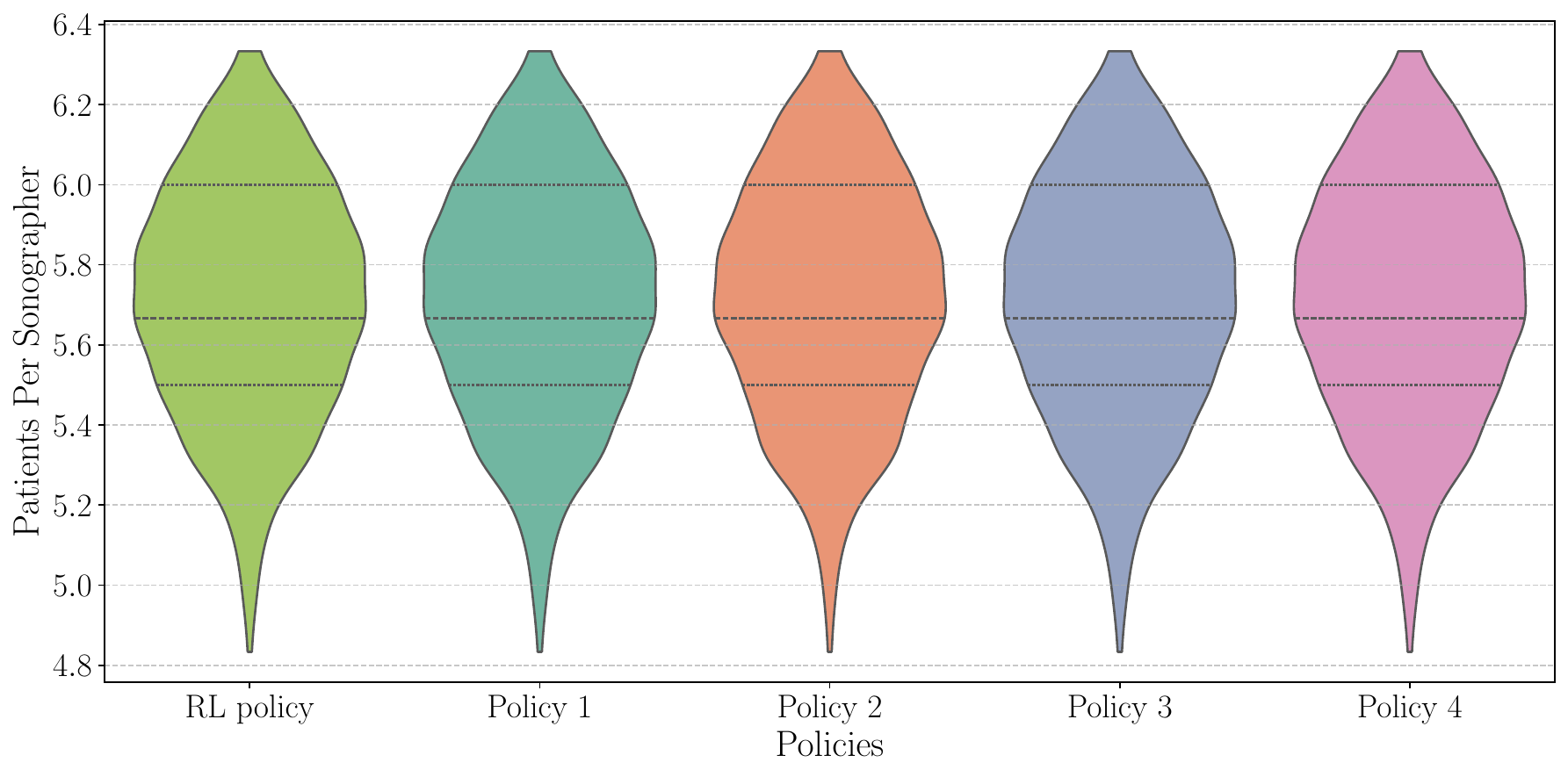}
\caption{Violin plot of average daily sonographer quotas under different policies.}
\label{fig:workload_comparison}
\end{minipage}
\end{figure}

To further compare RL and rule-based policies, we evaluate two key metrics over 365 independent replications of a representative day: average waiting time, and sonographer quota.
As shown in Figure~\ref{fig:waiting_time_comparison}, the average weighted waiting time of patients is similar among these 5 policies. As in Section~\ref{sec:predefined_policies}, all policies shown here (including the RL policy) share a common random seed for each of the 365 simulated days, so this comparison is not sensitive to sampling variability for the same reason.
However, policies~3, 4 and the RL policy incur relatively high waiting times in the extreme case (when some sonographers are on leave). In Figure~\ref{fig:workload_comparison}, all five policies yield comparable sonographer quotas, suggesting that under each policy their daily quota remains manageable, and can be completed within the same day.

\subsection{Policy comparison: state-by-state analysis}

\begin{figure}[ht!]
    \centering
    \begin{subfigure}[t]{0.48\textwidth}
        \centering
        \includegraphics[width=\textwidth]{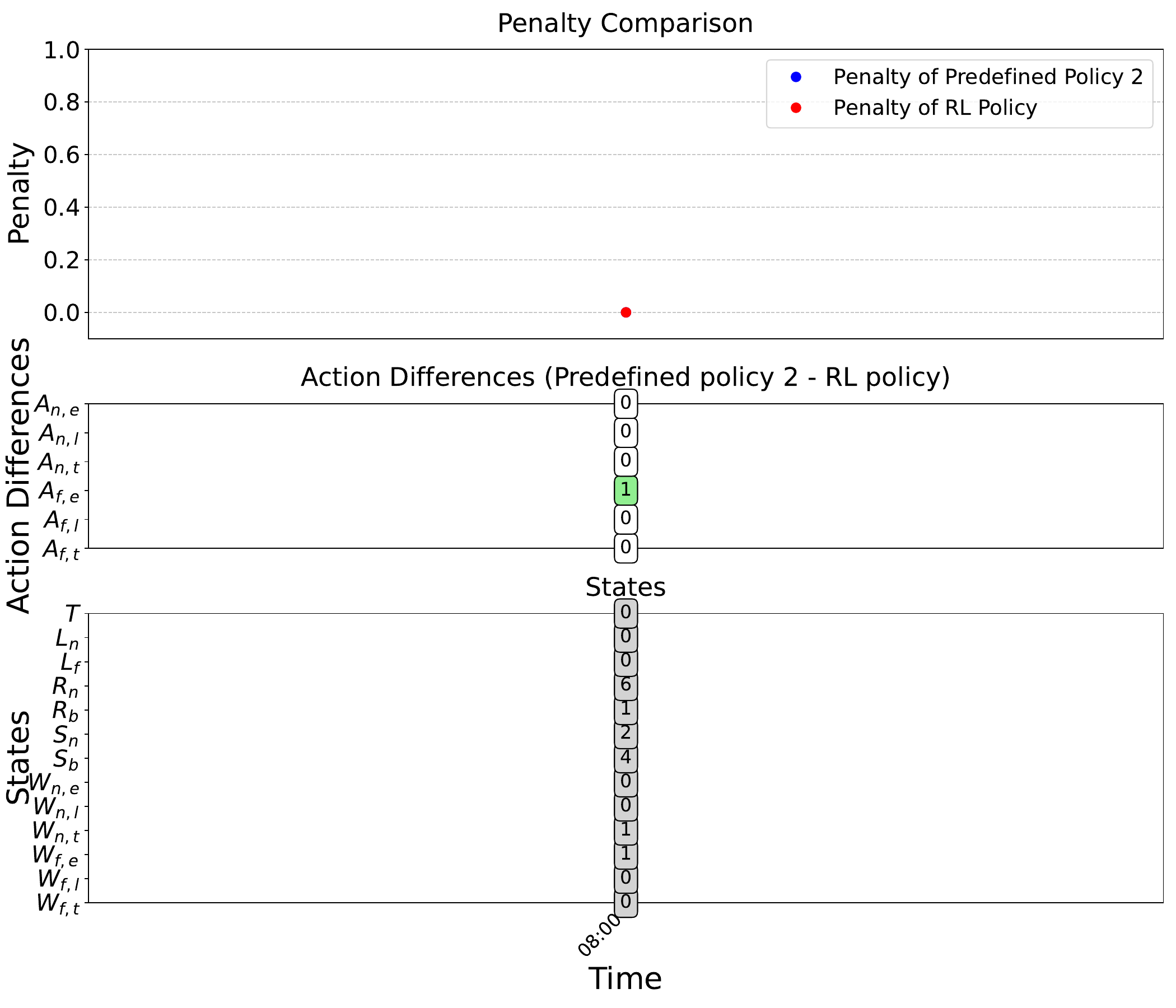}
        \caption{}
        \label{fig:state_by_state_comparison_1}
    \end{subfigure}
    \hfill
    \begin{subfigure}[t]{0.48\textwidth}
        \centering
        \includegraphics[width=\textwidth]{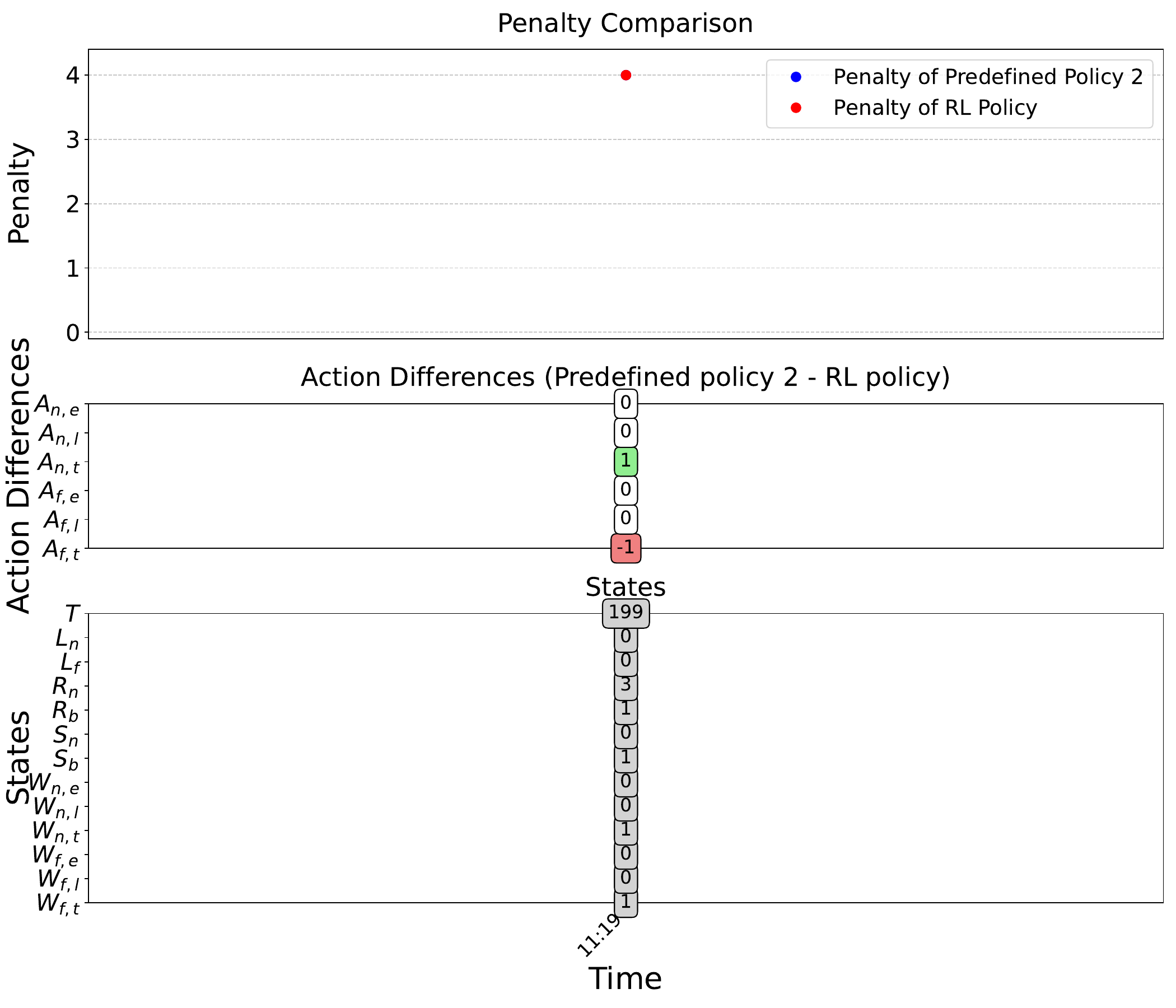}
        \caption{}
        \label{fig:state_by_state_comparison_2}
    \end{subfigure}
    \vskip\baselineskip
    \begin{subfigure}[t]{0.48\textwidth}
        \centering
        \includegraphics[width=\textwidth]{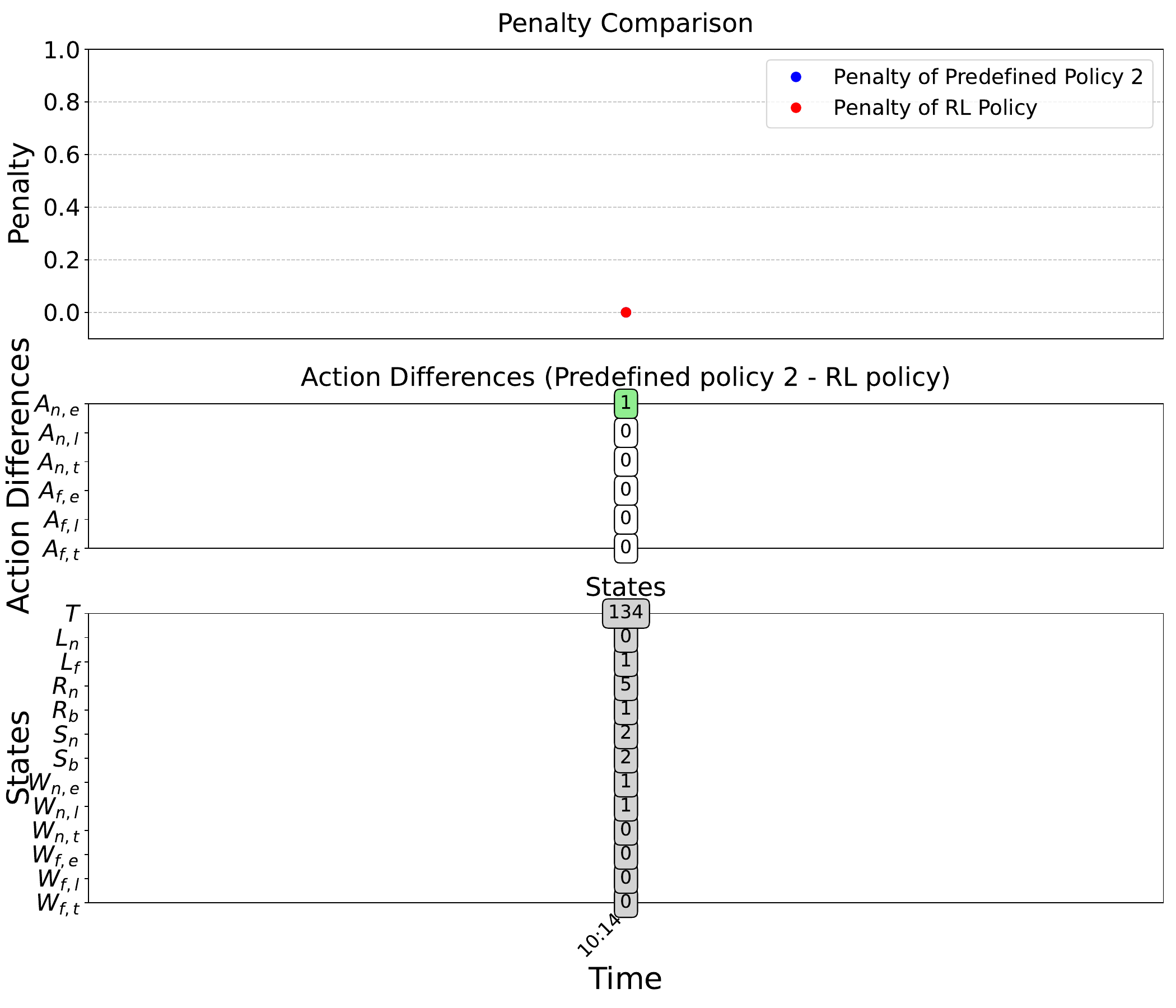}
        \caption{}
        \label{fig:state_by_state_comparison_3}
    \end{subfigure}
    \hfill
    \begin{subfigure}[t]{0.48\textwidth}
        \centering
        \includegraphics[width=\textwidth]{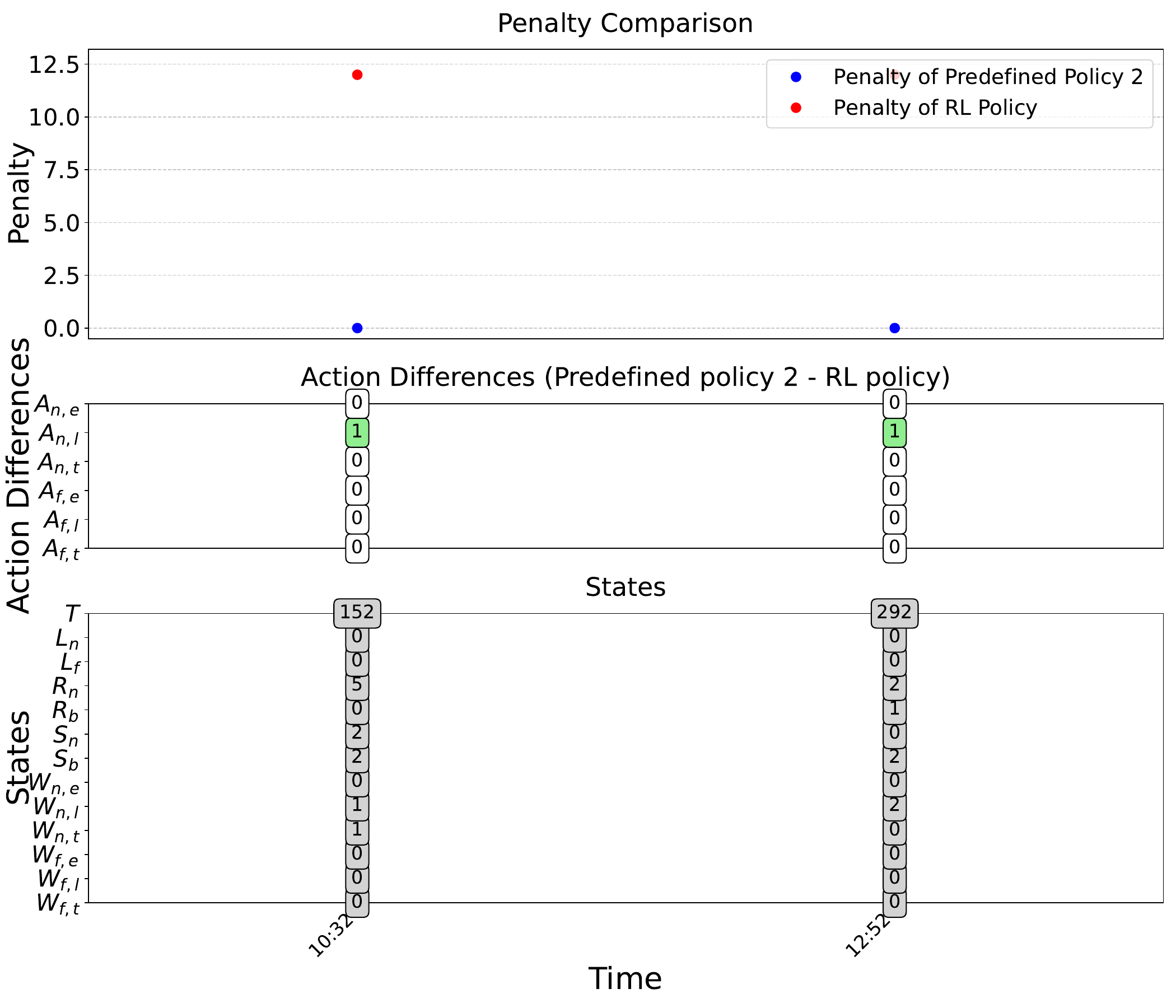}
        \caption{}
        \label{fig:state_by_state_comparison_4}
    \end{subfigure}
    \caption{
    State-by-state comparison between policy 2 and the RL policy. (a) At 08:00 AM, policy 2 accommodates non-fetal on-time patient ($A_{n,t}$) and fetal early patient ($A_{f,e}$). However, the RL policy only accommodates non-fetal on-time patient ($A_{n,t}$). (b) At 11:19 AM, policy 2 accommodates non-fetal on-time patient ($A_{n,t}$) while the RL policy accommodates the fetal on-time patient ($A_{f,t}$). (c) At 10:14 AM, policy 2 accommodates both the non-fetal late patient ($A_{n,l}$) and the non-fetal early patient ($A_{n,e}$). In contrast, the RL policy accommodates the non-fetal late patient ($A_{n,l}$) only. (d) At 10:32 AM, policy 2 accommodates the non-fetal late patient ($A_{n,l}$) and the non-fetal on-time patient ($A_{n,t}$); while the RL policy only accommodates the non-fetal on-time patient ($A_{n,t}$). At 12:52 PM, policy 2 accommodates both non-fetal patients that arrived late ($A_{n,l}$), while the RL policy only accommodates one.}
    \label{fig:state_comparison_2x2}
\end{figure}

To further compare the predefined on-the-fly policies with the RL policy, we examine the sequences of actions and the corresponding penalties generated under a fixed sequence of states.
Specifically, we first generate a sequence of states using policy 2 and then apply the RL policy to this same sequence, comparing both the selected actions and the resulting penalties.
This approach enables a direct comparison of the decision-making behavior of the two policies under identical environmental conditions.
To better visualize and interpret the differences, we present a plot composed of three panels. The bottom panel shows the states in which the RL policy and policy 2 exhibit different behaviors.
The central panel displays the difference of the actions using predefined policy 2 minus the RL policy. We use green if the difference of action is positive and red if the difference of action is negative.
The top panel illustrates the immediate penalties incurred by both policies at each time step.
By holding the state transitions constant, we compare the two policies' one-step action choices under policy 2's own state distribution.
We emphasize that this analysis is not intended as a full off-policy evaluation of the RL policy's performance; that evaluation is already provided by the 365-replication comparison reported earlier in this section. Policy 2's realized state sequence is used here only as a fixed set of concrete, representative states on which to visualize how the two policies' one-step decisions and immediate penalties differ.
Notably, the two policies behave identically on most days, with differences emerging only in a few isolated time slots.
On many days, there is no difference between these two policies throughout the whole day ($>80\%$ of the simulated days). We highlight four representative days where the policies diverge, as illustrated in Figure~\ref{fig:state_by_state_comparison_1}, Figures~\ref{fig:state_by_state_comparison_2}, ~\ref{fig:state_by_state_comparison_3}, and~\ref{fig:state_by_state_comparison_4}.
From these plots we draw four conclusions.
First, the RL policy has higher immediate penalty than policy 2, as the latter tends to use all the available echo resources to accommodate the waiting patients.
Second, fetal patients have higher priority than non-fetal patients when there is only fetal resources available, as fetal resources are relatively scarce with respect to non-fetal resources. Prioritizing fetal patients under such conditions allows for smoother and more efficient resource allocation.
Third, the RL policy has the tendency not to accommodate patients arriving early, as their waiting times are only partially considered in our penalty design.
Finally, the RL policy tends to be conservative in accommodating late-arriving patients during rush hours (10:30 AM to 3:30 PM), when echo resources are highly constrained.
Serving late patients at these times may delay subsequent on-time arrivals, leading to higher penalties, so the policy learns to avoid doing so.

\subsection{Fetal/non-fetal trade-off analysis}
\begin{figure}[ht!]
    \centering    
    \includegraphics[width=0.48\textwidth]{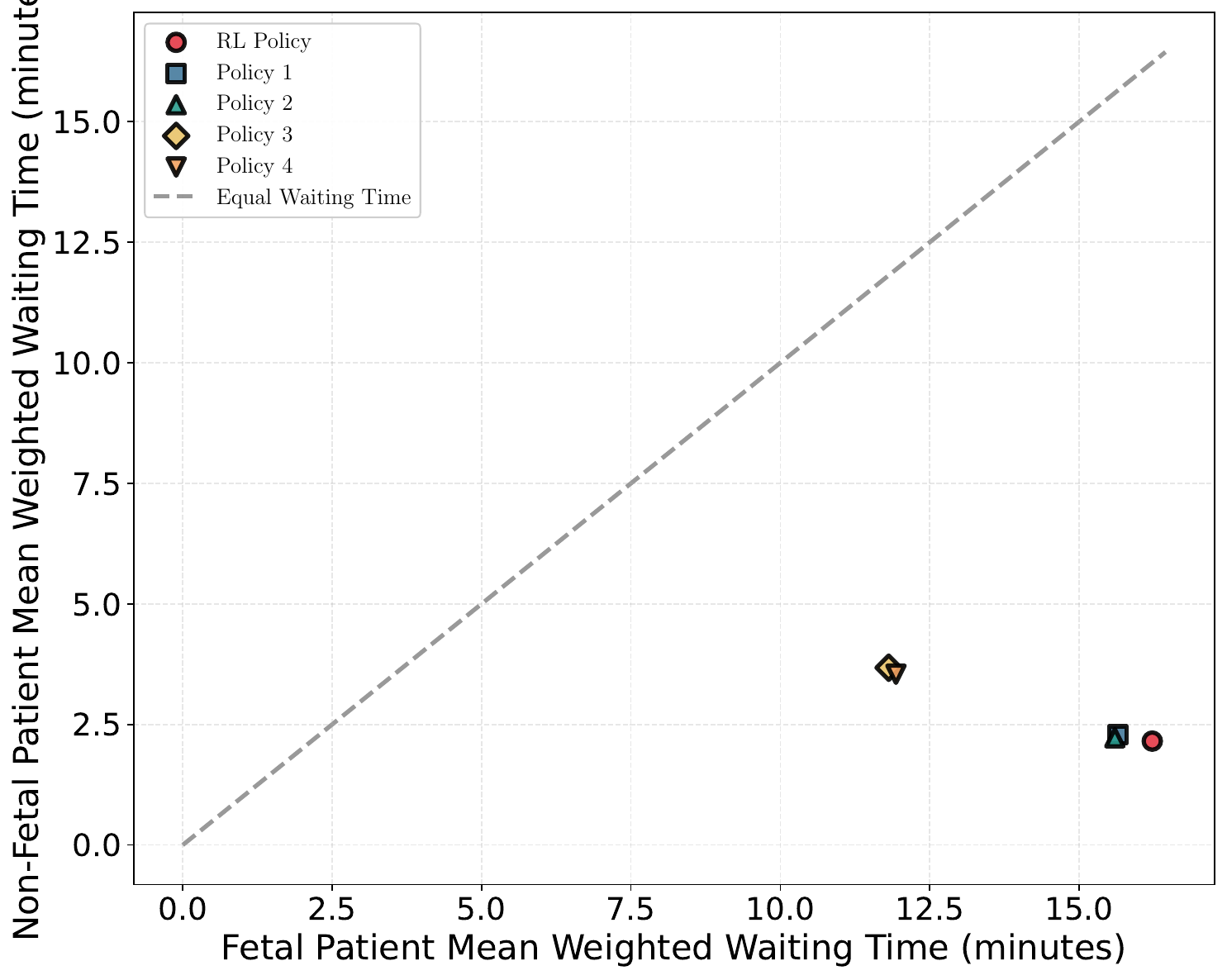}
    \caption{Fetal versus non-fetal mean weighted waiting time under each policy, averaged over 365 independent replications of a representative day. The dashed line marks equal waiting times.}
    \label{fig:boundary}
\end{figure}

%
In this section, we compare policies~1--4 and the RL policy by evaluating $W$ separately for the two patient groups in our dataset.
Specifically, we plot the waiting time for fetal patients on the $x$-axis and for non-fetal patients on the $y$-axis (matching Figure~\ref{fig:boundary}). Each point in the figure represents the mean waiting time over 365 days, allowing us to visualize the trade-offs achieved by different policies.
From the plot, we observe that although the overall $W$ is comparable among policies~1, 2, 3, 4, and the RL policy, the balance they strike between the two patient types differs.
Policies~3 and 4 achieve shorter waiting times for fetal patients but at the cost of noticeably longer waiting times for non-fetal patients.
Conversely, policies~1 and 2 produce shorter non-fetal waiting times while imposing longer waits on fetal patients.
The RL policy exhibits behavior similar to policies~1 and~2, with slightly longer waiting time for non-fetal patients and shorter waiting time for fetal patients.
Additionally, our results show that even when dedicating one fetal room and one fetal sonographer exclusively to fetal cases -- as in policies~3 and 4 -- fetal patients still tend to experience longer waiting times than non-fetal patients.
This suggests that fetal demand remains sufficiently high, and that additional fetal-specific capacity alone does not eliminate the imbalance in service delays.

\section{Sensitivity analysis}\label{sec:sensitivity}

In this section, we conduct a series of sensitivity analyses in which key environmental parameters are systematically varied to assess their impact on average waiting times for fetal versus non-fetal patients and for on-time versus late arrivals.

\subsection{No show rate}

\begin{figure}[ht!]
\centering
\begin{subfigure}[b]{0.48\textwidth}
\centering
\includegraphics[width=\textwidth]{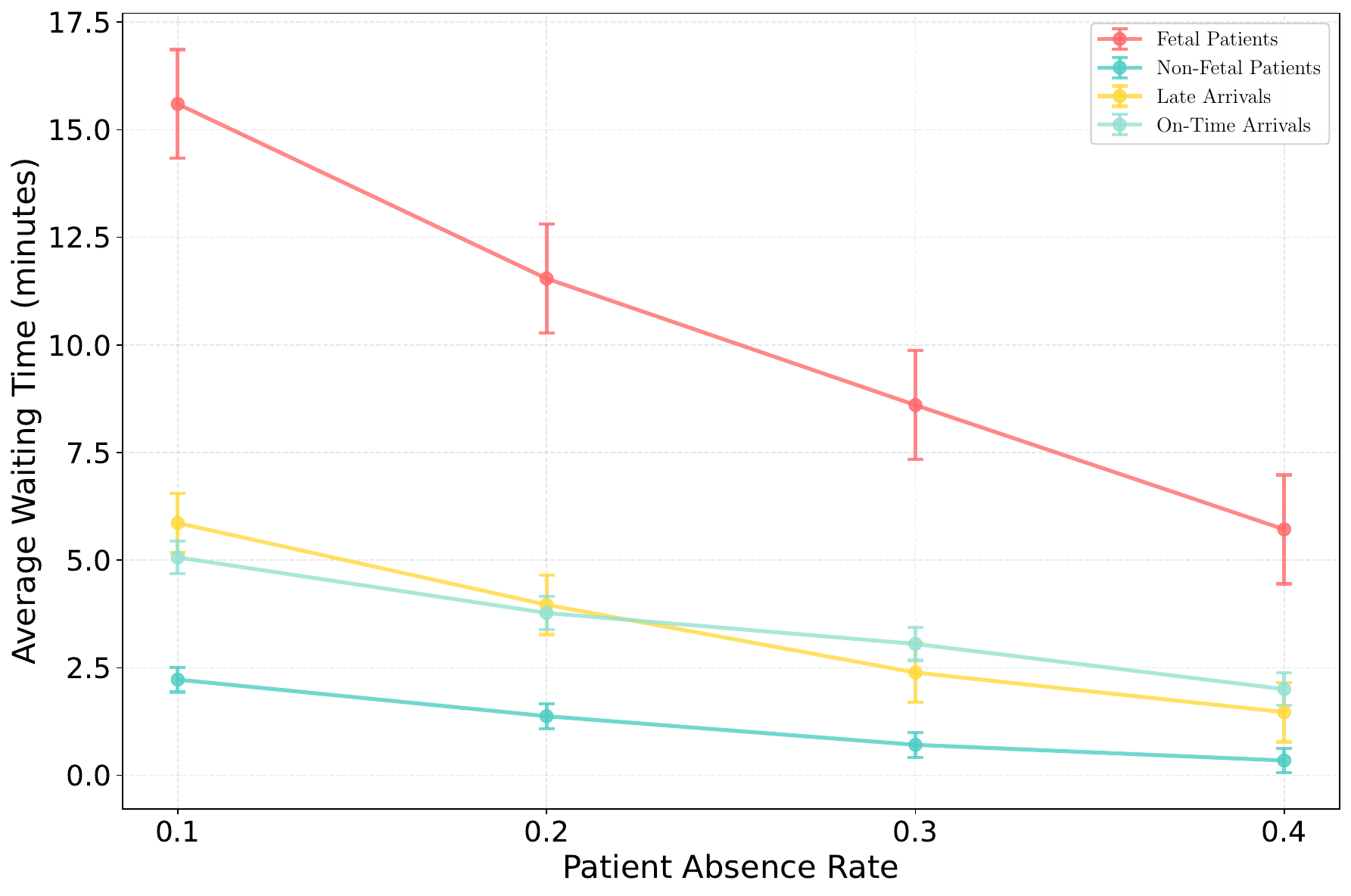}
\caption{Average weighted waiting time versus no-show rate under policy 2.}
\label{fig:sensitivity_analysis_no_show_P2}
\end{subfigure}
\hfill
\begin{subfigure}[b]{0.48\textwidth}
\centering
\includegraphics[width=\textwidth]{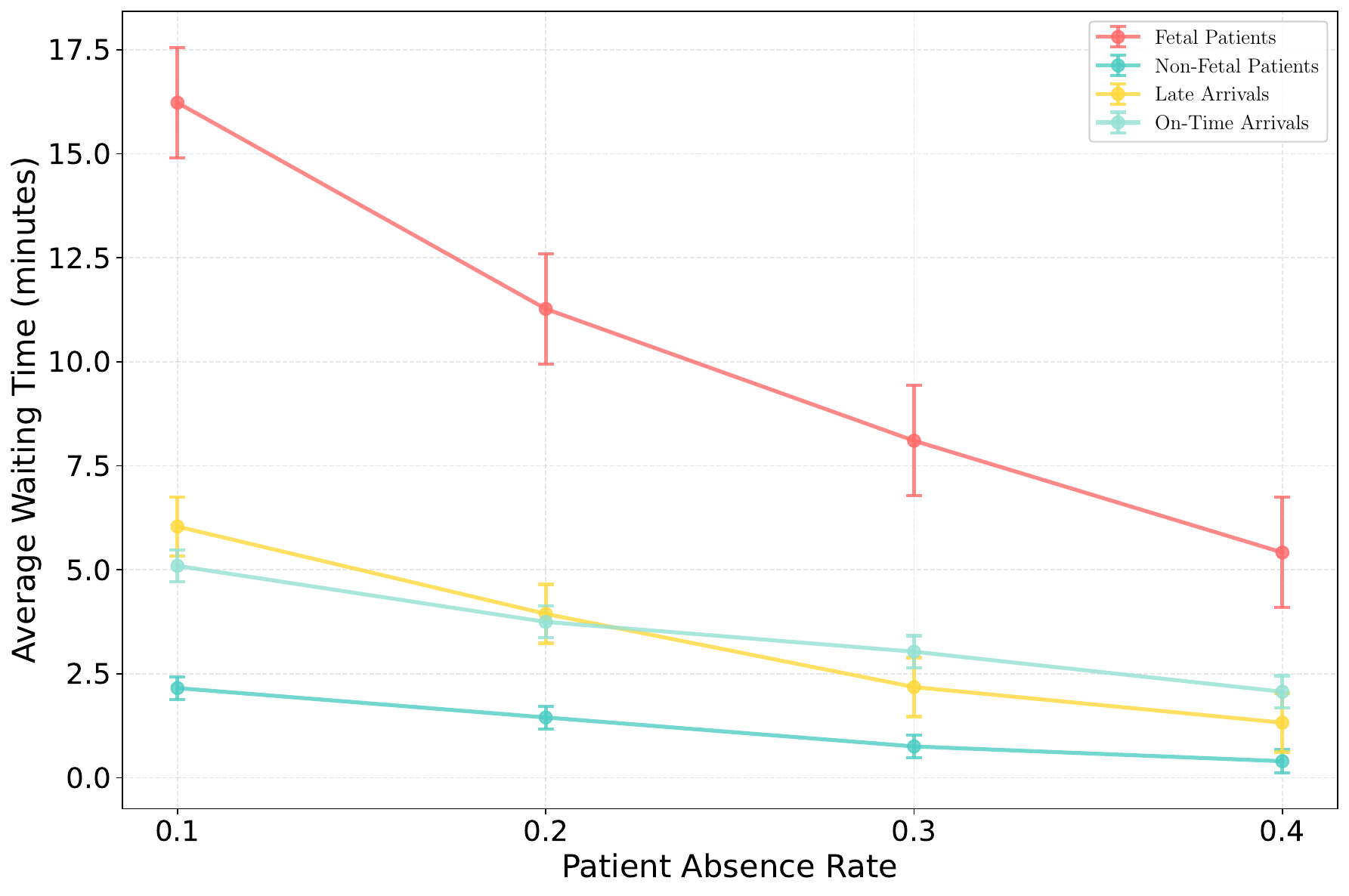}
\caption{Average weighted waiting time versus no-show rate under the RL policy.}
\label{fig:sensitivity_analysis_no_show_RL}
\end{subfigure}
\caption{Sensitivity of waiting times to varying no-show rates for both fetal and non-fetal patients.}
\label{fig:sensitivity_no_show}
\end{figure}

We varied the patient no-show rate from 0.1 to 0.4 to assess its impact on average waiting times and to evaluate the robustness of the learning algorithm. For both policy 2 and the RL-based policy, results are reported with their empirical 95\% confidence intervals (computed over the 365 replications at each setting). No-shows are modeled as i.i.d.\ Bernoulli draws at a fixed rate, independent of lead time or prior no-show history; we state this explicitly as a simplifying modeling assumption. The RL agent is trained once at a single fixed no-show rate and evaluated without retraining as the no-show rate is varied in this section and in Sections~\ref{sec:sensitivity} and~\ref{sec:dif}; the reported robustness is therefore zero-shot generalization to a shifted environment, not a learnability study.
As illustrated in Figures~\ref{fig:sensitivity_analysis_no_show_P2} and~\ref{fig:sensitivity_analysis_no_show_RL}, the weighted average waiting time consistently decreases across all patient categories as the no-show rate increases.
This is expected: higher absence rates effectively reduce system congestion, resulting in shorter queues and thus lower waiting times for patients who do attend their scheduled exams.
In addition, the RL policy exhibits a trend similar to that of policy 2, suggesting that the learning algorithm is robust under varying no-show rates.

\subsection{Echo duration}

\begin{figure}[ht!]
\centering

\begin{subfigure}[b]{0.48\textwidth}
\centering
\includegraphics[width=\textwidth]{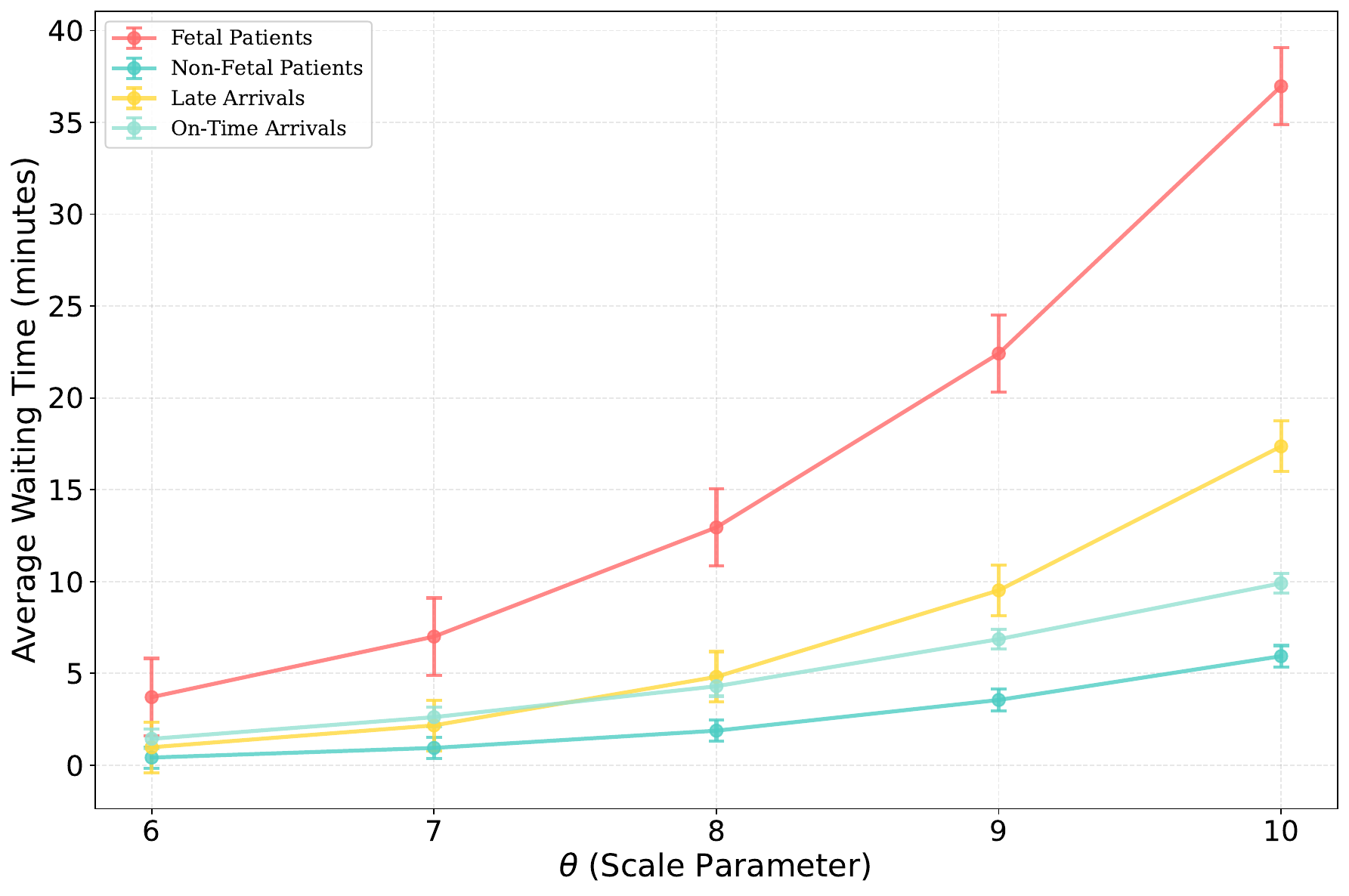}
\caption{Weighted waiting time versus $\theta$ under policy~2.}
\label{fig:sensitivity_theta_P2}
\end{subfigure}
\hfill
\begin{subfigure}[b]{0.48\textwidth}
\centering
\includegraphics[width=\textwidth]{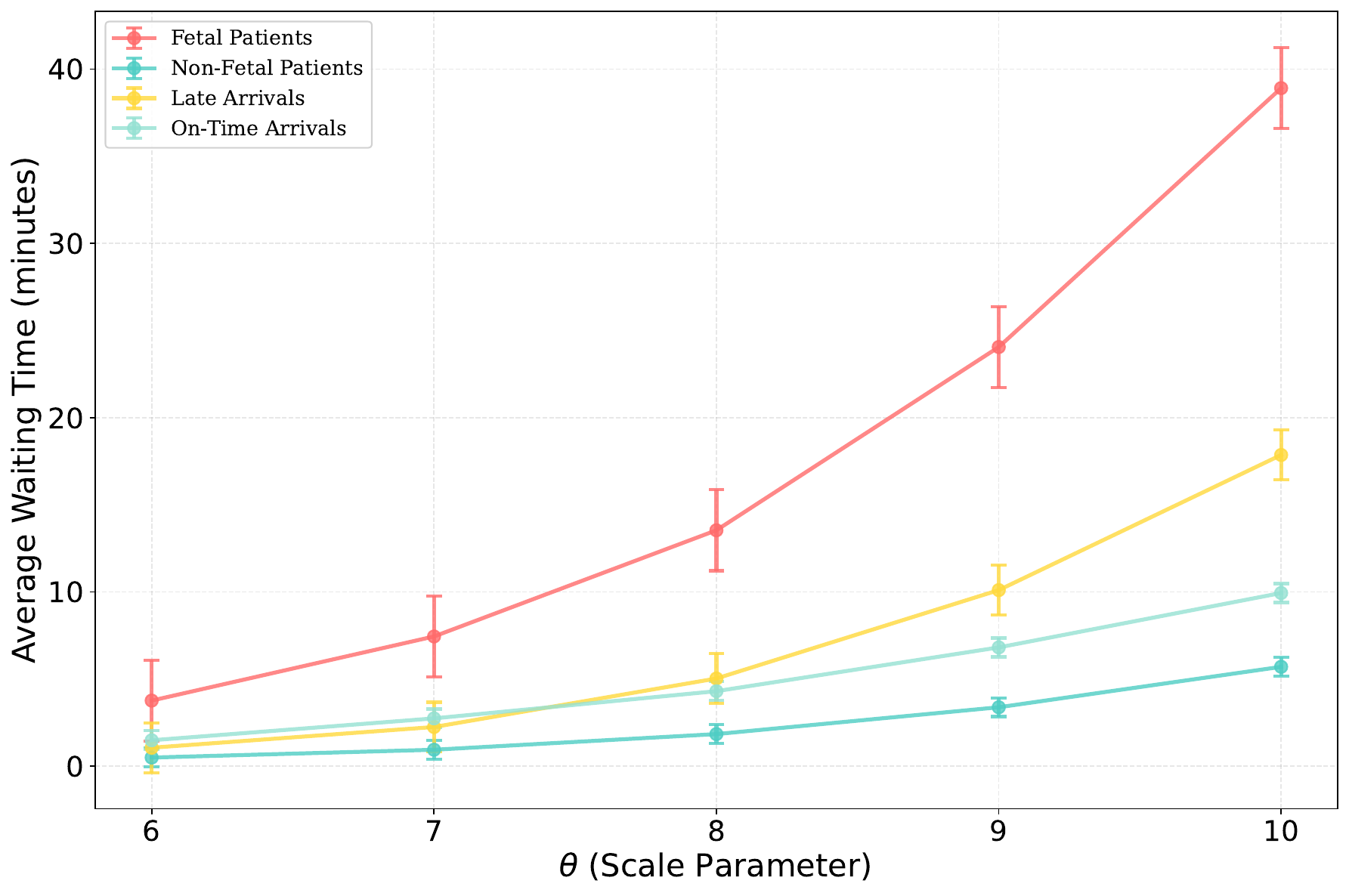}
\caption{Weighted waiting time versus $\theta$ under RL policy.}
\label{fig:sensitivity_theta_RL}
\end{subfigure}

\vspace{0.4cm}

\begin{subfigure}[b]{0.48\textwidth}
\centering
\includegraphics[width=\textwidth]{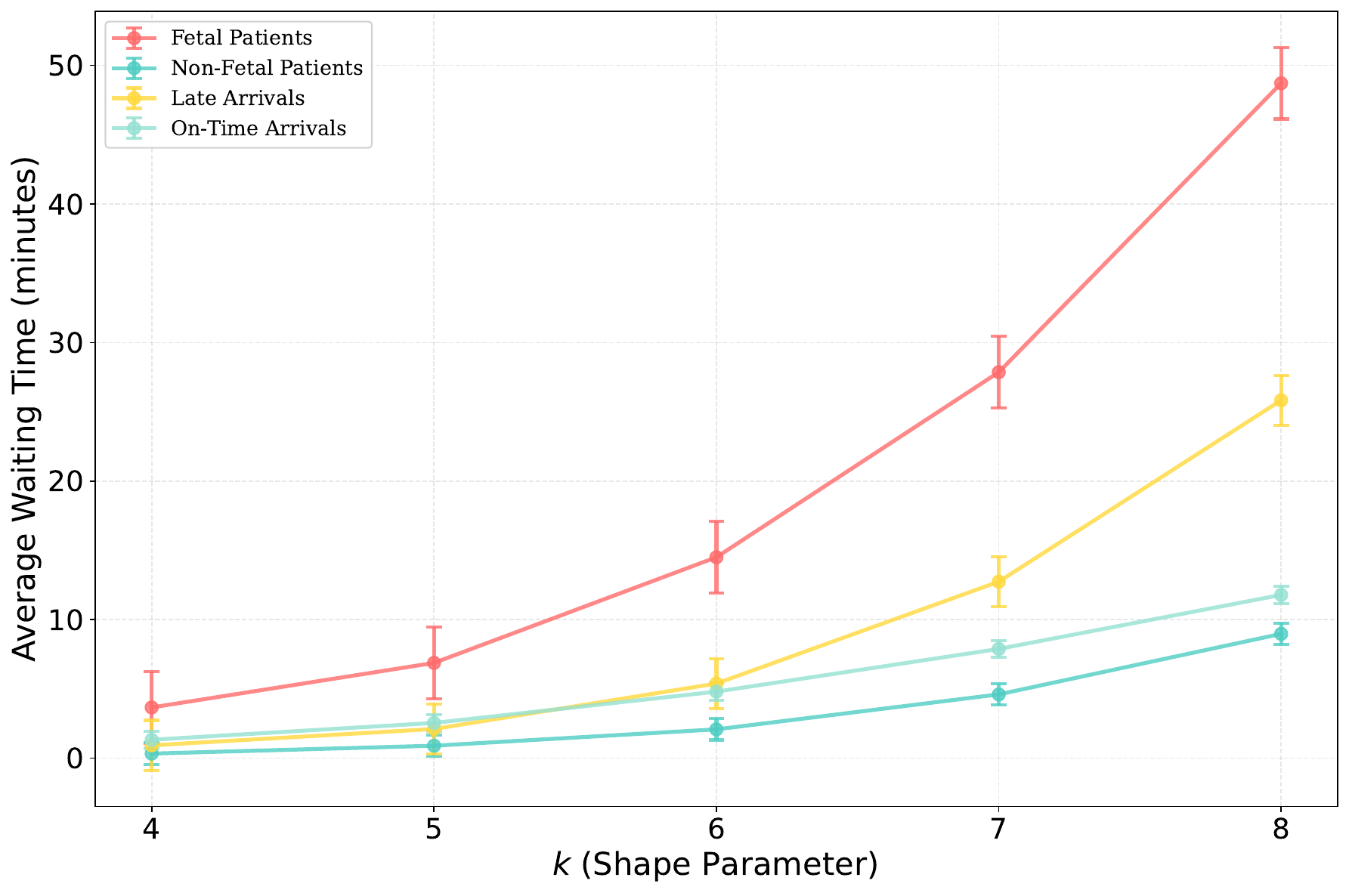}
\caption{Weighted waiting time versus $k$ under policy~2.}
\label{fig:sensitivity_k_P2}
\end{subfigure}
\hfill
\begin{subfigure}[b]{0.48\textwidth}
\centering
\includegraphics[width=\textwidth]{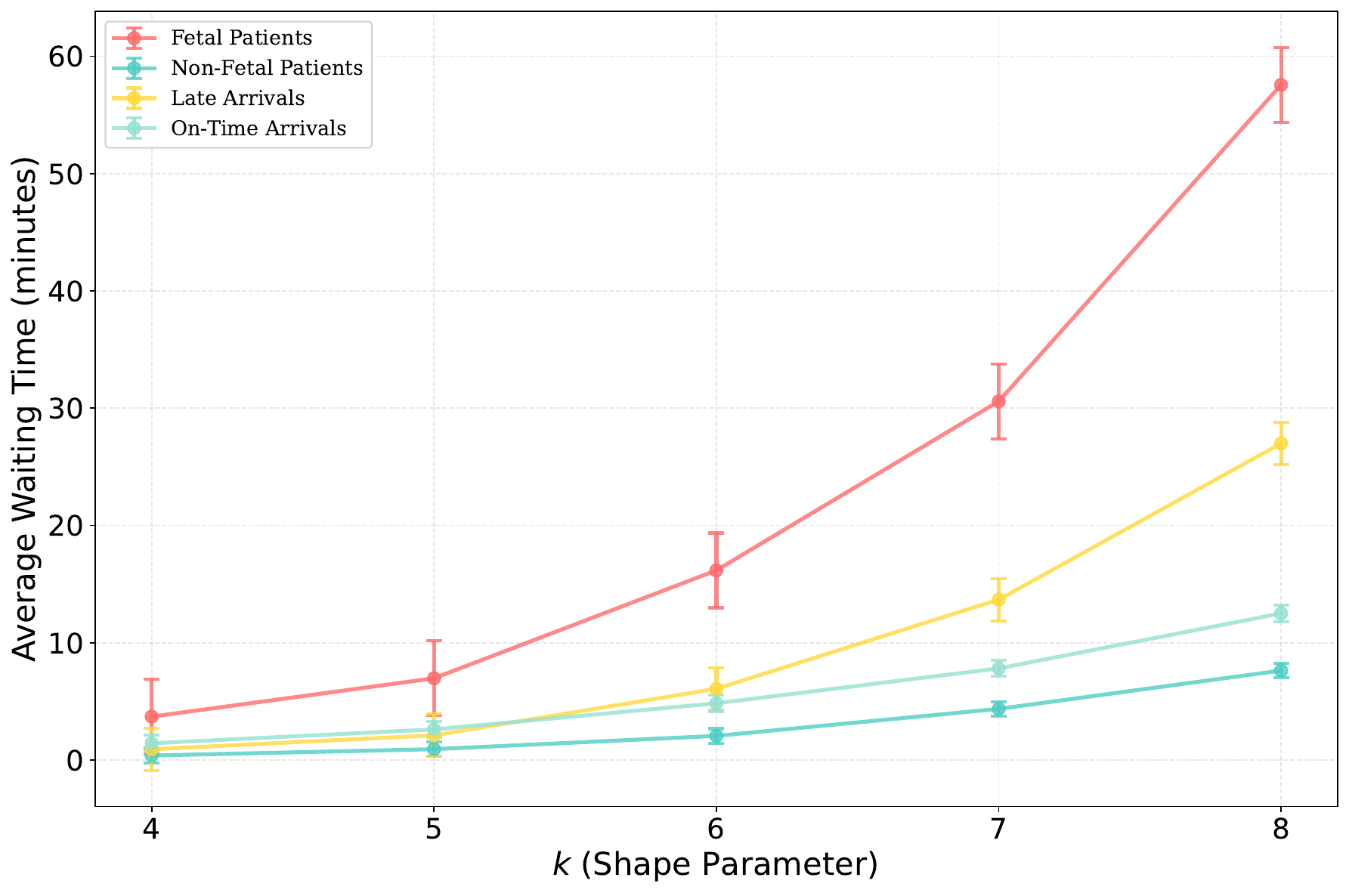}
\caption{Weighted waiting time versus $k$ under RL policy.}
\label{fig:sensitivity_k_RL}
\end{subfigure}

\caption{Sensitivity of $W$ to parameters $\theta$ and $k$ for both policy~2 and the RL policy.}
\label{fig:sensitivity_theta_k}
\end{figure}

We also examined how variations in the $k$ and $\theta$ parameters of the Gamma distribution modeling echo exam durations affect the average waiting time. The tested ranges (4--8 for $k$, 6--10 for $\theta$) bracket the fitted values ($k=6.12$, $\theta=8.30$, marked on each panel).
For both policy 2 and the RL policy, we report the results with their empirical 95\% confidence intervals.
As illustrated in Figures~\ref{fig:sensitivity_theta_P2}, ~\ref{fig:sensitivity_theta_RL}, ~\ref{fig:sensitivity_k_P2}, and ~\ref{fig:sensitivity_k_RL}, the average waiting times increase with both $k$ and $\theta$.
This is expected, as the mean of the untruncated Gamma distribution ($\mu = k\theta$) increases linearly with these parameters, leading to longer average service times and consequently longer waiting times (the simulator truncates durations to $[15,150]$ minutes, so the mean of the truncated distribution actually sampled is not exactly $k\theta$).

\subsection{Variance of patient's arrival time}

\begin{figure}[ht!]
\centering
\begin{subfigure}[b]{0.48\textwidth}
\centering
\includegraphics[width=\textwidth]{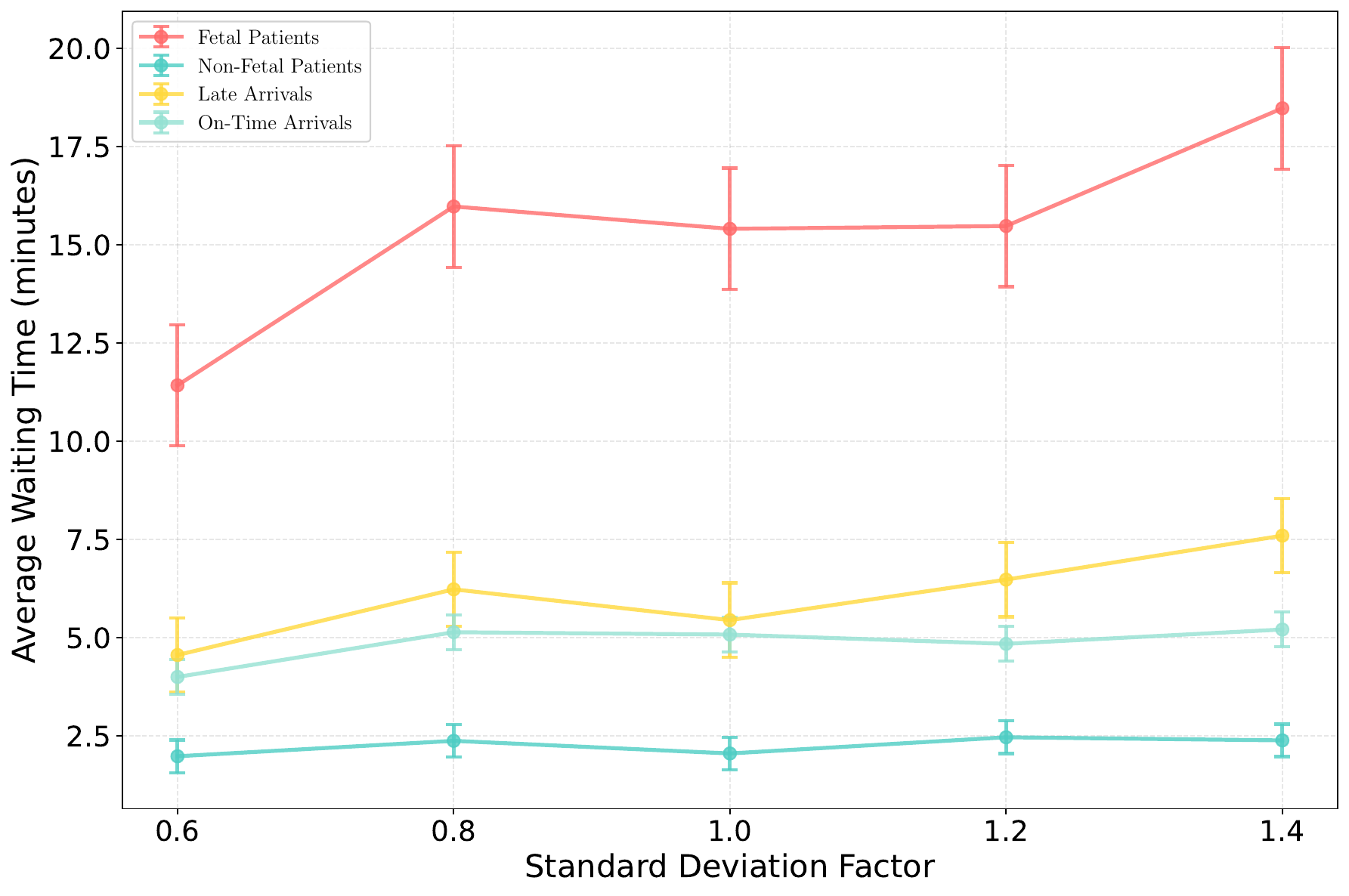}
\caption{Average patient waiting time versus varying standard deviations under policy 2.}
\label{fig:sensitivity_analysis_variance_P2}
\end{subfigure}
\hfill
\begin{subfigure}[b]{0.48\textwidth}
\centering
\includegraphics[width=\textwidth]{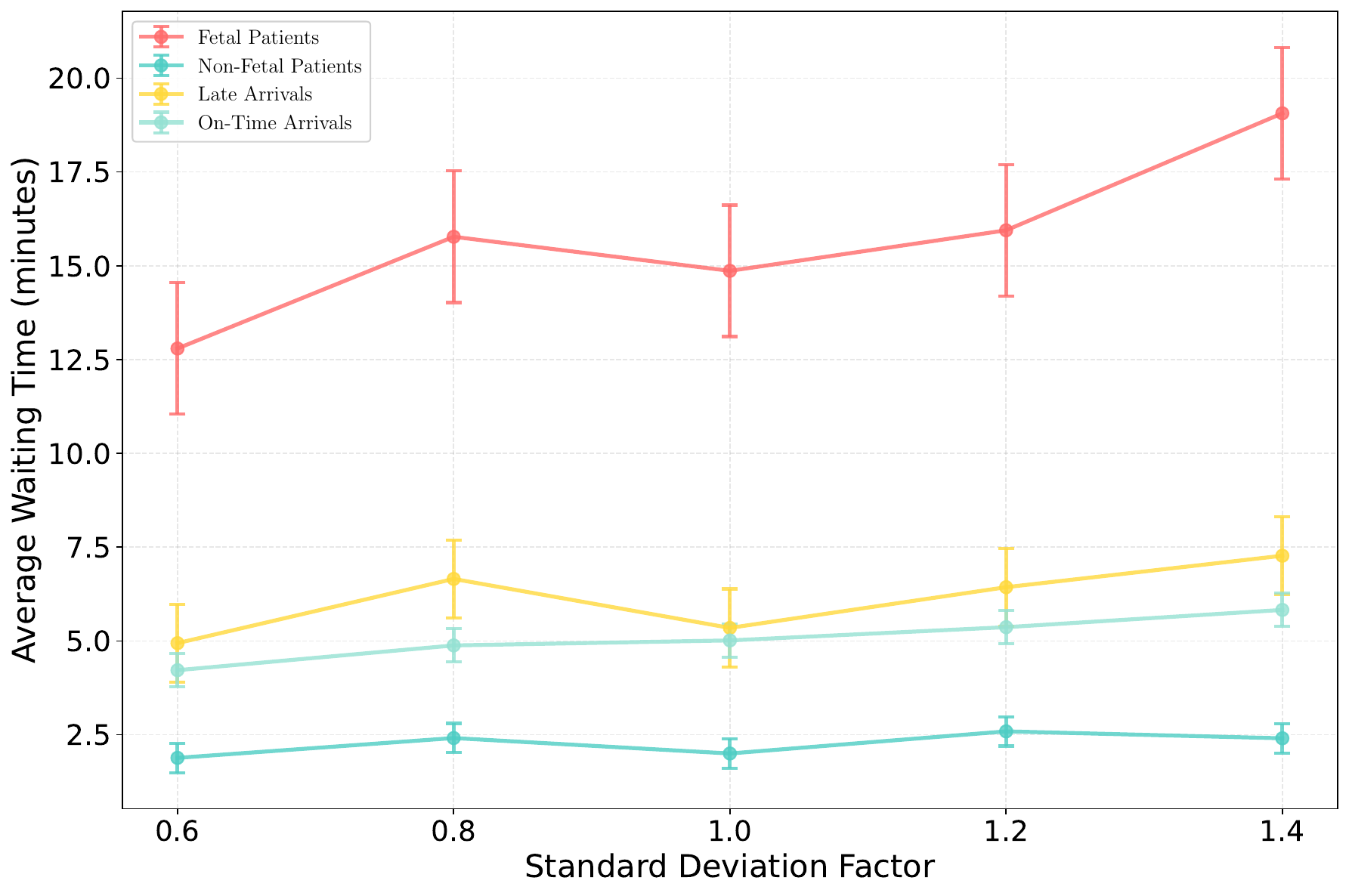}
\caption{Average patient waiting time versus varying standard deviations under the RL policy.}
\label{fig:sensitivity_analysis_variance_RL}
\end{subfigure}
\caption{Sensitivity of patient waiting times to varying standard deviations for both fetal and non-fetal populations.}
\label{fig:sensitivity_variance}
\end{figure}

We also evaluated the sensitivity of the learning algorithm to the time lag between a patient’s arrival and scheduled time by varying a multiplicative standard-deviation factor applied to the fitted arrival-delay distribution from 0.6$\times$ to 1.4$\times$.
For both policy 2 and the RL policy, we report the results with empirical 95\% confidence intervals based on 365 independent replications of a representative day.
As illustrated in Figures~\ref{fig:sensitivity_analysis_variance_P2} and~\ref{fig:sensitivity_analysis_variance_RL}, waiting times tend to increase across all patient categories as the standard deviation increases, with a similar sensitivity obtained for policy 2 and the RL policy.
Higher variance in patient arrivals introduces greater fluctuations in the system, leading to less stable operations, and more extreme scenarios, hence to an increase in $W$.

\section{Results under varying resource availability}\label{sec:dif}

We conduct a parametric study to evaluate our algorithm’s performance under environments with different echo resources.
We examine two scenarios: a case with \emph{abundant} resources and a case with \emph{scarce} resources.

\subsection{Abundant echo resources}

\begin{figure}[ht!]
\centering
\includegraphics[width=0.48\textwidth]{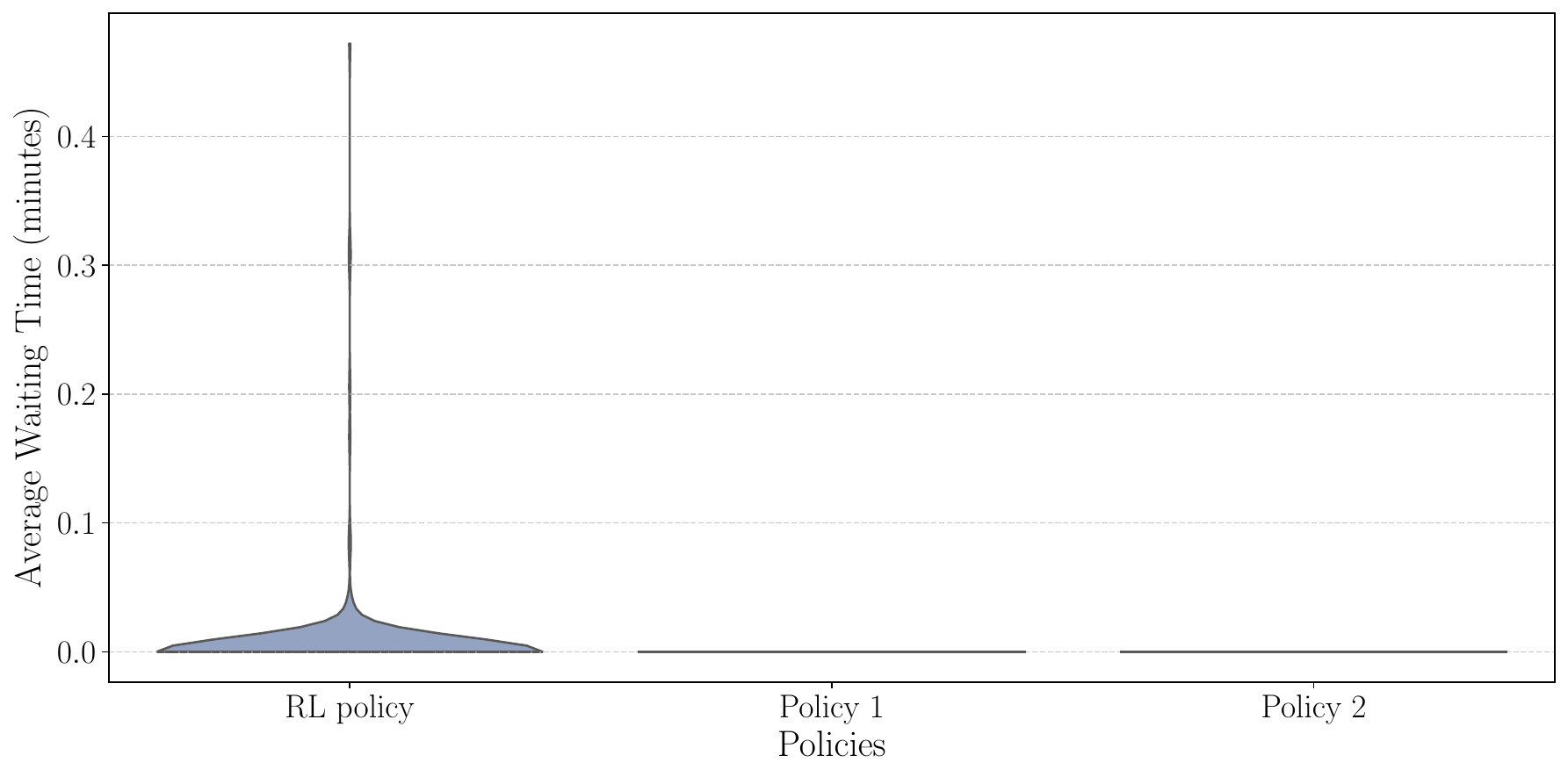}
\caption{Comparison of average daily waiting time produced by various policies under abundant echo resources.}
\label{fig:waiting_time_ab}
\end{figure}

In the abundant resource setting, the system is equipped with sufficient capacity in both echo rooms and sonographers. Since we assume unlimited echo resources are available, there is no penalty for idle resources, so the reward function optimized in this subsection differs from the one used in Section~\ref{sec:lp}; absolute penalty values are therefore not comparable across the abundant, normal, and scarce resource regimes (Figures~\ref{fig:learning_curve}, \ref{fig:abundant_learning_curve}, \ref{fig:scarce_learning_curve} have very different penalty scales). 
Moreover, since echo resources are unlimited, policies 3 and 4 are not considered.
In such cases, rule-based on-the-fly policies are always sufficient to accommodate patient demand due to the surplus of rooms and staff.
As shown in Figure~\ref{fig:abundant_learning_curve}, the learning curve for the RL policy decreases smoothly and converges to 0. Although the RL policy typically yields zero penalty on most days (similar to policies 1 and 2), it occasionally incurs a small penalty due to suboptimal decisions.
As shown in Figure~\ref{fig:waiting_time_ab}, the mean daily waiting time of the RL policy is not exactly zero over 10{,}000 replications of a representative day, indicating a small residual waiting time for some patients. 
This effect is too small to be reliably distinguished from zero at the 365 replications used throughout the rest of the paper, so we characterize it here using 10{,}000 replications instead.

\subsection{Scarce echo resources}

\begin{figure}[ht!]
\centering

\begin{minipage}[t]{0.48\textwidth}
\centering
\includegraphics[width=\textwidth]{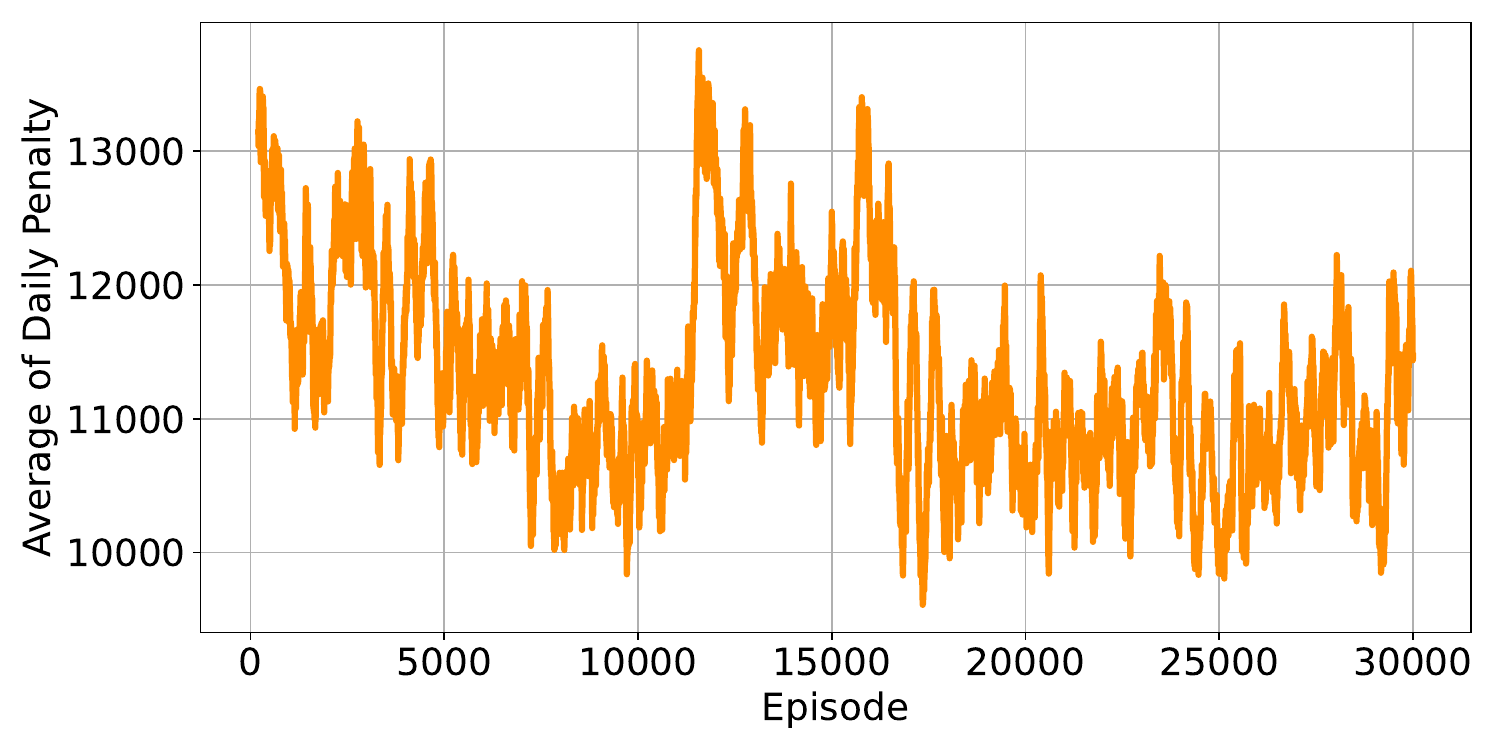}
\caption{Running average (200-day window) of daily penalty under scarce resources.}
\label{fig:scarce_learning_curve}
\end{minipage}
\hfill
\begin{minipage}[t]{0.48\textwidth}
\centering
\includegraphics[width=\textwidth]{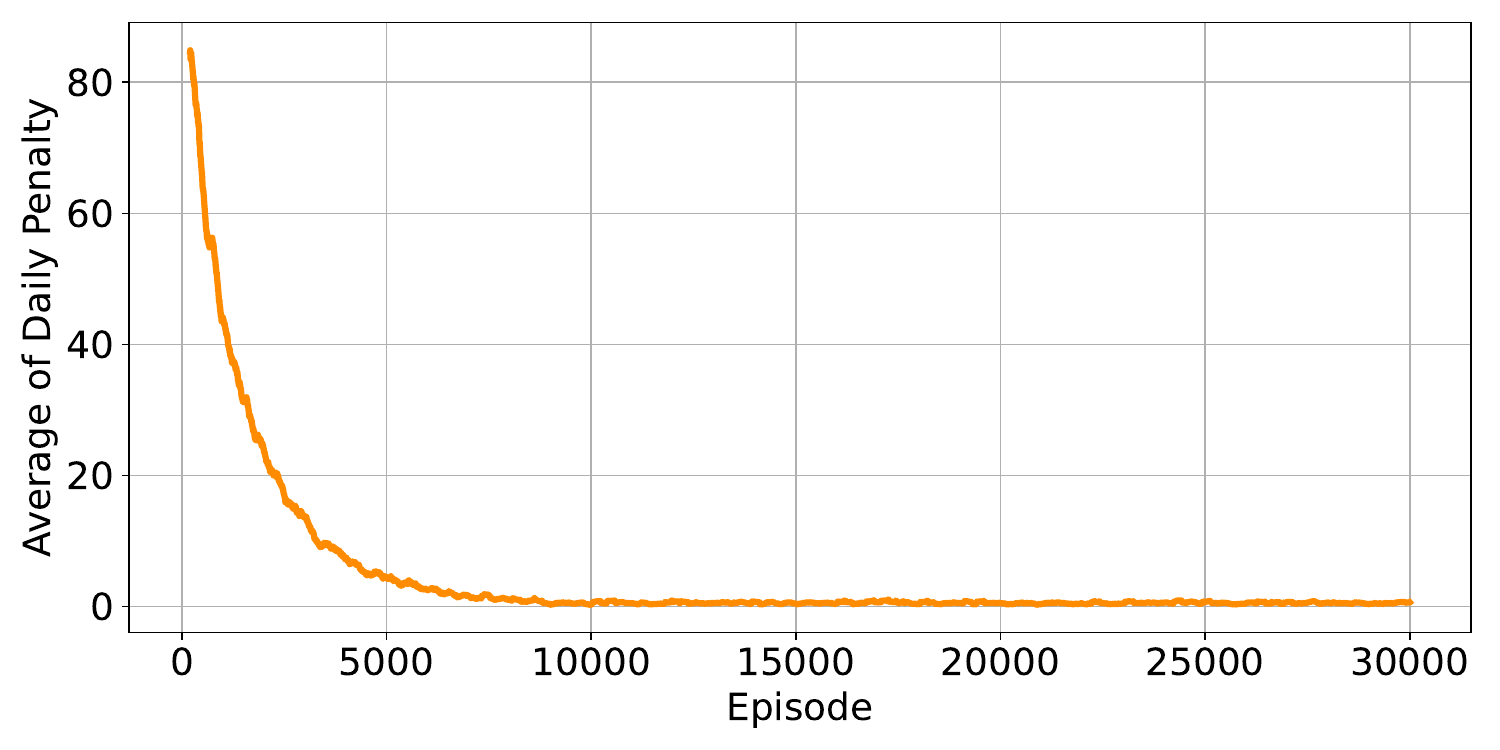}
\caption{Running average (200-day window) of daily penalty under abundant resource setting.}
\label{fig:abundant_learning_curve}
\end{minipage}

\end{figure}

\begin{figure}[ht!]
\centering

\begin{minipage}[t]{0.48\textwidth}
\centering
\includegraphics[width=\textwidth]{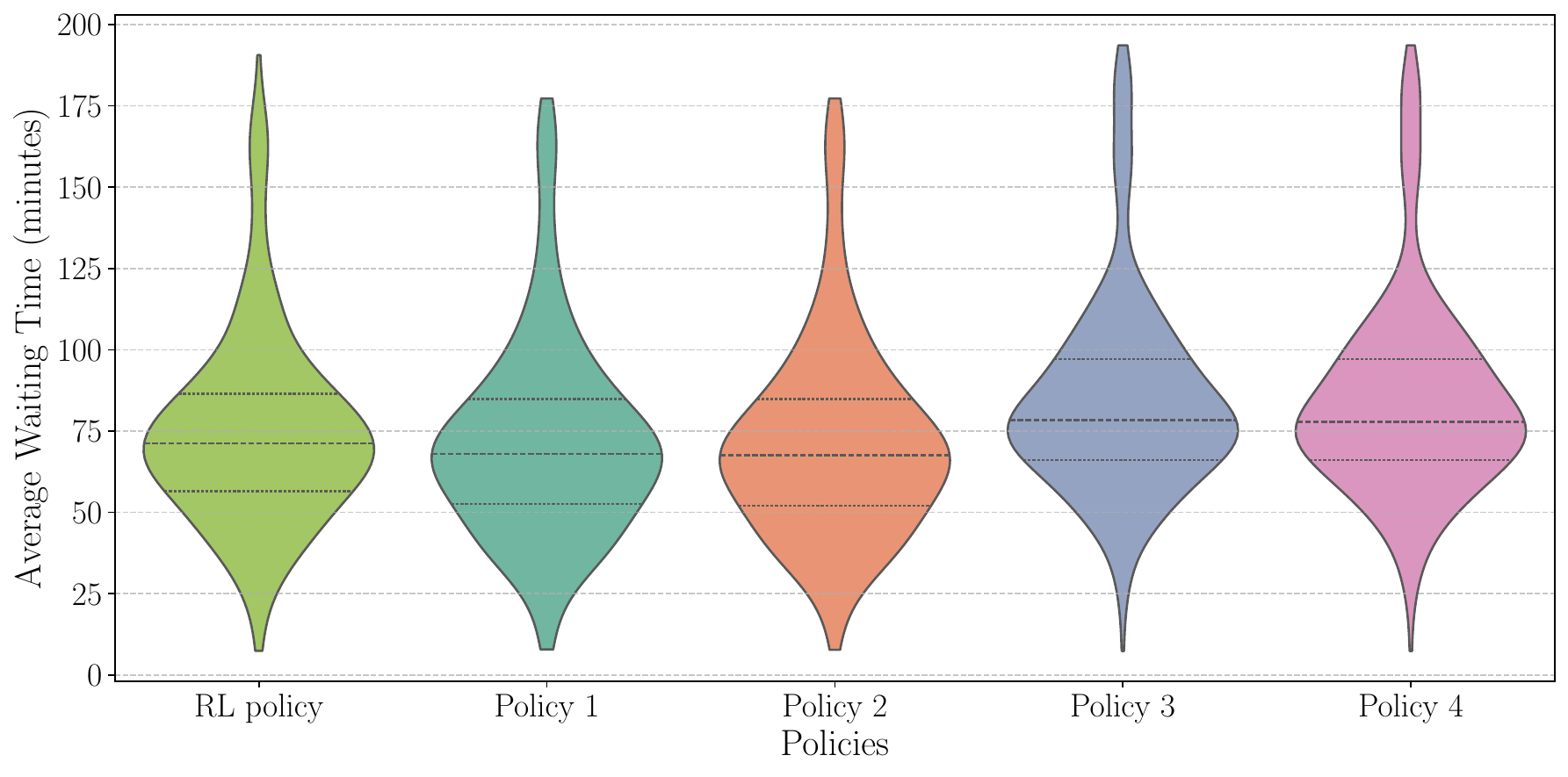}
\caption{Comparison of average daily waiting time produced by various policies under scarce resources.}
\label{fig:waiting_time_scarce}
\end{minipage}
\hfill
\begin{minipage}[t]{0.48\textwidth}
\centering
\includegraphics[width=\textwidth]{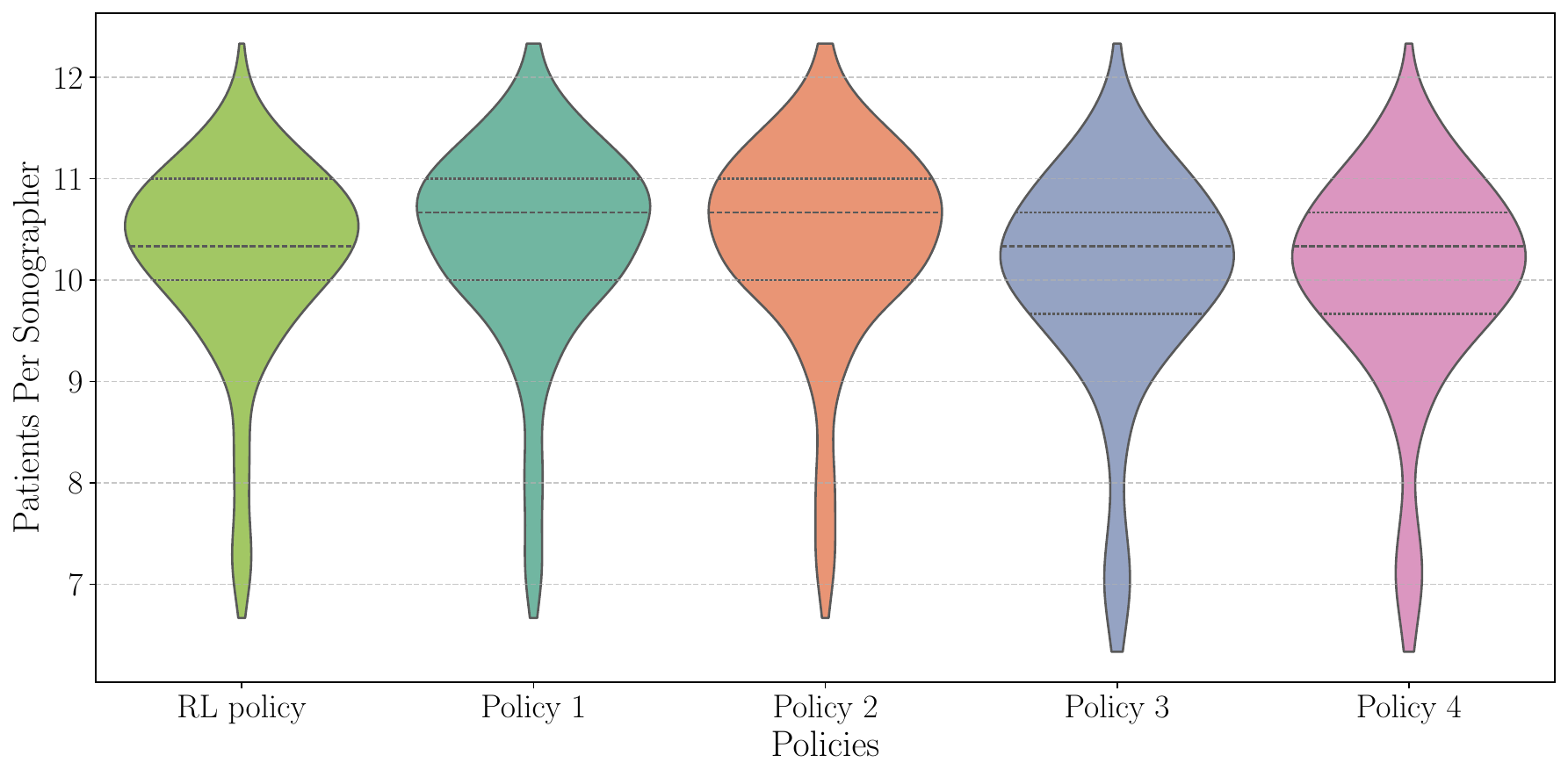}
\caption{Comparison of average daily sonographer quota produced by various policies under scarce resources.}
\label{fig:workload_scarce}
\end{minipage}

\end{figure}

In the scarce resource setting, the system operates under significant constraints in both room and sonographer availability.
Specifically, there is only one echo room capable of performing both fetal and non-fetal examinations, alongside three rooms designated exclusively for non-fetal procedures.
The sonographer workforce comprises two sonographers qualified for both fetal and non-fetal assessments, and one additional sonographer limited to non-fetal cases.
As shown in Figure~\ref{fig:scarce_learning_curve}, the RL learning curve has larger fluctuations than in the normal and abundant settings.
This outcome indicates that, under severe resource constraints, the current RL algorithm struggles to learn an effective allocation policy.
As shown in Figures~\ref{fig:workload_scarce} and~\ref{fig:waiting_time_scarce}, policies 1 and 2 result in the highest sonographer quotas and the lowest average waiting times. In contrast, policies 3 and 4 yield the lowest sonographer quotas but the highest average waiting times. The RL policy lies in between.

This behavior can be explained by the scarcity of echocardiography resources. Policies 3 and 4 reserve one sonographer and one room exclusively for fetal patients, which significantly restricts resource allocation flexibility. Under such constrained conditions, overall system efficiency decreases, leading to longer patient waiting times.
Moreover, the results suggest that in a resource-scarce setting, the RL policy faces greater difficulty in learning an optimal allocation strategy, resulting in intermediate performance.
It suggests the need to explore alternative learning frameworks or incorporate domain-specific constraints to improve performance in this setting.

\section{Discussion}\label{sec:discussion}

This study compared rule-based and reinforcement-learning resource allocation policies for echocardiography (echo) tests.
After evaluating several rule-based policies, we adopted an \textit{on-the-fly} allocation scheme.
Under this scheme, we trained a reinforcement learning (RL) agent to learn an allocation policy directly through interaction with the environment.
Our results suggest that although the RL-based policy exhibits distinct behavioral characteristics—most notably, a more conservative decision-making tendency in anticipation of environmental uncertainty compared with policies 1 and 2—it achieves performance comparable to, but not exceeding, the other rule-based \textit{on-the-fly} policies, and is worse on end-of-day overruns (Section~\ref{sec:lp}). 
In different environments, the RL policy could achieve similar performance to policy 2, which shows the robustness of our learning algorithm under distribution shift; as confirmed in Section~\ref{sec:sensitivity}, this reflects zero-shot generalization to a shifted environment rather than retraining at each setting.

We observed clear behavioral differences among the \textit{on-the-fly} policies, which can be ranked by their level of operational aggressiveness toward fetal patients.
Policies 1 and 2 are the most aggressive, as they do not reserve dedicated fetal resources for fetal patients.
By contrast, policies 3 and 4 are the most conservative, as they always reserve fetal resources exclusively for fetal patients.
The RL policy adopts a more balanced strategy, prioritizing fetal patients when both fetal and non-fetal patients are waiting, while still allowing flexible utilization of shared resources.
Despite these behavioral differences, all policies exhibit similar performance in terms of the average patient waiting time, demonstrating the agent’s ability to learn a competitive and adaptive allocation policy without explicit rule-based design.
Nevertheless, the RL policy still exhibits several limitations.
It cannot consistently allocate patients in a manner that minimizes the completion time of all tests, which makes it not optimal among all scenarios. For example, through state-by-state policy comparisons, we identified limitations in the RL agent’s decision-making. 
In Figure~\ref{fig:state_by_state_comparison_1}, early in the day, it is generally beneficial to serve early-arriving patients since resource usage at this stage has minimal impact on later arrivals. However, the agent does not consistently learn this behavior. This oversight is likely due to the current design of the penalty function and the limited training duration, both of which may prevent the agent from capturing such operational nuances.

More broadly, the ability of the RL policy to consistently outperform rule-based strategies is constrained by several factors.
First, due to the stochastic nature of the environment, it is inherently impossible to identify a single policy that outperforms all alternatives across every possible scenario.
Second, the current state representation omits several key features that characterize real-world hospital operations, effectively rendering the problem a partially observable Markov decision process (POMDP). We note that the current time step is already part of the state, and the underlying daily appointment schedule is deterministic and identical across simulated days; only the realized arrival time of each patient, drawn from an empirical noise distribution around its scheduled slot, varies from day to day.
In principle, the value network could therefore learn how many scheduled patients remain at a given time of day purely through repeated exposure to this fixed template during training, without an explicit agenda feature; what current time alone cannot convey is each day's specific realization of arrival noise and absences. Whether the network's capacity and training budget are sufficient to extract this time-dependent regularity in practice, however, is untested, and explicitly providing it -- either as a remaining-schedule feature or through a recurrent architecture (DRQN) \citep{hausknecht2015drqn} -- may still ease learning even where the information is, in principle, already recoverable from the state.
Important contextual variables—such as individual patient waiting times, the timing and duration of sonographer breaks, and daily staffing levels—are not explicitly modeled.
While incorporating these variables could improve policy performance, doing so would substantially increase the dimensionality of the state space and the computational burden of training.

Third, the complexity of the environment, combined with the high-dimensional state and action spaces, makes it challenging for the agent to sufficiently explore and experience all possible scenarios during training.
Limited exposure to rare or infrequently encountered states can hinder the agent’s ability to generalize effectively.
Moreover, the large dimensionality of the problem increases the risk of the training process becoming trapped in local optima, making it difficult for the agent to discover truly optimal policies.

Finally, due to the long decision horizon ($t$, see Section~\ref{sec:mdp}), penalties incurred at the end of an episode have limited influence on earlier decisions under a discounted reward formulation ($\gamma=0.99$ over an episode of several hundred one-minute steps gives an effective horizon well under two hours).
This temporal credit assignment challenge further complicates the learning of globally optimal long-term behaviors.

All RL results reported here come from a single trained model, i.e., a single training seed; the 365 independent daily replications used throughout this paper account for environment-level evaluation noise -- how this one trained policy performs across different arrival patterns -- but do not address whether retraining with a different seed would yield a policy of meaningfully different quality, a well-documented risk for deep RL~\citep{henderson2018deep}. 
Properly ruling this out would require retraining the full pipeline across several seeds and reporting the resulting performance distribution~\citep{agarwal2021deep}; given the computational cost of a single training run, we did not do so within the scope of this revision. We therefore report seed-to-seed variability as an open limitation.

\subsection{Limitations}

Beyond the algorithmic limitations discussed above, several limitations bear on the external validity and completeness of the study as a whole:
\begin{itemize}
    \item \textbf{Single center, single week.} All parameters are estimated from one week of hand-recorded data at one quaternary pediatric echo lab; a single fixed daily appointment template is used throughout.
    \item \textbf{Outpatient only.} The simulation does not model inpatient, emergency, or add-on cases, which compete for the same rooms and sonographers in practice.
    \item \textbf{Unused operational data.} The source dataset also records vitals, EKG, and clinic-visit timestamps (Section~\ref{sec:data}); the ripple-effect motivation given in the Introduction is not itself modeled, since the simulator only tracks the echo step.
    \item \textbf{Simulator not validated against observed operations}, and \textbf{no statistical comparison between policies} (both noted above, Comments \#2--\#3).
    \item \textbf{Single training seed} for all reported RL results (see Comments \#26 in the Discussion); seed-to-seed variability has not been characterized.
\end{itemize}

\subsection{Future work}

To improve the effectiveness and robustness of the RL-based policy, future work should focus on the following areas:

\begin{enumerate}
    \item \textbf{Improving the agent’s understanding of the environment}: Instead of relying entirely on exploration, we propose warm-starting the agent with domain knowledge, e.g., the patient scheduling agenda and the distribution of testing durations, which could help the agent learn more efficiently and reduce the complexity of the training process. Additionally, extra \textit{data augmentation and exploration techniques} may expose the agent to a broader variety of stochastic scenarios, improving robustness and helping to avoid local optima.
    \item \textbf{Enhancing the reward and state representation}: The current design relies solely on the number of waiting patients and does not account for individual waiting times. This limits the agent’s ability to make patient-level decisions. A more nuanced modeling approach -- incorporating patient-level waiting times -- could provide a richer foundation for policy learning. Expanding the state space to include more contextual features, alongside a redefined reward function that captures waiting time penalties, would promote a more holistic optimization objective.
    \item \textbf{Increasing simulation realism}: The current framework focuses solely on outpatient workflows and simplifies staff schedules.
    A more realistic simulation environment could model interactions with other diagnostic departments and incorporate explicit scheduling of staff activities, such as lunch breaks, handovers, and variable shift patterns.
    These enhancements would increase the fidelity of the simulation and provide a more robust testbed for evaluating practical echo resource allocation policies.
    \item \textbf{Adopting continuous representations and policy-based methods}:
    The current approach relies on discrete state and action spaces, which may limit the expressiveness of the learned policy in capturing subtle operational trade-offs.
    In future work, we plan to adopt continuous state--action representations and explore policy gradient methods, such as actor--critic frameworks, to enable smoother policy updates and improved scalability.
    These methods may also help mitigate some of the challenges associated with high-dimensional state spaces and partial observability.
    \item \textbf{Addressing temporal credit assignment through intermediate penalties}:
    As noted above (Section~\ref{sec:discussion}), the long decision horizon limits the influence of end-of-episode penalties on earlier decisions under a discounted reward formulation.
    Introducing intermediate penalties—such as backlog growth costs or projected unfinished workload penalties—could provide denser feedback signals and improve temporal credit assignment.
    Such reward shaping strategies may encourage the agent to account for long-term system congestion earlier in the decision process, leading to more globally efficient allocation behavior under resource scarcity.
\end{enumerate}

\section{Conclusions}\label{sec:conclusions}
This study developed a simulation framework for comparing resource allocation schemes in a pediatric echocardiography laboratory, comparing on-the-fly, reservation-based, and hybrid allocation schemes through stochastic simulation.
By modeling patient arrivals, examination durations, and resource constraints, calibrated against one week of operational data, we demonstrated that the on-the-fly allocation scheme reduces mean patient waiting time relative to every reservation-based configuration we tested (Section~\ref{sec:predefined_policies}). We then compared both rule-based and reinforcement learning (RL)-based policies under the on-the-fly scheme.

Reinforcement learning was introduced as a data-driven approach capable of learning competitive allocation behaviors without reliance on explicit rule-based programming.
Concretely, the RL policy matched but did not exceed policy 2, and accumulated a \emph{higher} penalty after the 5:00~PM closing time (Section~\ref{sec:lp}), in line with similar findings elsewhere~\citep{gijsbrechts2022deep}.
These findings underscore the importance of careful reward design, robust state observability, and training stability.
The RL agent’s conservative behavior under uncertainty reflects a cautious trade-off strategy, but also reveals opportunities for enhancing long-term optimization.
Looking ahead, providing heuristic guidance to the agent, refining the state representation, and implementing more advanced training strategies could substantially enhance policy performance. Furthermore, incorporating additional real-world constraints and operational complexities into the simulation environment would improve the practical applicability and robustness of the proposed methods.
Overall, our findings highlight the potential of adaptive, learning-based allocation policies in healthcare operations and point to the value of continued interdisciplinary collaboration among data science, operations research, and clinical practice.

\section*{Ethics, Data Availability, and Author Contributions}

This study was approved by the Stanford University Institutional Review Board, and relied exclusively on de-identified aggregate data and did not involve identifiable private information.

\textbf{Software.} This work relies on SimPy~\cite{simpyteam}, Gymnasium~\cite{towers2024gymnasium}, PyTorch~\cite{paszke2019pytorch}, SciPy~\cite{virtanen2020scipy}, NumPy~\cite{harris2020numpy}, and Matplotlib~\cite{hunter2007matplotlib}.

\section*{AI Usage Statement}

The policy algorithm design, deep reinforcement learning network architectures, and simulation environment presented in this work were designed and implemented entirely by the authors.
During manuscript preparation, the authors used Claude (Anthropic), an AI-based language model, to assist with editing tasks, including refining prose, formatting hyperparameter and descriptive-statistics tables, writing plotting code used to generate the manuscript's figures, and cross-checking the manuscript's textual descriptions of the algorithm against the authors' own training code for accuracy. 
The AI tool was not involved in the design, development, or implementation of the study's methodology, and was not used to generate, analyze, or interpret the study's scientific results, data, or conclusions. All AI-assisted edits were reviewed and verified by the authors, who take full responsibility for the accuracy, originality, and content of the published work.

\section*{Acknowledgements}

DES acknowledges support from an NSF CAREER award \#1942662 and an NSF CDS\&E award \#2104831. BS, DES, JAF, and ALM acknowledge support from NIH grant \#1R01HL167516 ``{\it Uncertainty-aware virtual treatment planning for peripheral pulmonary artery stenosis}'' (PI ALM). The authors would like to thank the Center for Research Computing at the University of Notre Dame for providing computational resources and support that were essential to generate model results for this study.

\bibliography{draft}

\appendix

\section{Appendix}
\subsection{Data analysis}
\subsubsection{Analysis of patient's no-show rate}\label{sec:nalysis of patient's no-show rate}
\begin{figure}[ht!]
\centering
\includegraphics[width=1\textwidth]{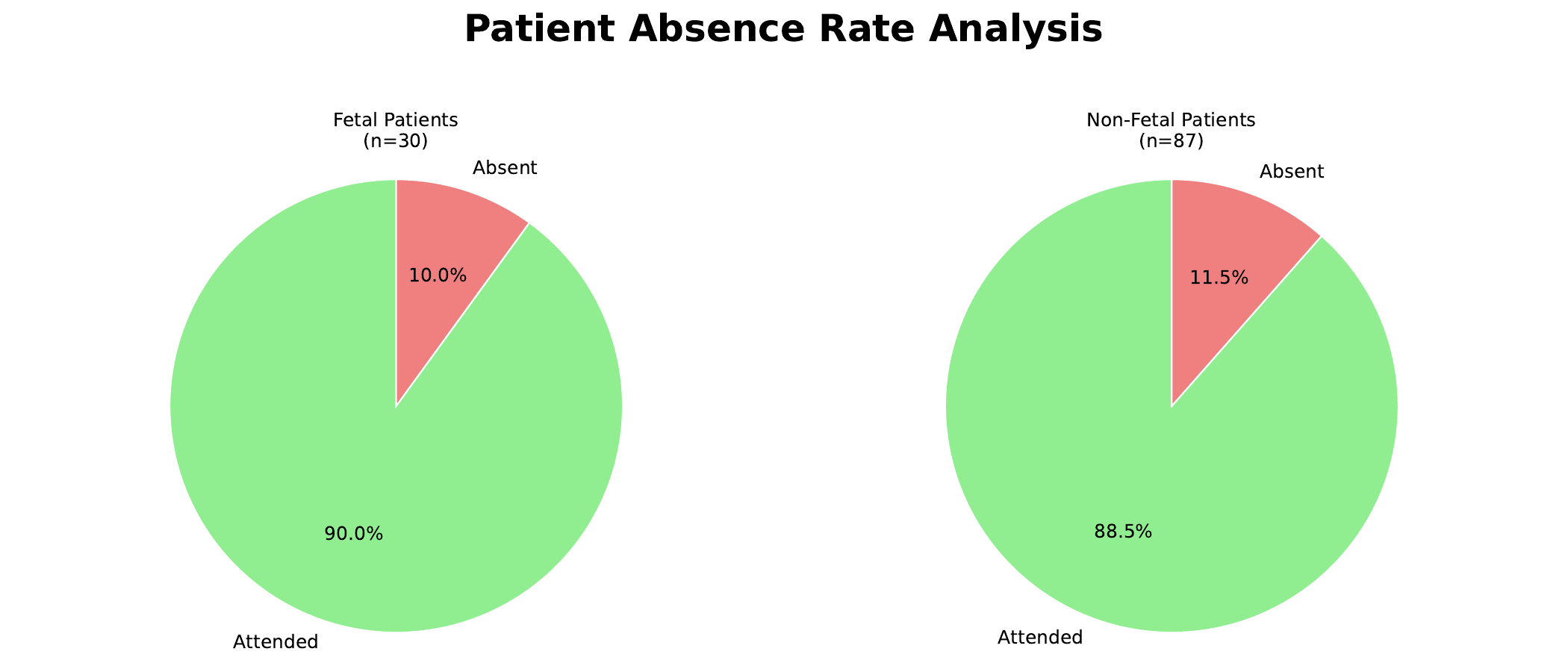}
\caption{Pie charts for patient's absence rate.}
\label{fig:absence_rate}
\end{figure}

\begin{figure}[ht!]
\centering
\includegraphics[width=1\textwidth]{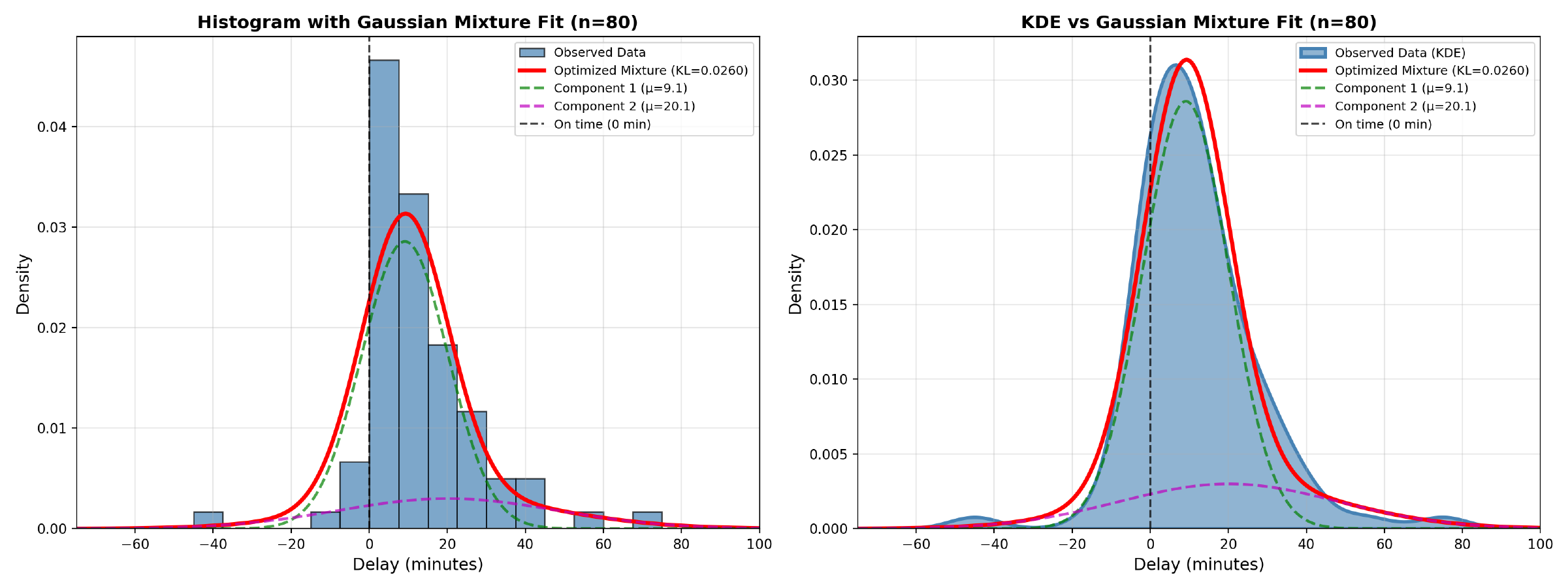}
\caption{Distribution of patients' arrival delay (difference between actual and scheduled arrival time)}
\label{fig:arrival_time}
\end{figure}

We define as \emph{absence} (or \emph{no-show}) a patient with a scheduled echo time and no recorded echo exam start time.
Using this definition, the estimated no-show rates in the dataset we received were 10.0\% for fetal patients and 11.5\% for non-fetal patients, as shown in Figure~\ref{fig:absence_rate}.
Given the small difference, we assume a common absence rate of 10\% for subsequent analyses.
\rev{
\subsubsection{Analysis of patient's delay time}\label{sec:Analysis of patient's delay time}
We consider as \emph{patient's delay} the difference between each patient's arrival time and their scheduled visit time.
Due to the limited sample size, we conducted a series of non-parametric statistical tests to evaluate whether the distribution of such delay differs significantly between fetal and non-fetal patients.
These tests provide a robust, distribution-free framework that avoids the unrealistic assumption of normality often violated in clinical data.
Specifically, the Mann–Whitney U test~\cite{mann1947test} is a test of stochastic dominance, which reduces to a test of medians under a location-shift assumption that we do not separately verify; the Kolmogorov–Smirnov test~\cite{kolmogorov1933sulla,smirnov1948table} evaluates whether the overall distribution shapes differ—capturing discrepancies in spread or skewness, as well as location.
Both tests are fully non-parametric and remain valid under non-normal, skewed, or multimodal conditions. To complement these, Levene's test~\cite{levene1960robust} (itself a parametric ANOVA on absolute deviations from group medians) examines equality of variances, identifying whether one group exhibits more variability in delays even when medians are similar.
To ensure reliability, we interpret results only when both non-parametric tests yield consistent conclusions, reducing the likelihood of spurious findings.
Finally, Cohen's $d$~\cite{cohen1988statistical} (which presumes approximately normal groups with comparable variances -- an assumption not verified here, and not obviously satisfied given the difference in dispersion between groups reported in Table~\ref{tab:desc_stats}) bridges statistical and practical significance by quantifying the magnitude of observed differences. This is essential because small datasets can mask clinically meaningful effects or, conversely, large samples can make trivial effects appear significant.
Together, these methods offer an assessment of whether the two delay-time distributions differ both statistically and clinically. Because the fetal subgroup is small ($n\approx$30 vs.\ 87 for non-fetal), the failure to reject the null hypothesis in these tests more likely reflects limited statistical power than true equivalence between the two groups; readers should interpret the pooled analysis with this caveat in mind.
All patients with both arrival and scheduled timestamps were included in the analysis. Descriptive statistics and test results are summarized in Tables~\ref{tab:desc_stats} and~\ref{tab:tests_effect}.
Because the sample size was limited and the non-parametric tests indicated that the differences between fetal and non-fetal patients were not statistically significant, we merged the two groups together for subsequent modeling.
Given the heavy-tailed nature of the data, we modeled the empirical distribution as a mixture of two Gaussian components.
The parameters of the mixture were derived by minimizing the Kullback–Leibler (KL) divergence between the kernel density estimate (KDE) of the empirical data and the fitted parametric model $g_\theta(x)$.
In practice, we minimized $D_{\mathrm{KL}}(\hat f\,\|\,g_\theta)$ using SciPy's~\cite{virtanen2020scipy} \texttt{minimize} with the Nelder–Mead simplex (derivative-free) algorithm (tolerances $x_{\text{atol}}=10^{-8}$, $f_{\text{atol}}=10^{-8}$, maximum 10{,}000 iterations), using reparameterization to enforce positivity.
The resulting fitted components are as follows: the first component has a mean of $\mu_1 = 9.12$ minutes and a standard deviation of $\sigma_1 = 11.02$ minutes, contributing $79.0\%$ of the total weight, while the second component has a mean of $\mu_2 = 20.14$ minutes and a standard deviation of $\sigma_2 = 27.98$ minutes, contributing the remaining $21.0\%$.
The final KL divergence after optimization was $0.0260$, indicating a close match between the analytical mixture and the empirical KDE.
Figure~\ref{fig:arrival_time} displays the histogram, kernel density estimate (KDE), and the fitted parametric mixture curves.}
%
\begin{table}[h!]
\centering
\renewcommand{\arraystretch}{1.15}
\begin{tabular}{lcc}
\toprule
\textbf{Metric} & \textbf{Fetal} & \textbf{Non-Fetal} \\
\midrule
Mean & 6.72 & 12.81 \\
Median & 5.00 & 10.50 \\
Standard Deviation & 8.38 & 16.89 \\
Minimum & -5.00 & -45.00 \\
Maximum & 25.00 & 75.00 \\
Q1 (25th percentile) & 0.00 & 1.25 \\
Q3 (75th percentile) & 11.50 & 20.00 \\
\bottomrule
\end{tabular}
\caption{Descriptive statistics of patient delay times (minutes).}
\label{tab:desc_stats}
\end{table}
\begin{table}[h!]
\centering
\resizebox{\textwidth}{!}{
\begin{tabular}{lccc}
\toprule
\textbf{Test} & \textbf{Statistic} & \textbf{p-value} & \textbf{Interpretation (at $\alpha=0.05$)} \\
\midrule
Mann--Whitney U & $U=404.50$ & 0.0765 & Not significant (no median difference) \\
Kolmogorov--Smirnov & $D=0.2401$ & 0.3355 & Not significant (shapes similar) \\
Levene’s Test & $W=2.4709$ & 0.1200 & Not significant (variances similar) \\
\midrule
Cohen’s $d$ & $d=-0.3941$ & --- & Small-to-medium effect (non-fetal larger delays) \\
\bottomrule
\end{tabular}}
\caption{Non-parametric test results comparing fetal vs.\ non-fetal delay distributions and effect size.}
\label{tab:tests_effect}
\end{table}

\subsubsection{Analysis of patient's echo duration}\label{Analysis of patient's echo duration}
\begin{figure}[ht!]
\centering
\includegraphics[width=1\textwidth]{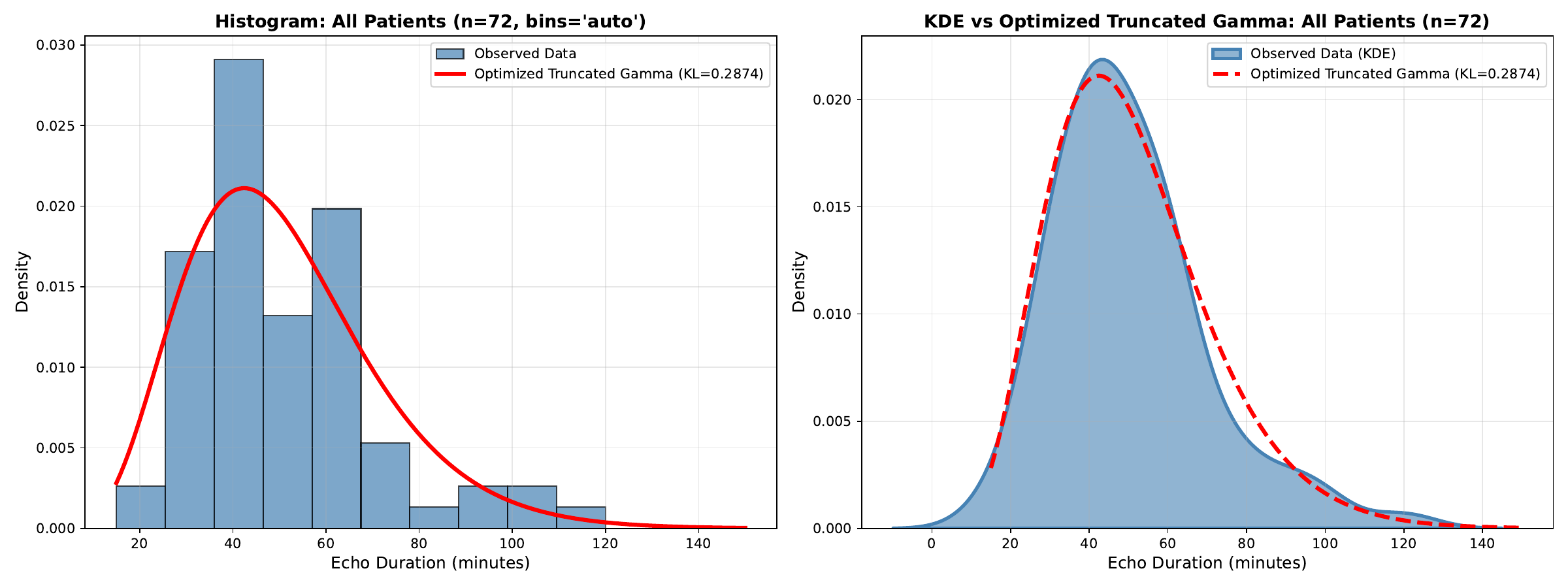}
\caption{Distribution of patients' echo duration.}
\label{fig:echo_duration}
\end{figure}

We recorded all patients with available echo duration data. We then estimated the underlying parametric distribution of their echo duration times and visualized the results using a histogram, a kernel density estimate (KDE), and the fitted parametric distribution curve, as shown in Figure~\ref{fig:echo_duration}.
Because the sample size was limited and the statistical tests indicated no significant differences between fetal and non-fetal patients, we combined the two groups for modeling (see the pooling caveat and equivalence-test discussion in Section~\ref{sec:Analysis of patient's delay time}). 
Prior to analysis, echo durations were truncated to the range of 15 to 150 minutes; no observation in the underlying data falls outside this range (Table~\ref{tab:echo_desc_stats}: max 120/99, min 28/15 minutes for fetal/non-fetal), so the truncation bounds the simulator's sampling range rather than removing any outlier.
To model the empirical distribution, we used a truncated Gamma distribution, which is well-suited for capturing skewed data bounded within a finite interval.
The parameters of the truncated Gamma distribution were derived by minimizing the Kullback–Leibler (KL) divergence between the KDE-derived empirical distribution $\hat f(x)$ and the candidate parametric model $g_\theta(x)$.
We optimized $D_{\mathrm{KL}}(\hat f\,\|\,g_\theta)$ with SciPy's~\cite{virtanen2020scipy} \texttt{minimize} using the Nelder–Mead simplex (derivative-free) method (tolerances $xatol=10^{-8}$, $fatol=10^{-8}$, maximum 10,000 iterations), enforcing positivity of the shape parameters via reparameterization.
This approach provides a stable, gradient-free fit that closely tracks the smoothed empirical density while preserving parametric interpretability.
Descriptive statistics and test results are summarized in Tables~\ref{tab:echo_desc_stats} and~\ref{tab:echo_tests}. Overall, the average echo duration was approximately $52$ minutes, with moderate variability across patients.
The Mann–Whitney U test ($U = 586.5$, $p = 0.193$) indicated no significant difference in median duration between fetal and non-fetal groups. Similarly, the Kolmogorov–Smirnov test ($D = 0.259$, $p = 0.301$) and Levene's test ($W = 0.292$, $p = 0.591$) found no significant differences in distribution shape or variance, respectively. Cohen’s $d = 0.378$ suggested a small-to-medium practical effect, with slightly longer average durations for fetal examinations.
The fitted truncated Gamma distribution yielded a shape parameter of $k = 6.1151$ and a scale parameter of $\theta = 8.2967$ (reported to two decimals in the main text, Section~\ref{Analysis of patient's echo duration}, as $k=6.12,\theta=8.30$; the two should be reconciled to a single precision), with truncation bounds of $[15, 150]$ minutes. 
The final Kullback–Leibler divergence between the fitted model and the KDE was $0.287$, indicating that the truncated Gamma model provided a reasonable but imperfect approximation of the empirical data. Figure~\ref{fig:echo_duration} illustrates the agreement between the observed and modeled distributions.
\begin{table}[h!]
\centering
\renewcommand{\arraystretch}{1.15}
\begin{tabular}{lcc}
\toprule
\textbf{Metric} & \textbf{Fetal} & \textbf{Non-Fetal} \\
\midrule
Mean & 55.78 & 48.44 \\
Median & 57.50 & 45.00 \\
Standard Deviation & 23.01 & 18.06 \\
Minimum & 28.00 & 15.00 \\
Maximum & 120.00 & 99.00 \\
Q1 (25th percentile) & 41.25 & 36.00 \\
Q3 (75th percentile) & 60.00 & 58.00 \\
\bottomrule
\end{tabular}
\caption{Descriptive statistics of echo duration times (minutes).}
\label{tab:echo_desc_stats}
\end{table}
\begin{table}[h!]
\centering
\resizebox{\textwidth}{!}{
\begin{tabular}{lccc}
\toprule
\textbf{Test} & \textbf{Statistic} & \textbf{p-value} & \textbf{Interpretation (at $\alpha=0.05$)} \\
\midrule
Mann--Whitney U & $U = 586.50$ & 0.1927 & Not significant (no median difference) \\
Kolmogorov--Smirnov & $D = 0.2593$ & 0.3012 & Not significant (shapes similar) \\
Levene’s Test & $W = 0.2922$ & 0.5905 & Not significant (variances similar) \\
\midrule
Cohen’s $d$ & $d = 0.3784$ & --- & Small-to-medium effect (fetal durations slightly longer) \\
\bottomrule
\end{tabular}}
\caption{Non-parametric test results comparing fetal vs.\ non-fetal echo duration distributions and effect size.}
\label{tab:echo_tests}
\end{table}
\subsection{Result plots for policy 5 ($\alpha, \beta$) and policy 6 ($\alpha, \beta$)}
\subsubsection{Policy 5 ($\alpha, \beta$)}\label{policy 5}
This subsection reports, for each tested reservation fraction $(\alpha,\beta)$, the average echo-resource utilization and waiting-patient counts (mean and standard deviation) for policy 5, across 365 independent replications of a representative day; Section~\ref{sec:predefined_policies} discusses the main-text case ($\alpha=100\%,\beta=100\%$) and the overall pattern across cases.
\begin{figure}[ht!]
\centering
\includegraphics[width=0.32\textwidth]{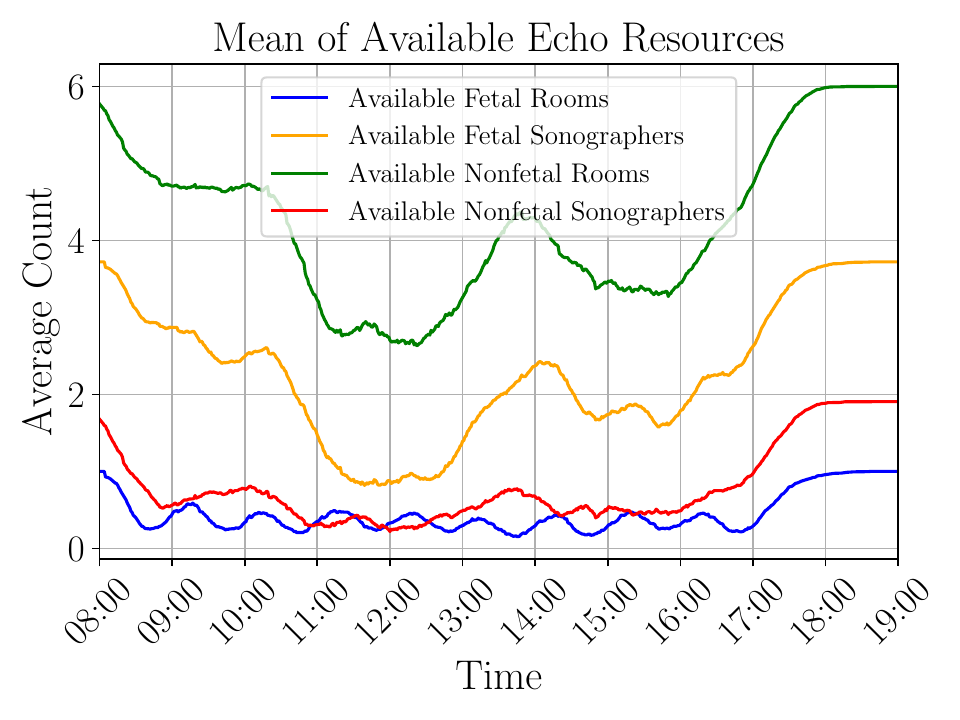}
\hfill
\includegraphics[width=0.32\textwidth]{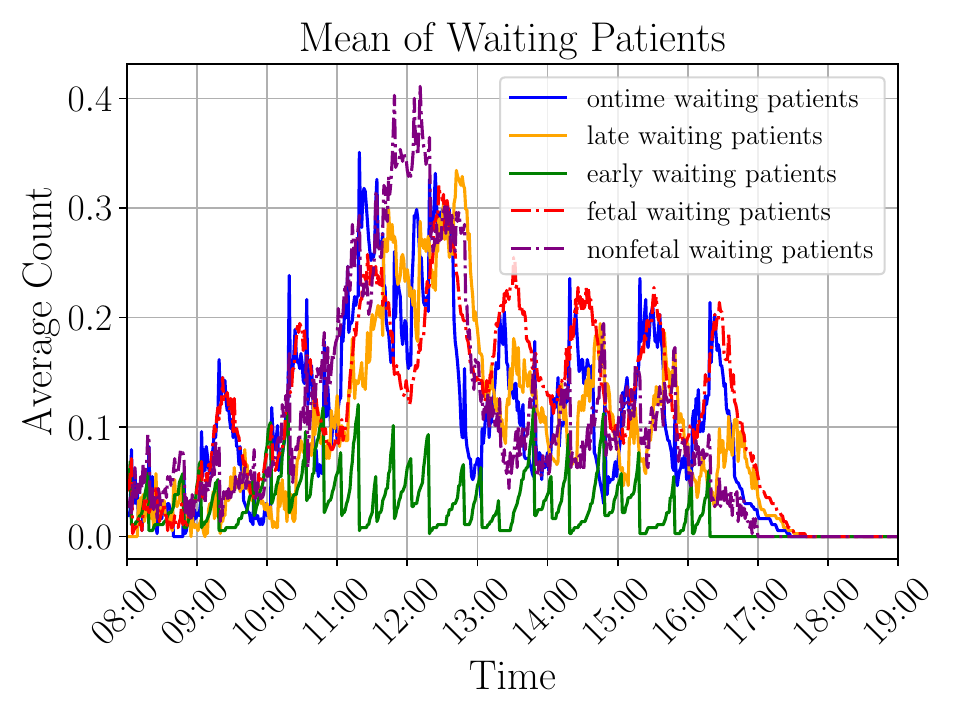}
\hfill
\includegraphics[width=0.32\textwidth]{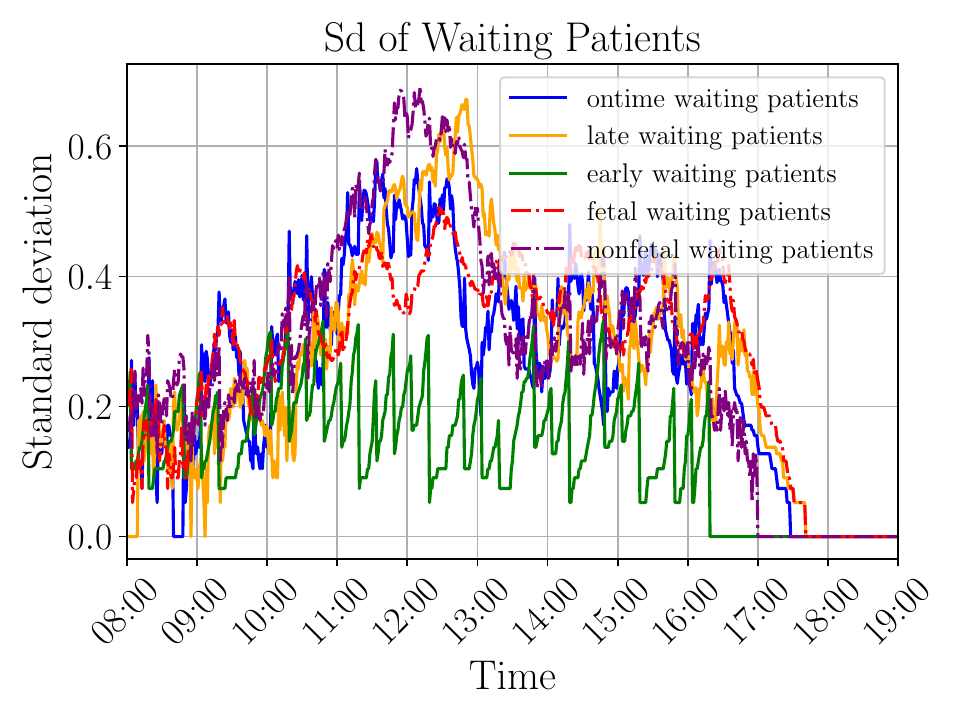}
\caption{Average results in terms of echo resources and waiting patients over 365 independent replications of a representative day for the stochastic hospital process enforcing policy 5 with $\alpha = 0\%$ and $\beta = 0\%$.}\label{fig:res_Policy5 0 0}
\end{figure}
\begin{figure}[ht!]
\centering
\includegraphics[width=0.32\textwidth]{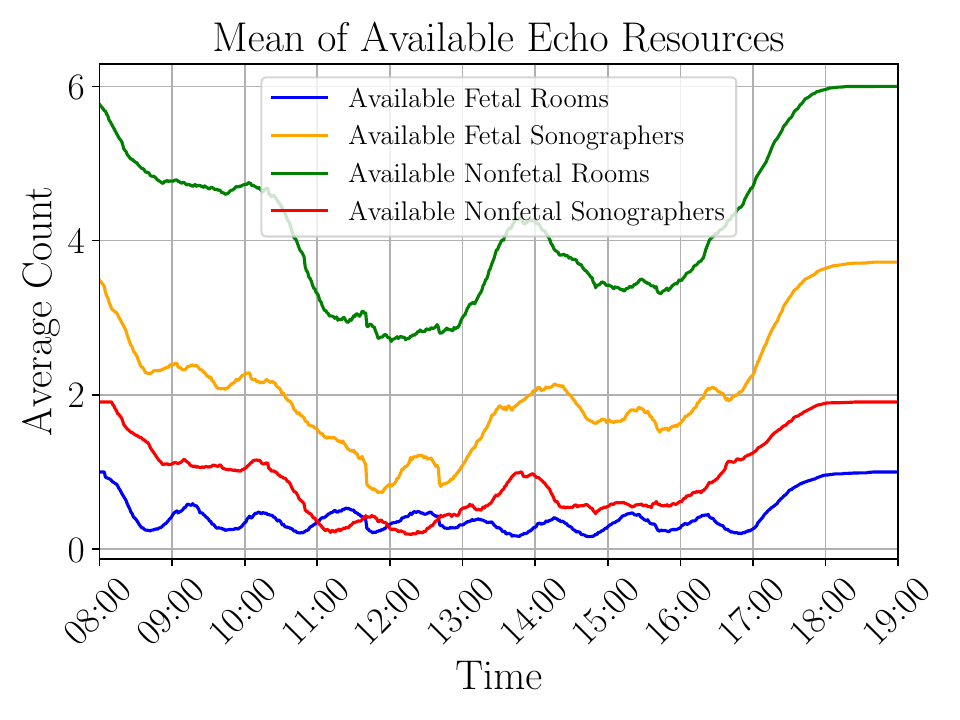}
\hfill
\includegraphics[width=0.32\textwidth]{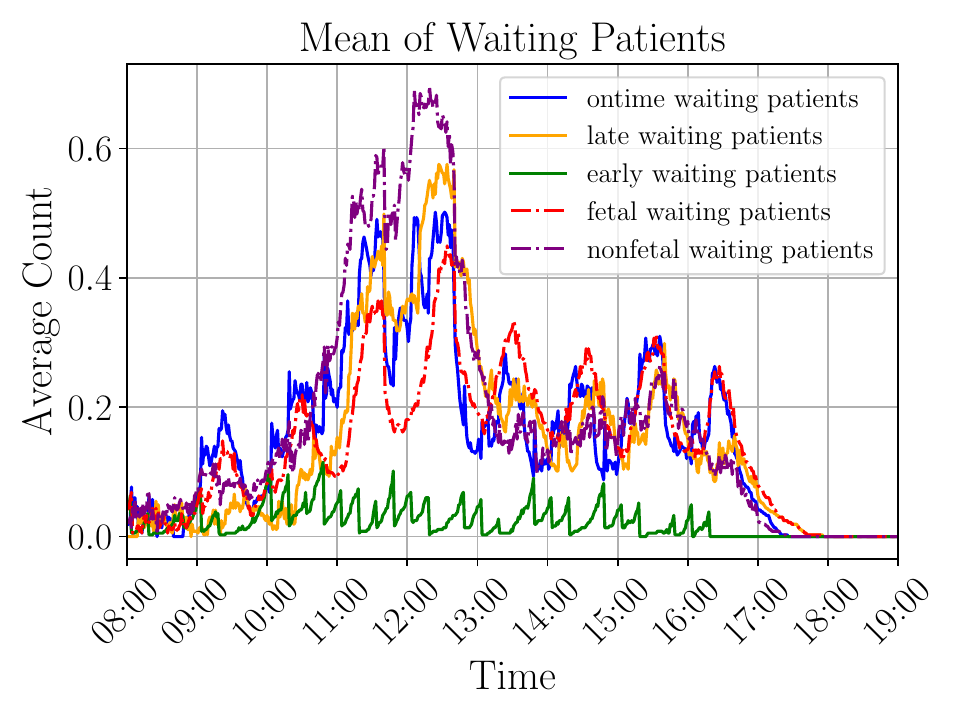}
\hfill
\includegraphics[width=0.32\textwidth]{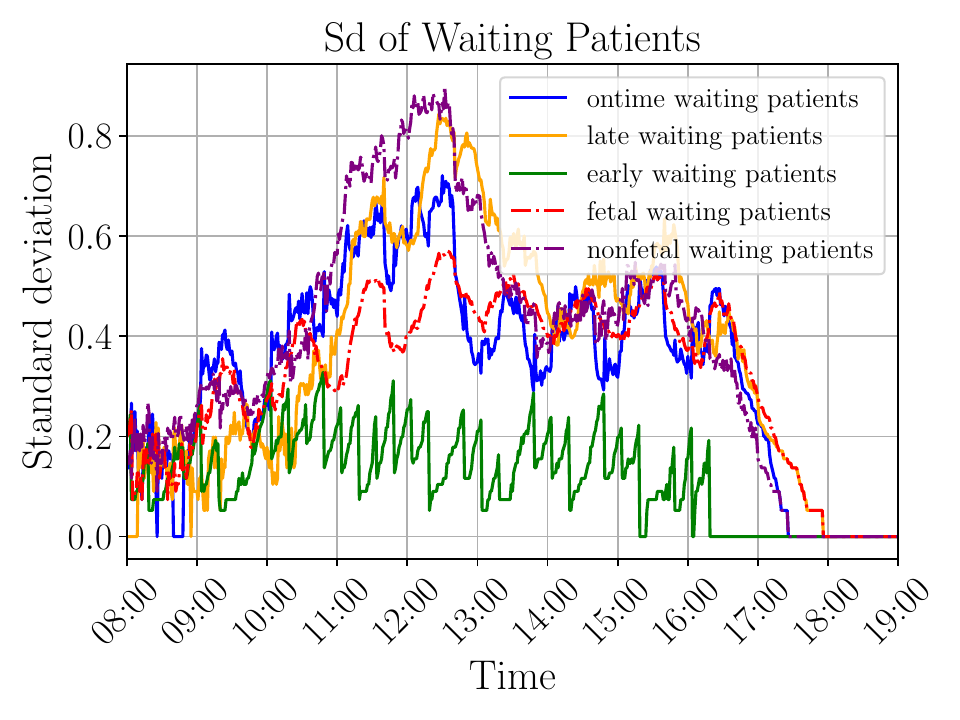}
\caption{Average results in terms of echo resources and waiting patients over 365 independent replications of a representative day for the stochastic hospital process enforcing policy 5 with $\alpha = 0\%$ and $\beta = 25\%$}\label{fig:res_Policy5 0 25}
\end{figure}
\begin{figure}[ht!]
\centering
\includegraphics[width=0.32\textwidth]{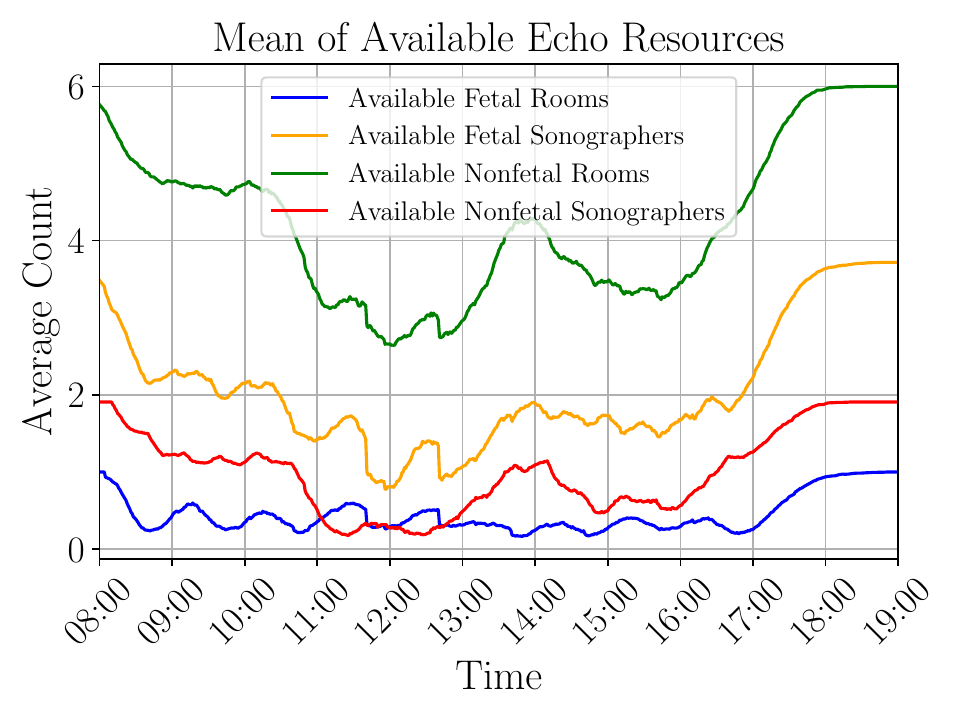}
\hfill
\includegraphics[width=0.32\textwidth]{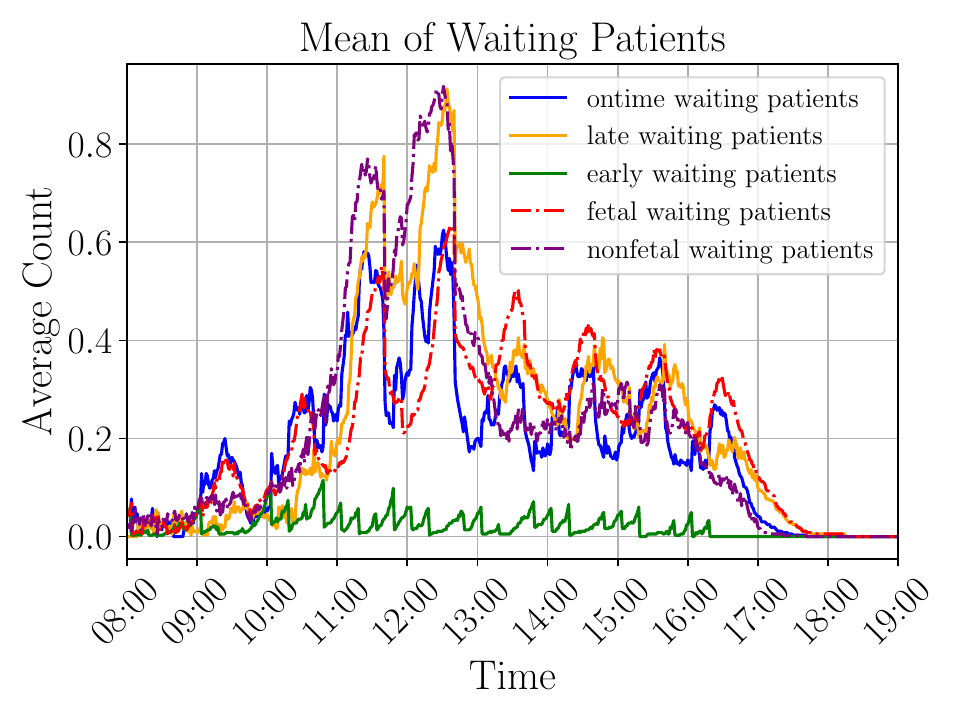}
\hfill
\includegraphics[width=0.32\textwidth]{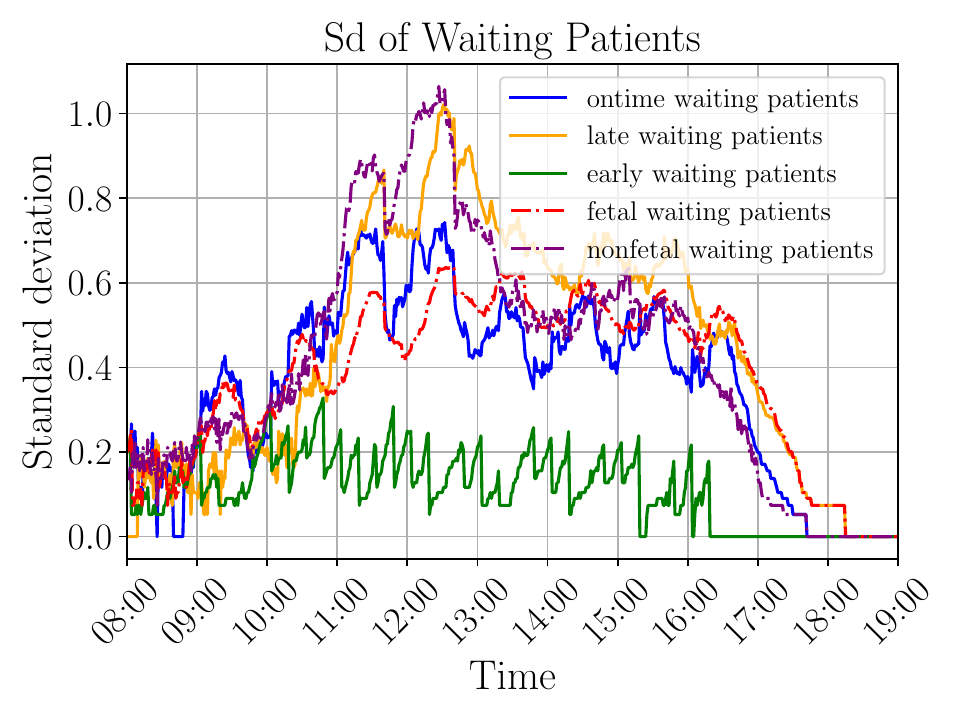}
\caption{Average results in terms of echo resources and waiting patients over 365 independent replications of a representative day for the stochastic hospital process enforcing policy 5 with $\alpha = 0\%$ and $\beta = 50\%$}\label{fig:res_Policy5 0 50}
\end{figure}
\begin{figure}[ht!]
\centering
\includegraphics[width=0.32\textwidth]{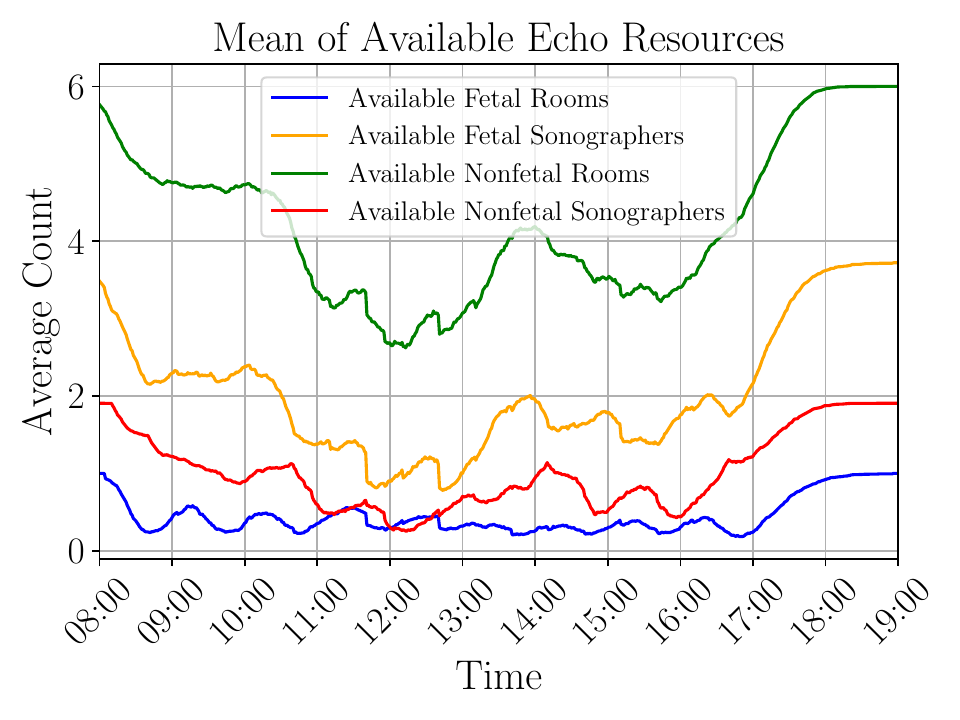}
\hfill
\includegraphics[width=0.32\textwidth]{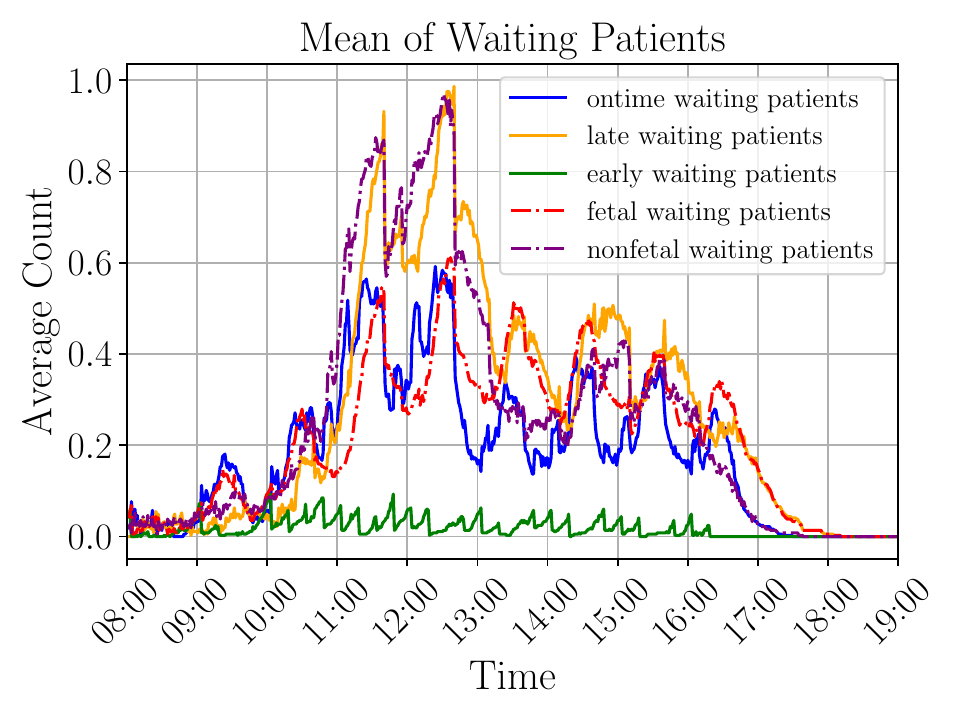}
\hfill
\includegraphics[width=0.32\textwidth]{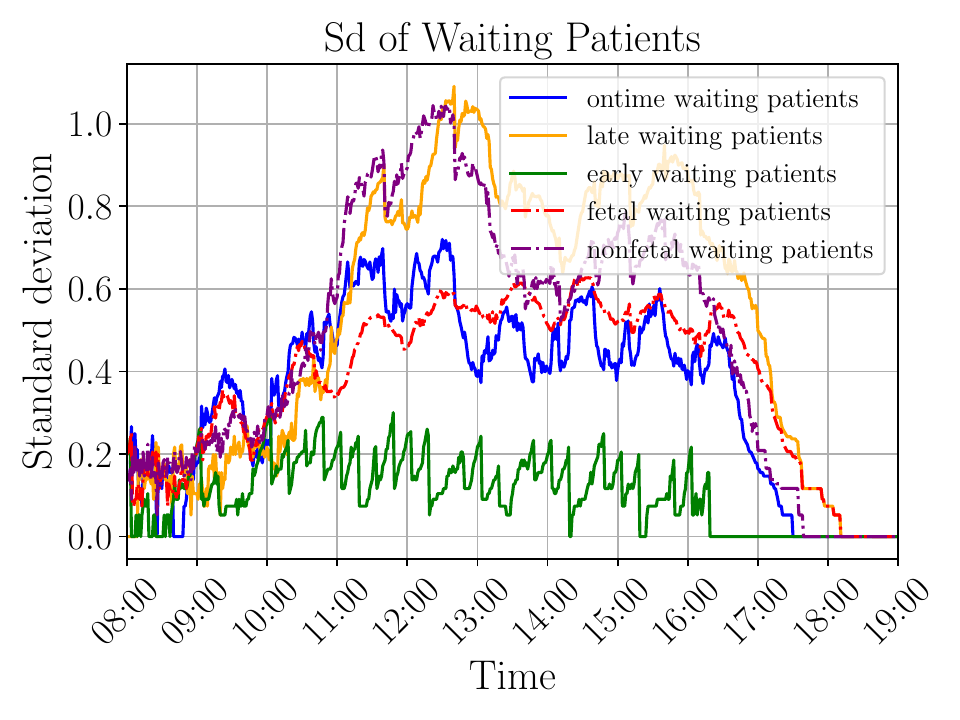}
\caption{Average results in terms of echo resources and waiting patients over 365 independent replications of a representative day for the stochastic hospital process enforcing policy 5 with $\alpha = 0\%$ and $\beta = 75\%$}\label{fig:res_Policy5 0 75}
\end{figure}
\begin{figure}[ht!]
\centering
\includegraphics[width=0.32\textwidth]{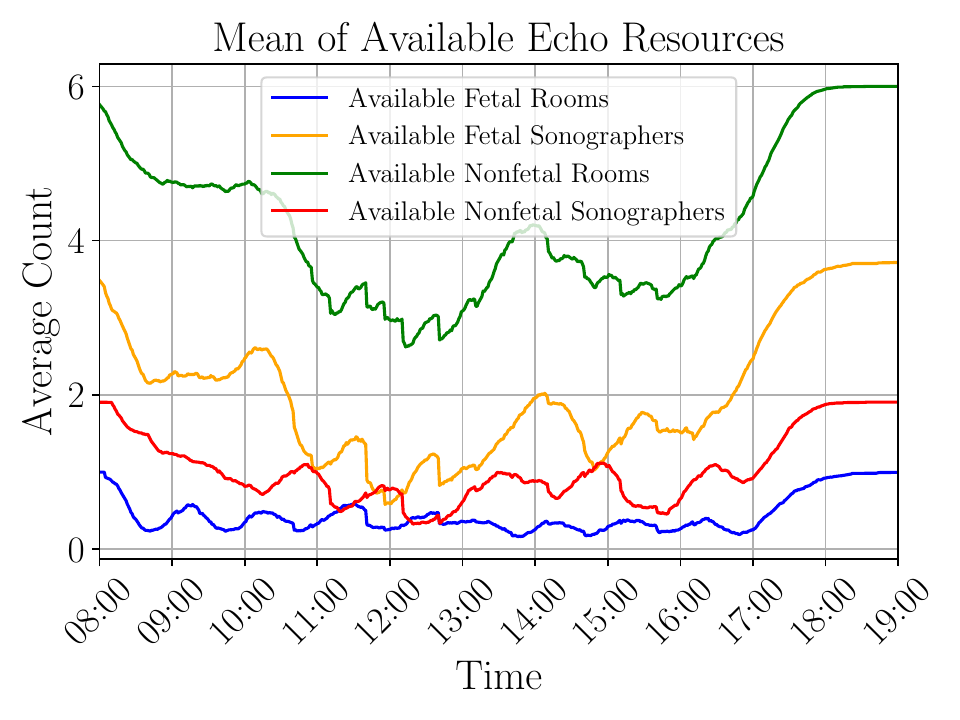}
\hfill
\includegraphics[width=0.32\textwidth]{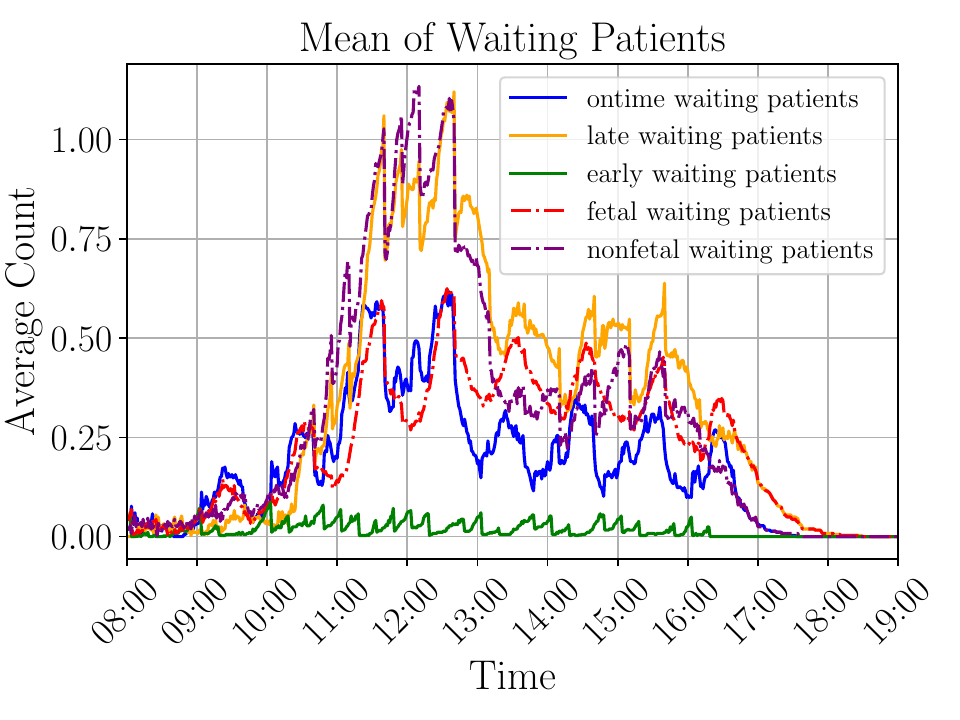}
\hfill
\includegraphics[width=0.32\textwidth]{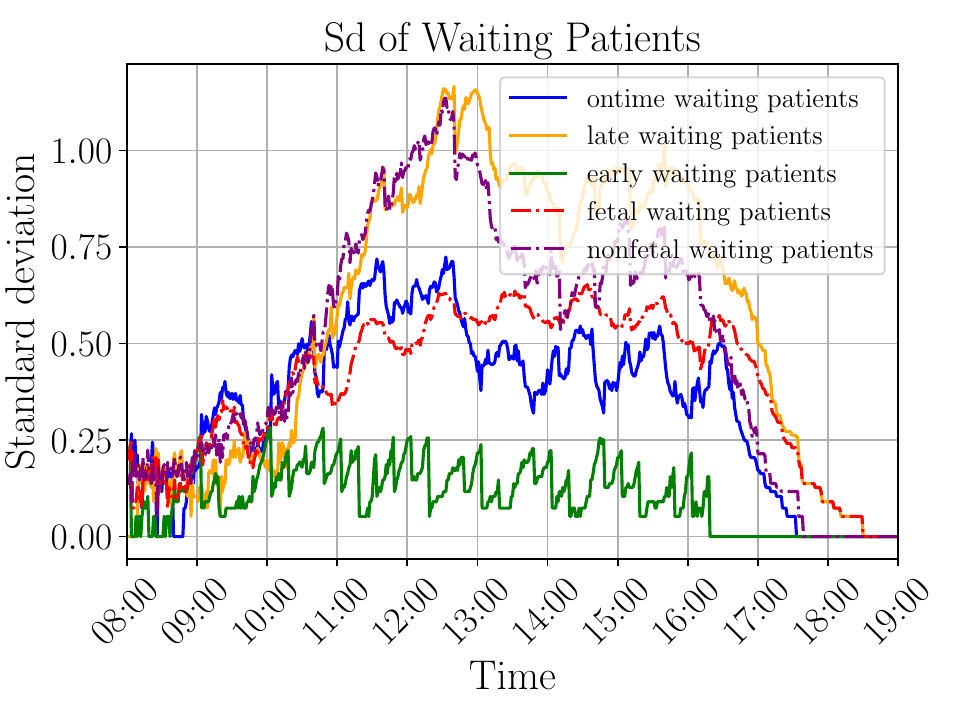}
\caption{Average results in terms of echo resources and waiting patients over 365 independent replications of a representative day for the stochastic hospital process enforcing policy 5 with $\alpha = 0\%$ and $\beta = 100\%$}\label{fig:res_Policy5 0 100}
\end{figure}
\begin{figure}[ht!]
\centering
\includegraphics[width=0.32\textwidth]{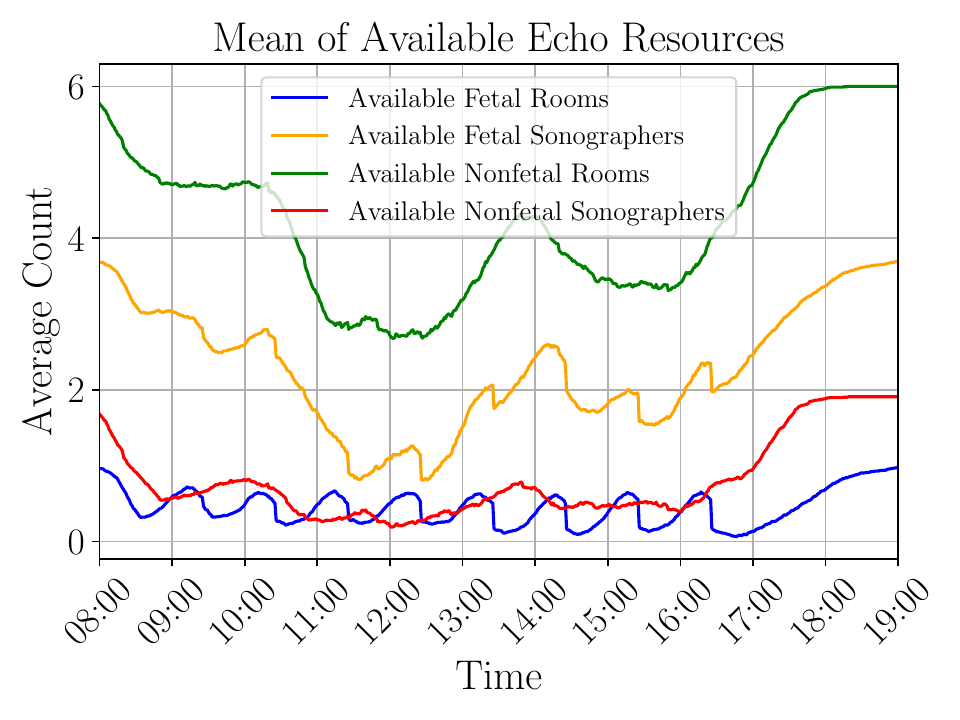}
\hfill
\includegraphics[width=0.32\textwidth]{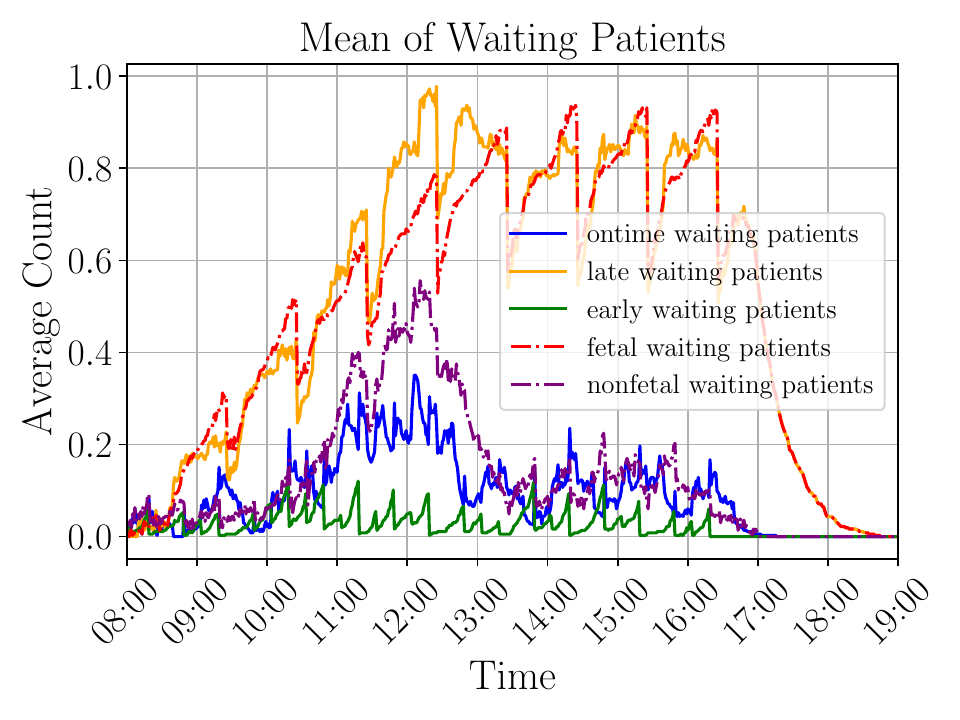}
\hfill
\includegraphics[width=0.32\textwidth]{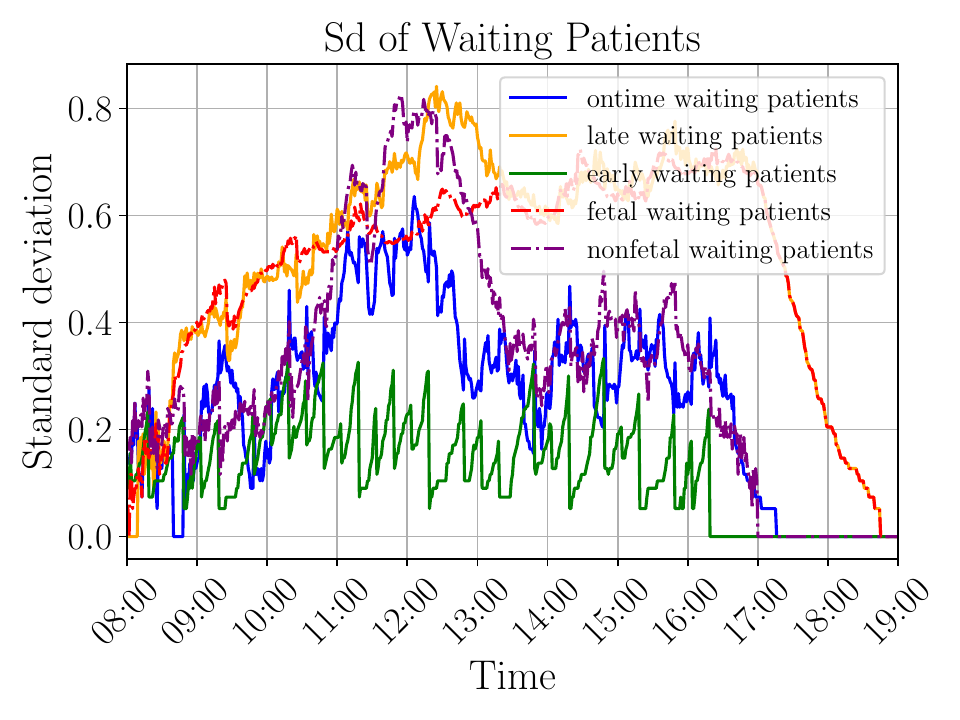}
\caption{Average results in terms of echo resources and waiting patients over 365 independent replications of a representative day for the stochastic hospital process enforcing policy 5 with $\alpha = 100\%$ and $\beta = 0\%$}\label{fig:res_Policy5 100 0}
\end{figure}
\begin{figure}[ht!]
\centering
\includegraphics[width=0.32\textwidth]{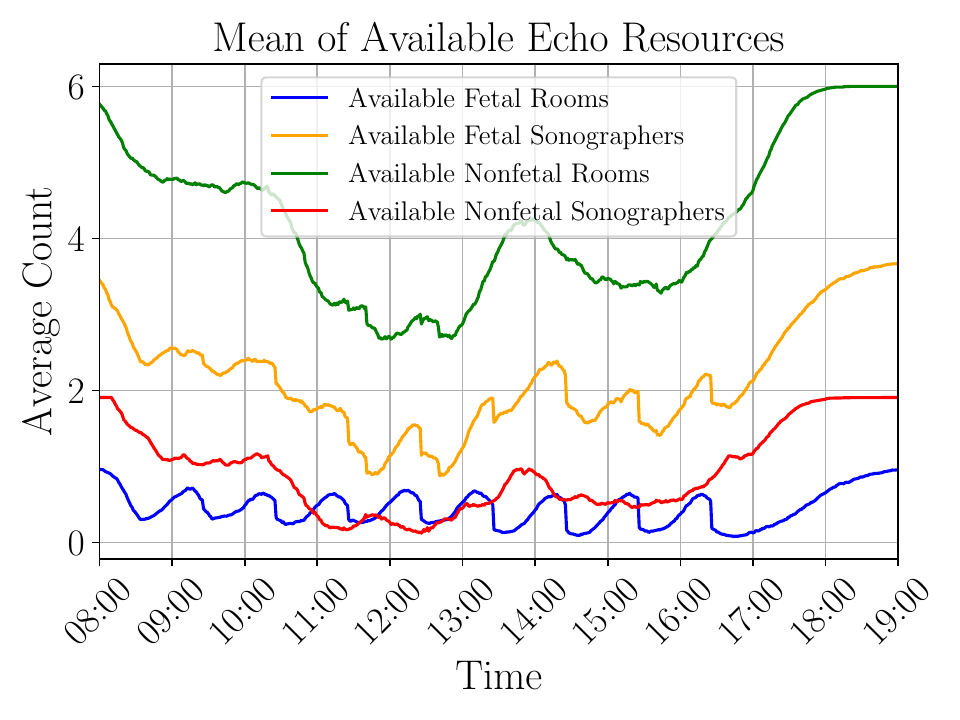}
\hfill
\includegraphics[width=0.32\textwidth]{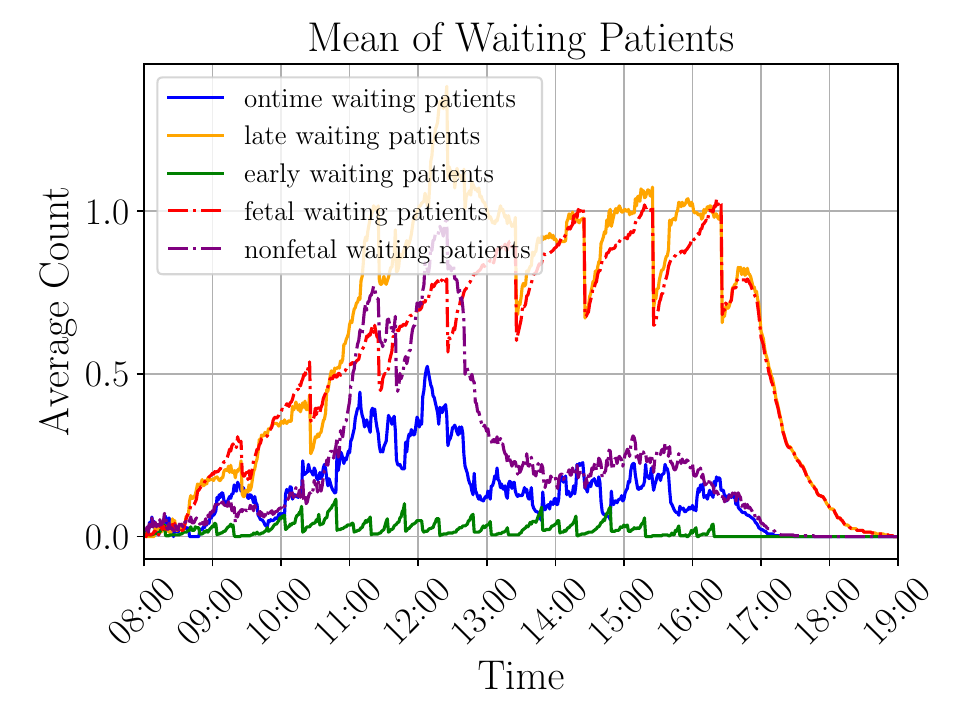}
\hfill
\includegraphics[width=0.32\textwidth]{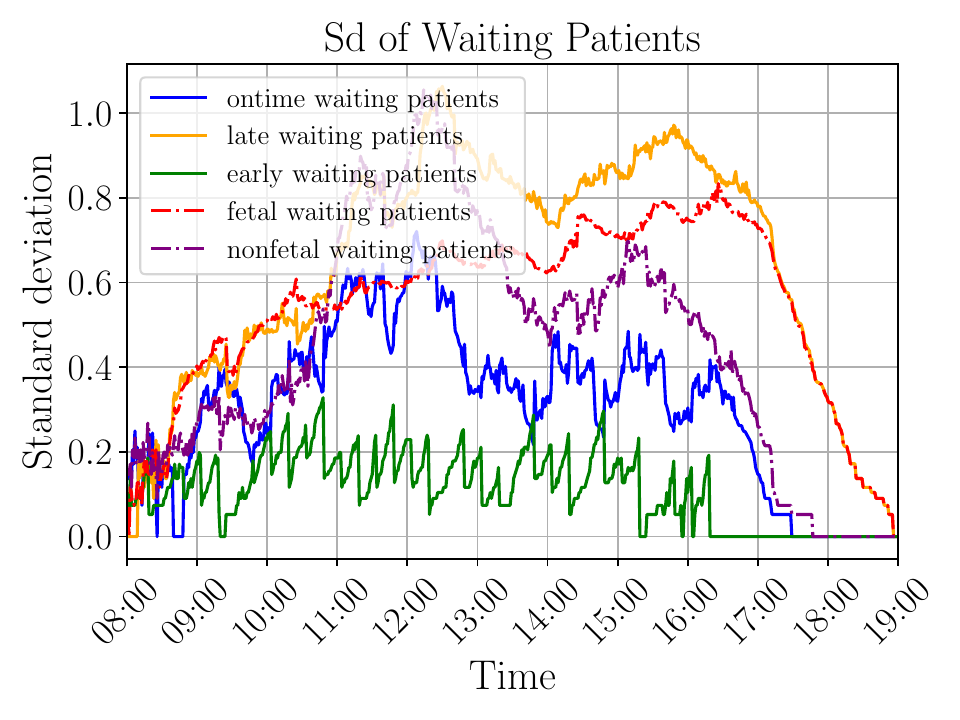}
\caption{Average results in terms of echo resources and waiting patients over 365 independent replications of a representative day for the stochastic hospital process enforcing policy 5 with $\alpha = 100\%$ and $\beta = 25\%$}\label{fig:res_Policy5 100 25}
\end{figure}
\begin{figure}[ht!]
\centering
\includegraphics[width=0.32\textwidth]{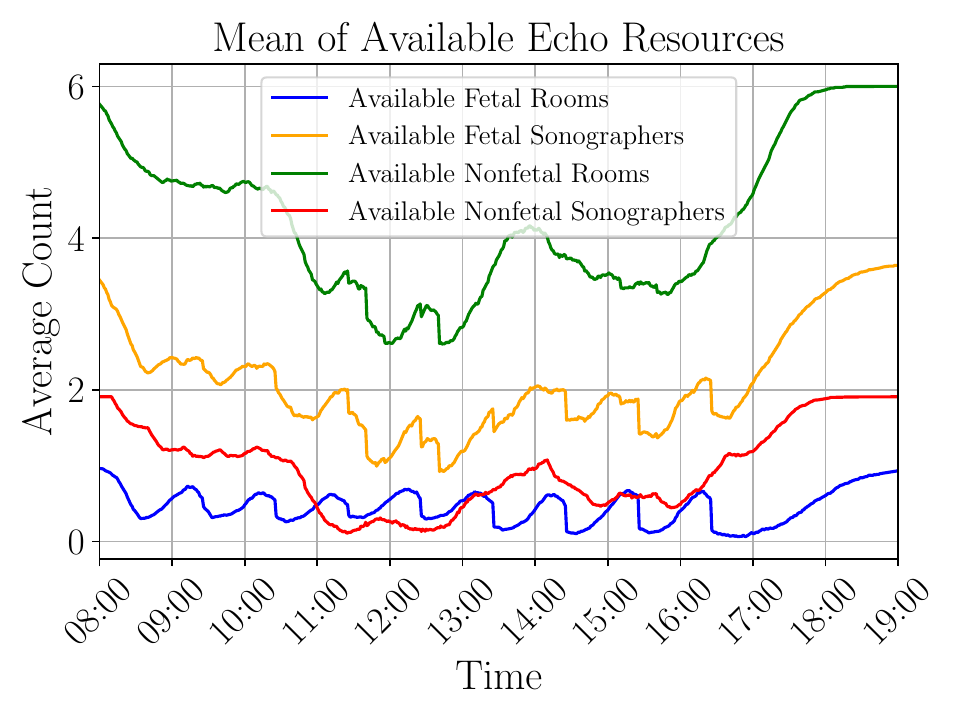}
\hfill
\includegraphics[width=0.32\textwidth]{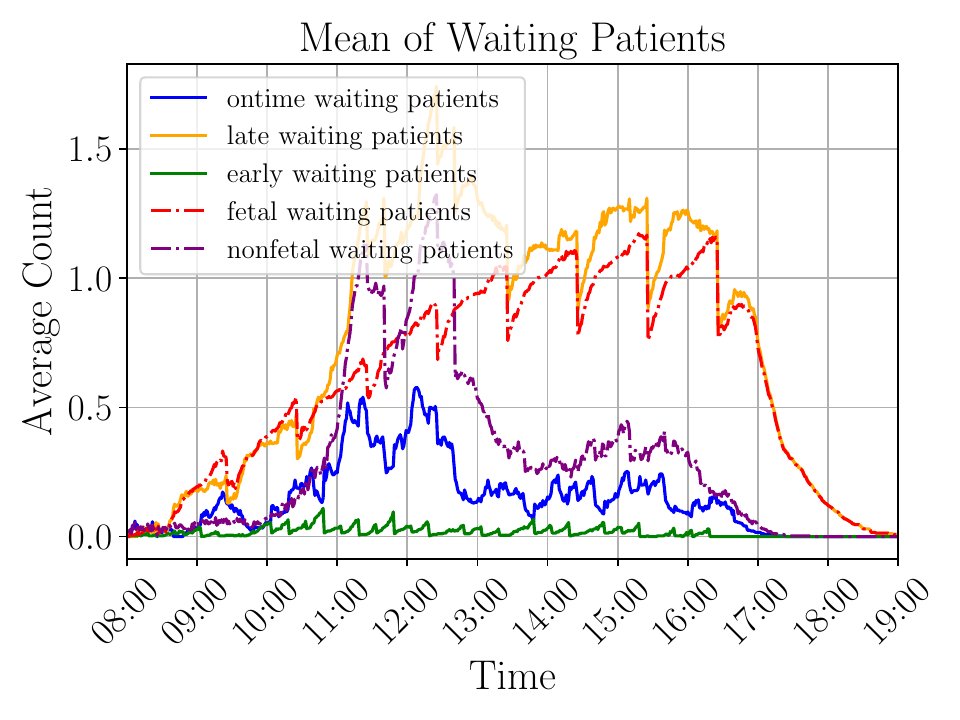}
\hfill
\includegraphics[width=0.32\textwidth]{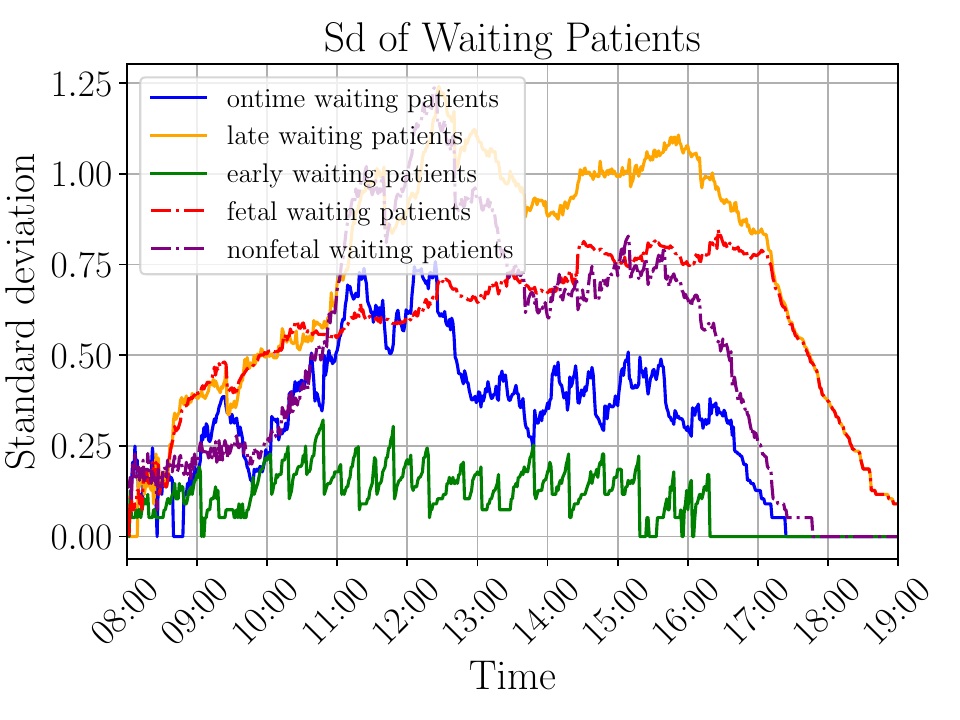}
\caption{Average results in terms of echo resources and waiting patients over 365 independent replications of a representative day for the stochastic hospital process enforcing policy 5 with $\alpha = 100\%$ and $\beta = 50\%$}\label{fig:res_Policy5 100 50}
\end{figure}
\begin{figure}[ht!]
\centering
\includegraphics[width=0.32\textwidth]{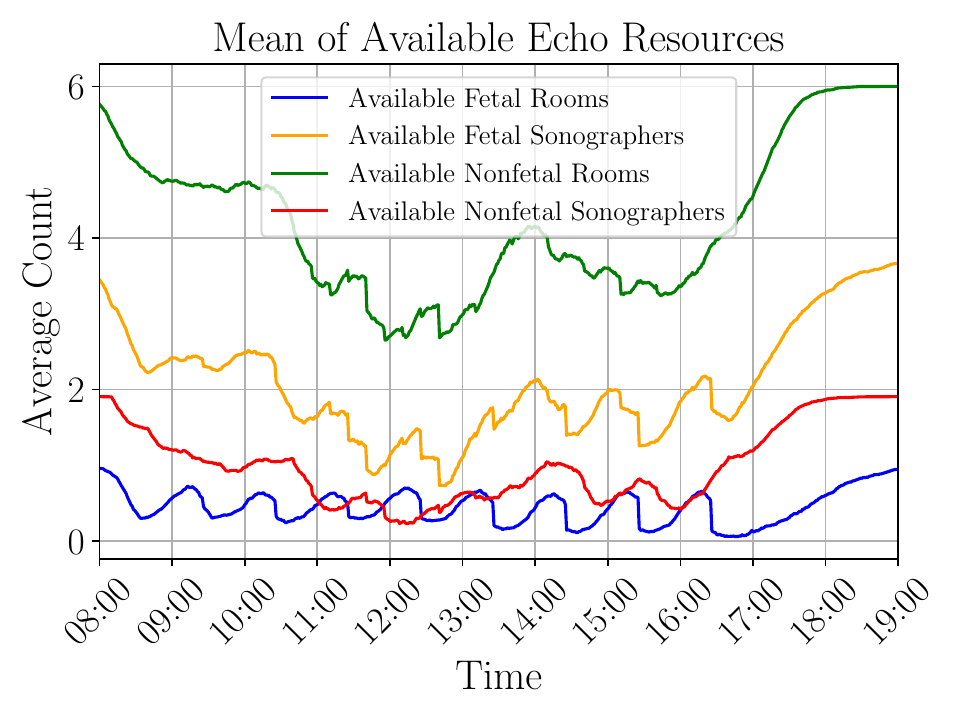}
\hfill
\includegraphics[width=0.32\textwidth]{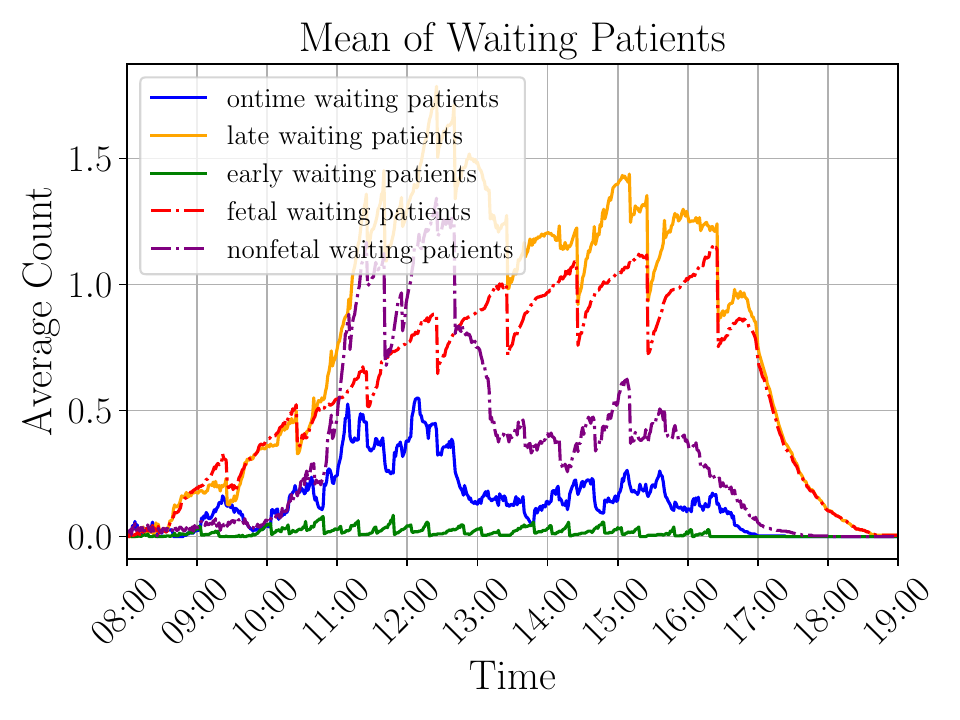}
\hfill
\includegraphics[width=0.32\textwidth]{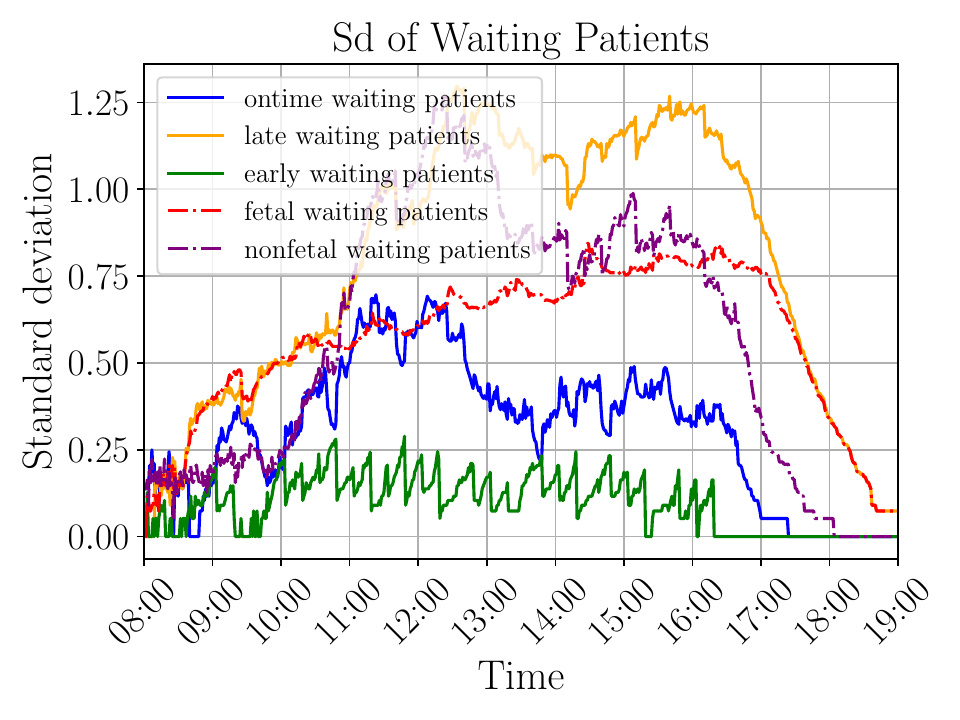}
\caption{Average results in terms of echo resources and waiting patients over 365 independent replications of a representative day for the stochastic hospital process enforcing policy 5 with $\alpha = 100\%$ and $\beta = 75\%$}\label{fig:res_Policy5 100 75}
\end{figure}
\begin{figure}[ht!]
\centering
\includegraphics[width=0.32\textwidth]{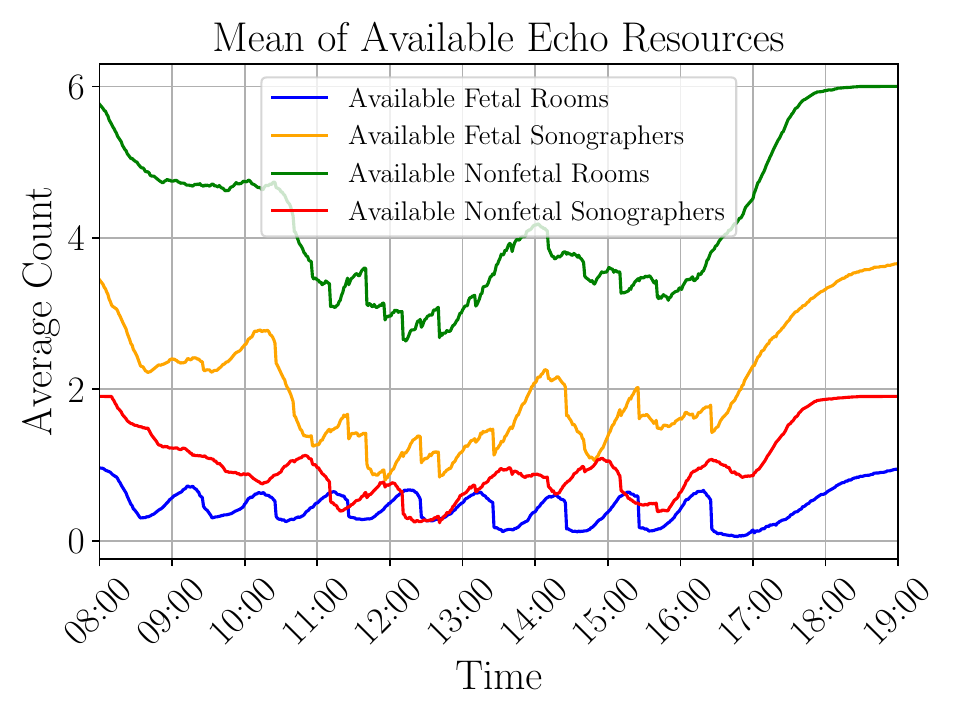}
\hfill
\includegraphics[width=0.32\textwidth]{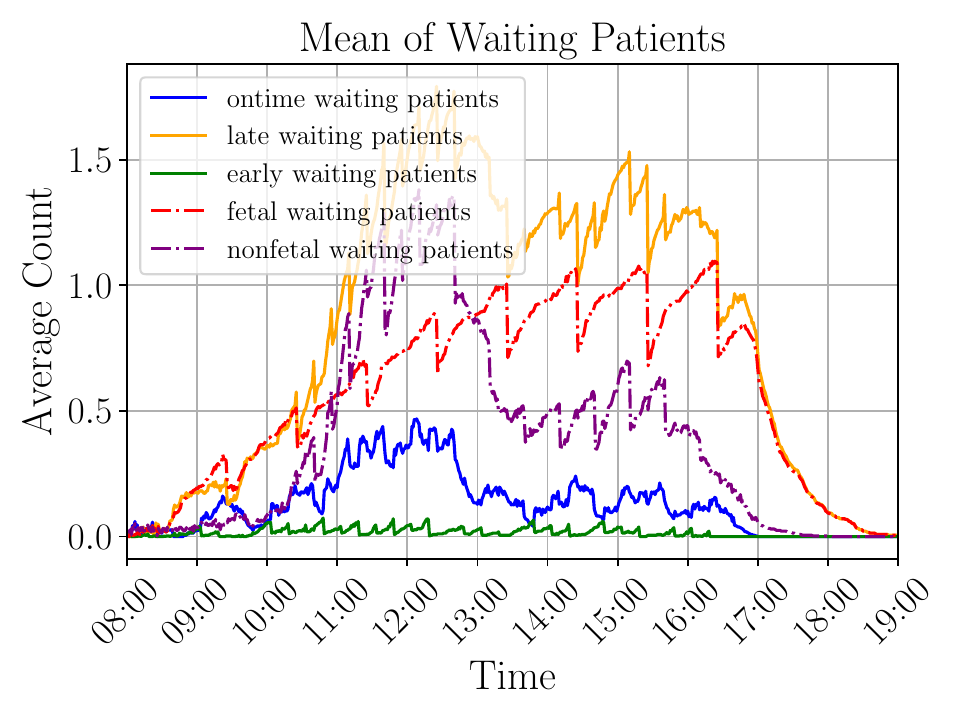}
\hfill
\includegraphics[width=0.32\textwidth]{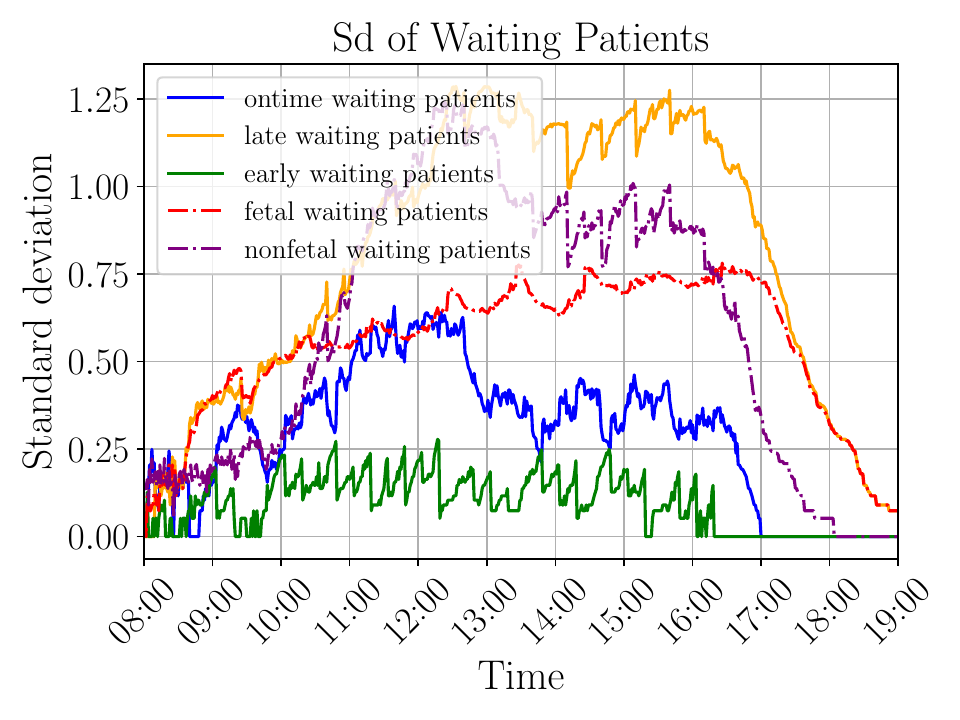}
\caption{Average results in terms of echo resources and waiting patients over 365 independent replications of a representative day for the stochastic hospital process enforcing policy 5 with $\alpha = 100\%$ and $\beta = 100\%$}\label{fig:res_Policy5 100 100}
\end{figure}
\subsubsection{Policy 6 ($\alpha, \beta$)}\label{policy 6}
As in the previous subsection, we report average echo-resource utilization and waiting-patient counts for policy 6 across all tested $(\alpha,\beta)$ reservation fractions; Figure~\ref{fig:res_Policy6 1 1_main} above (Section~\ref{sec:predefined_policies}) duplicates the $\alpha=100\%,\beta=100\%$ case shown here as a convenience for the main-text discussion.
\begin{figure}[ht!]
\centering
\includegraphics[width=0.32\textwidth]{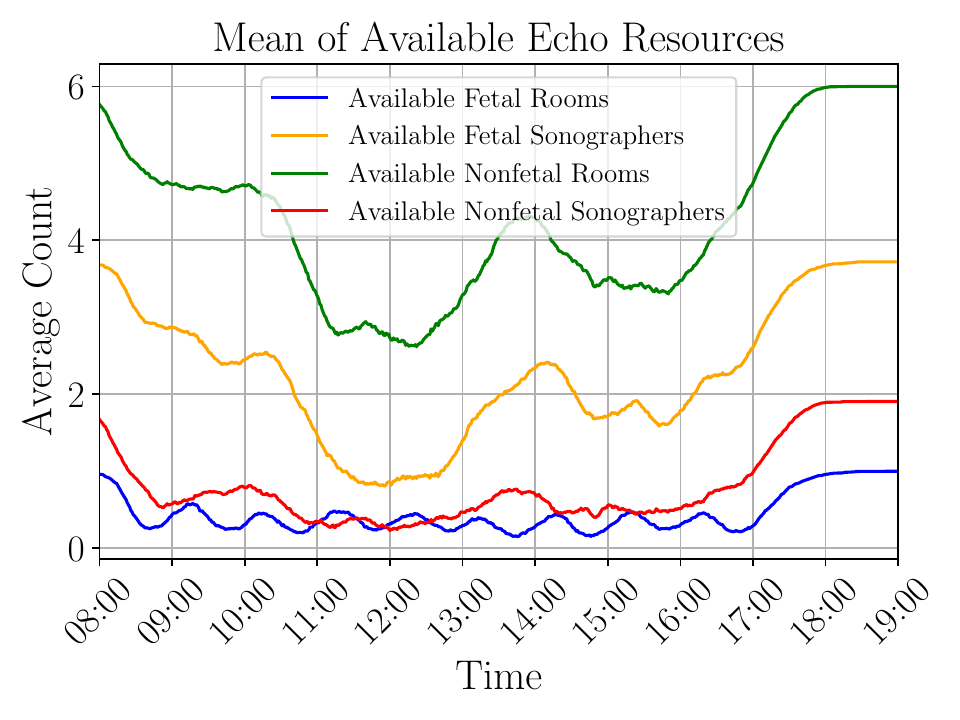}
\hfill
\includegraphics[width=0.32\textwidth]{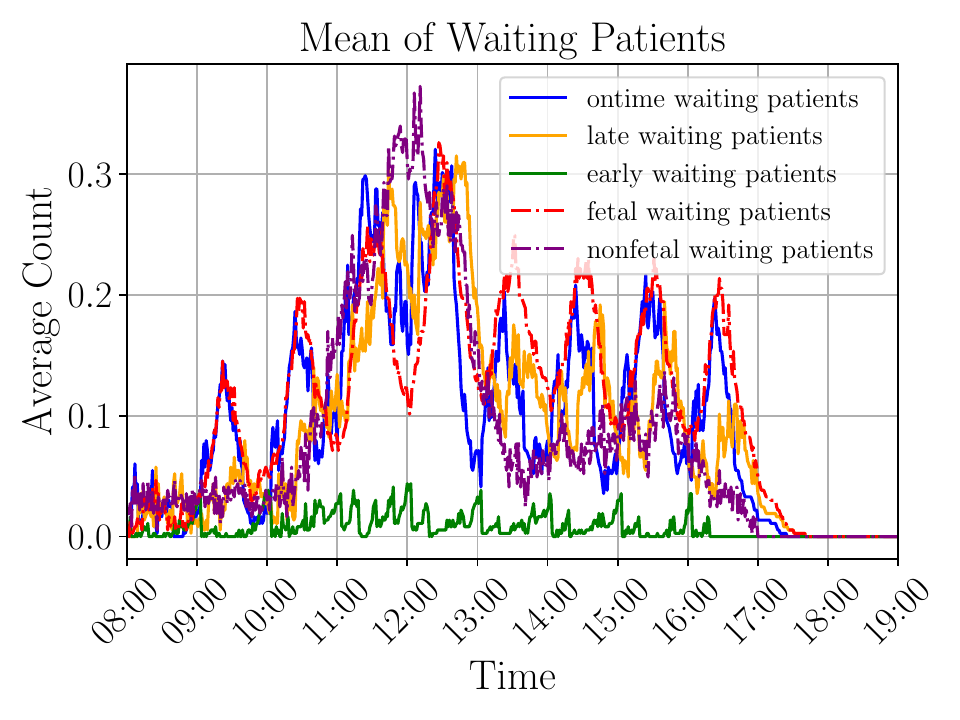}
\hfill
\includegraphics[width=0.32\textwidth]{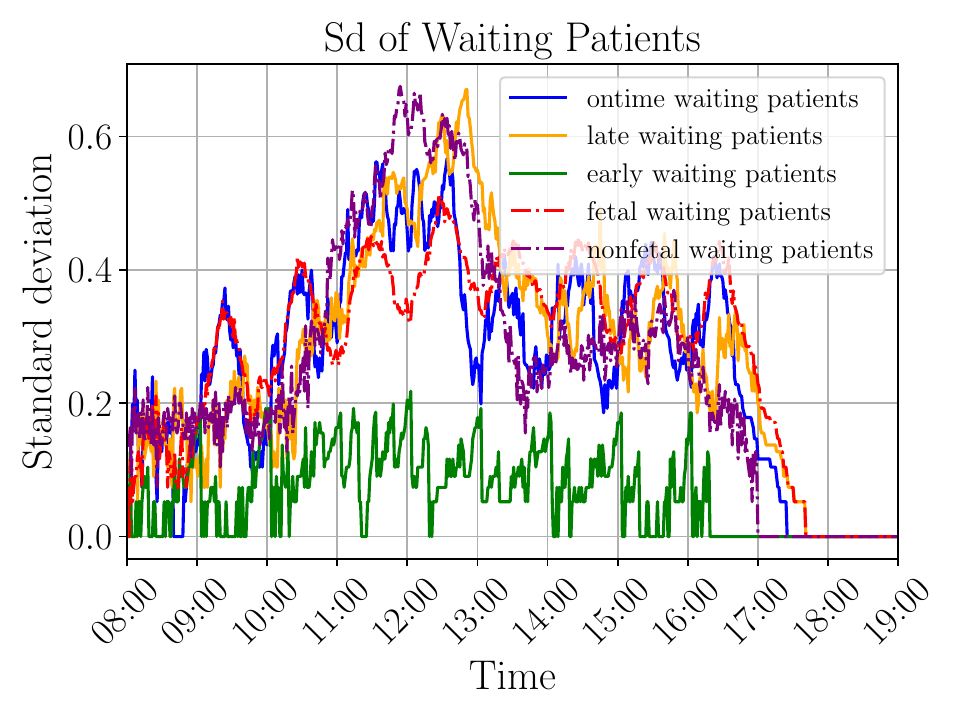}
\caption{Average results in terms of echo resources and waiting patients over 365 independent replications of a representative day for the stochastic hospital process enforcing policy 6 with $\alpha = 0\%$ and $\beta = 0\%$.}\label{fig:res_Policy6 0 0}
\end{figure}
\begin{figure}[ht!]
\centering
\includegraphics[width=0.32\textwidth]{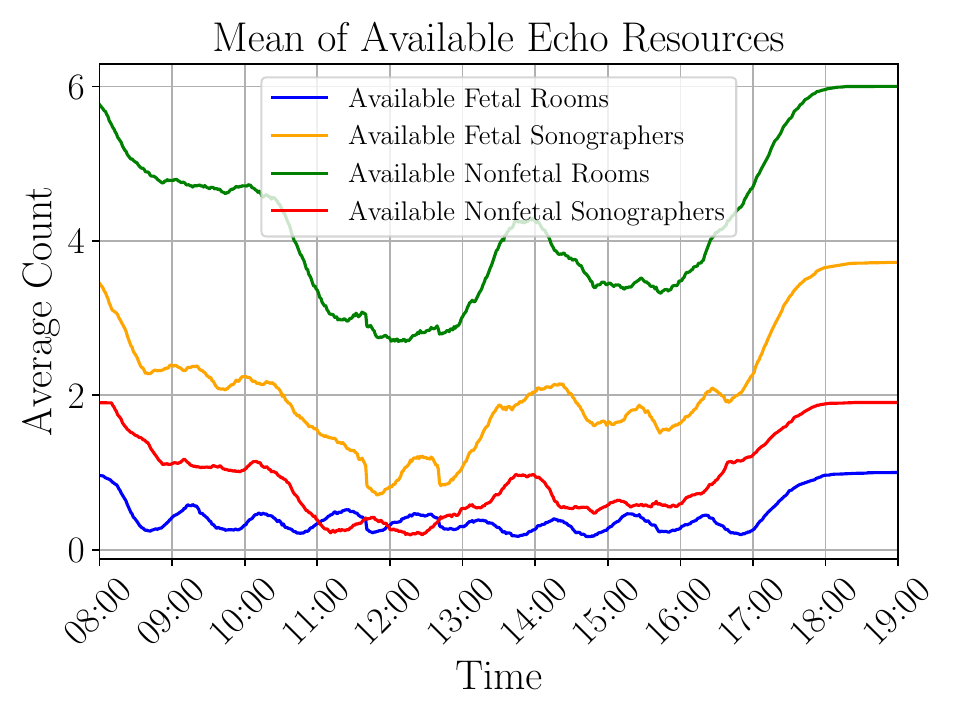}
\hfill
\includegraphics[width=0.32\textwidth]{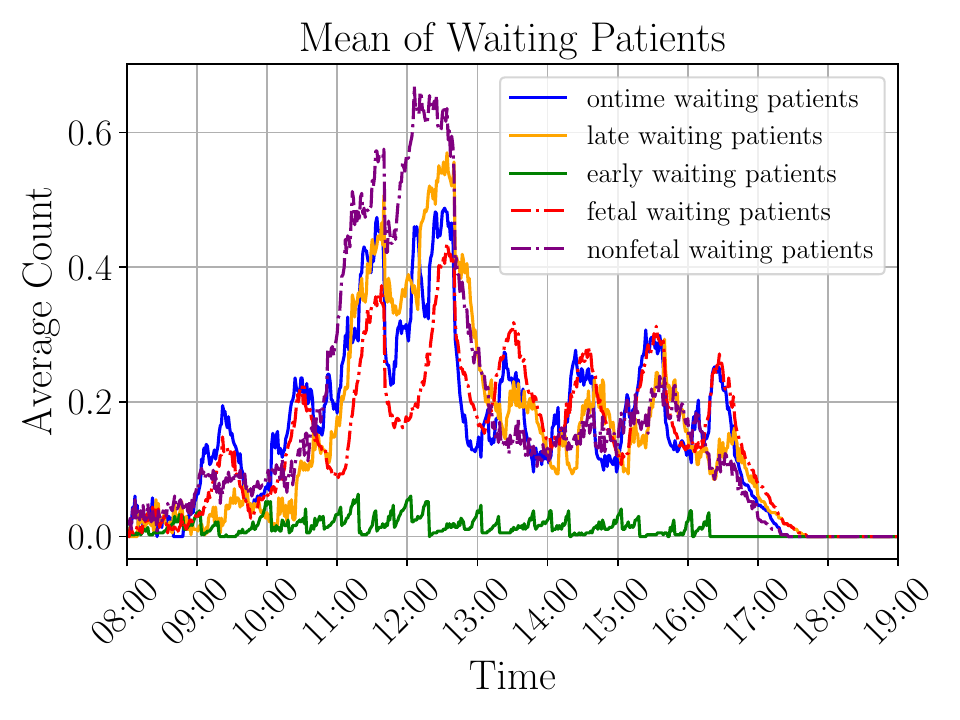}
\hfill
\includegraphics[width=0.32\textwidth]{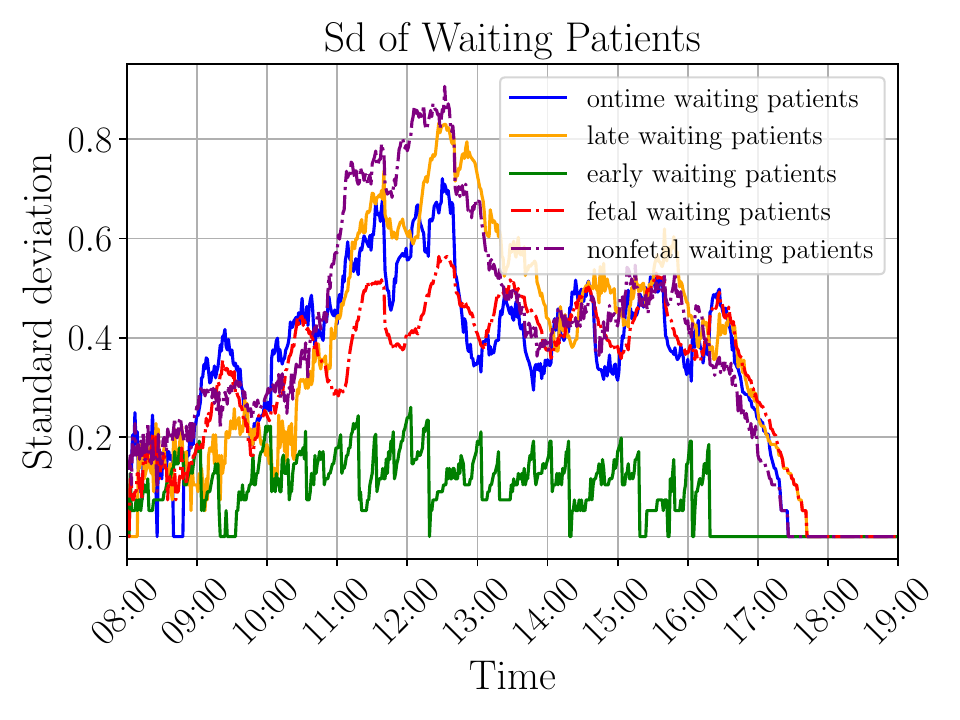}
\caption{Average results in terms of echo resources and waiting patients over 365 independent replications of a representative day for the stochastic hospital process enforcing policy 6 with $\alpha = 0\%$ and $\beta = 25\%$}\label{fig:res_Policy6 0 25}
\end{figure}
\begin{figure}[ht!]
\centering
\includegraphics[width=0.32\textwidth]{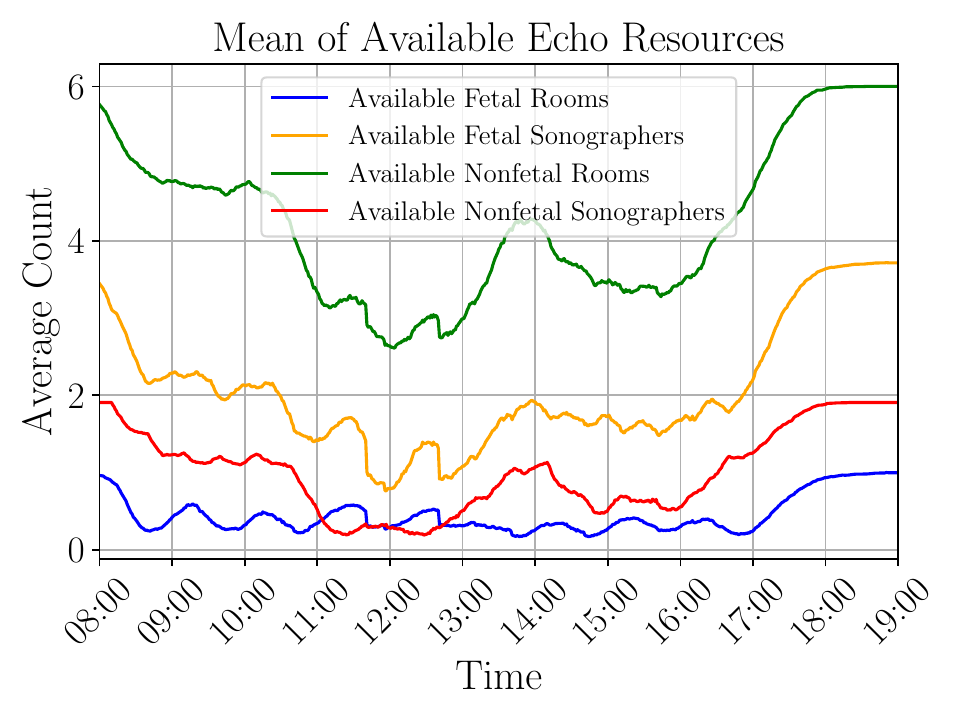}
\hfill
\includegraphics[width=0.32\textwidth]{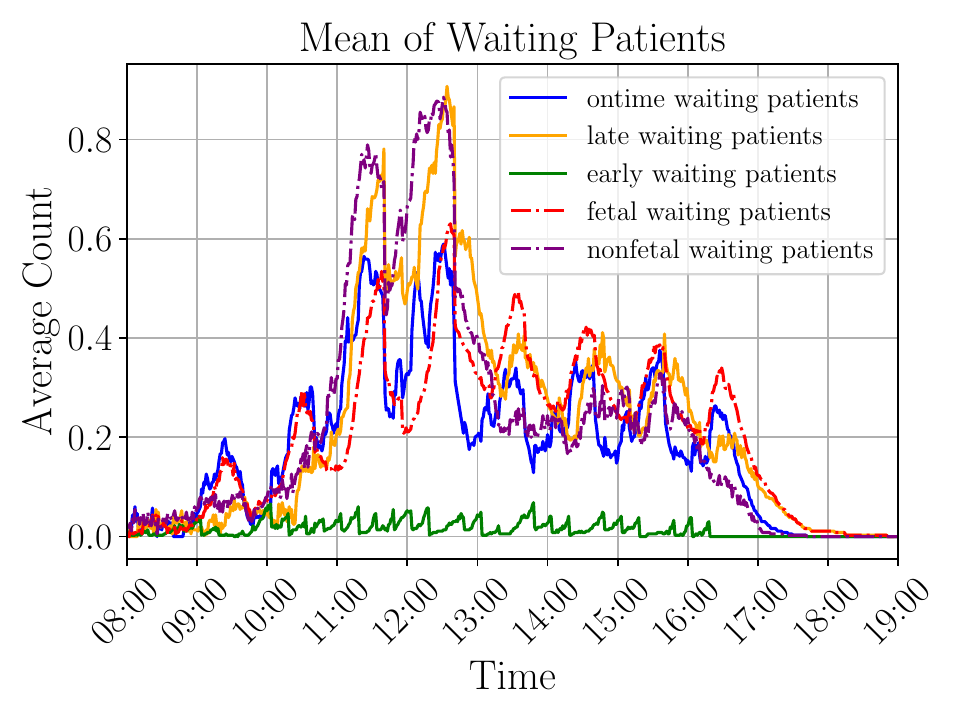}
\hfill
\includegraphics[width=0.32\textwidth]{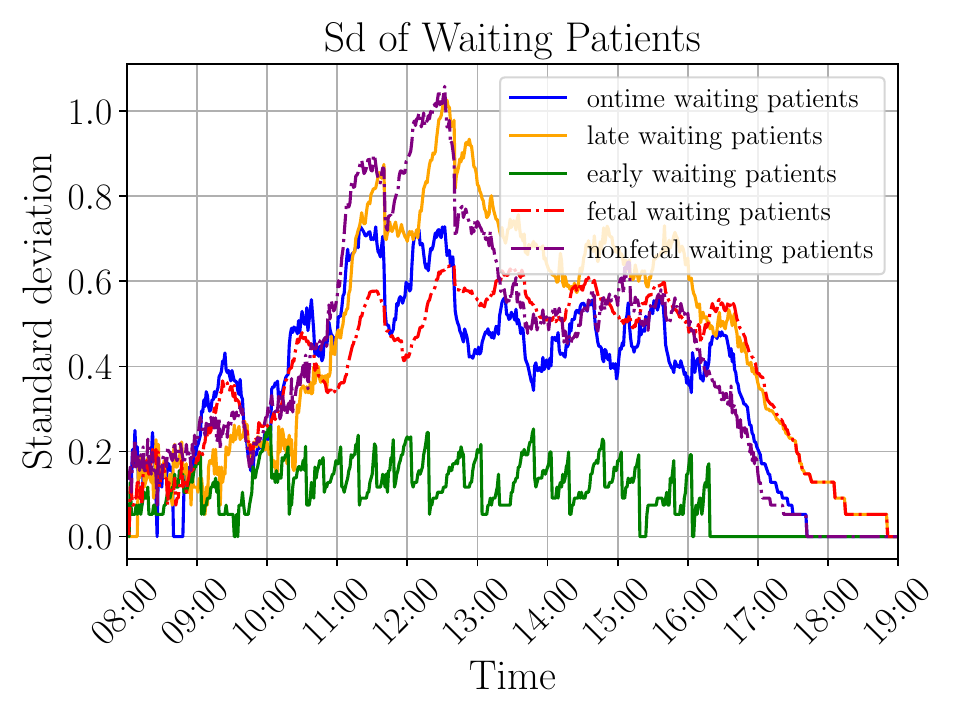}
\caption{Average results in terms of echo resources and waiting patients over 365 independent replications of a representative day for the stochastic hospital process enforcing policy 6 with $\alpha = 0\%$ and $\beta = 50\%$}\label{fig:res_Policy6 0 50}
\end{figure}
\begin{figure}[ht!]
\centering
\includegraphics[width=0.32\textwidth]{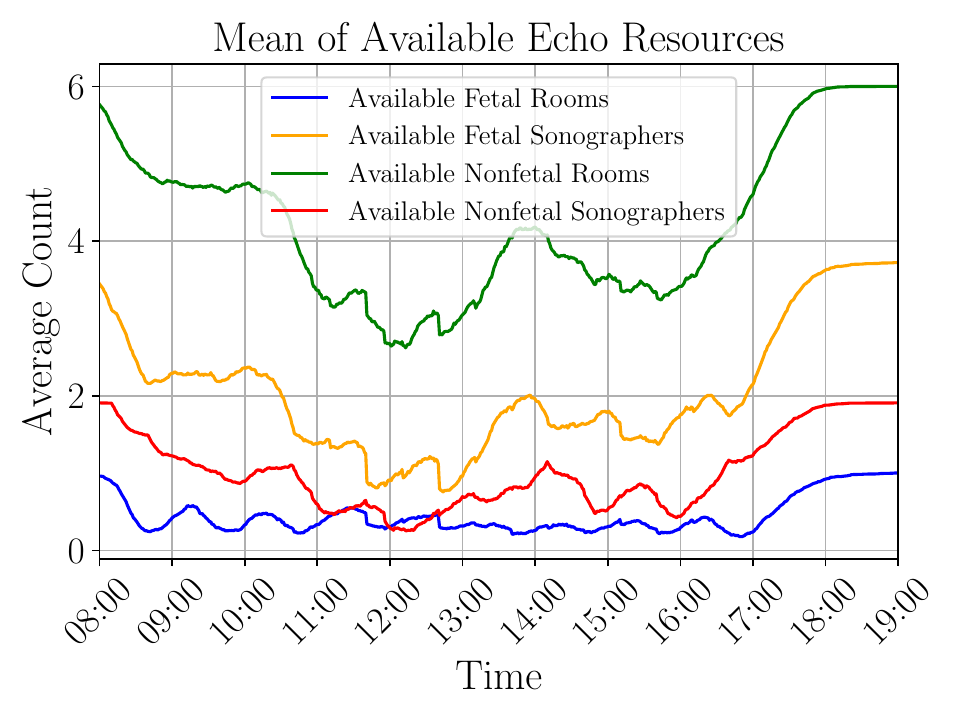}
\hfill
\includegraphics[width=0.32\textwidth]{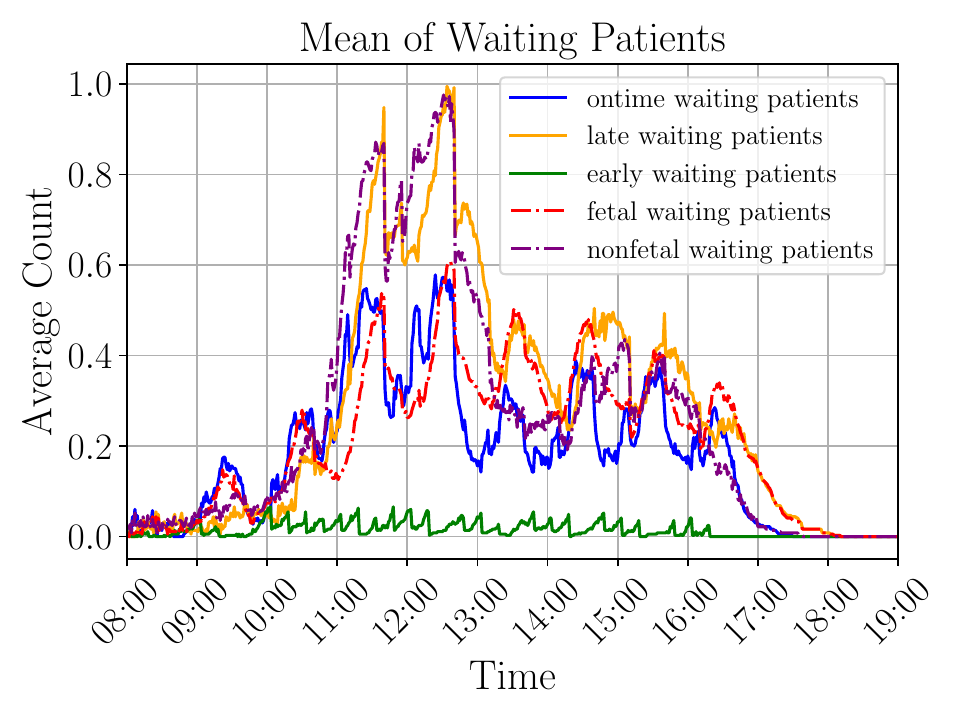}
\hfill
\includegraphics[width=0.32\textwidth]{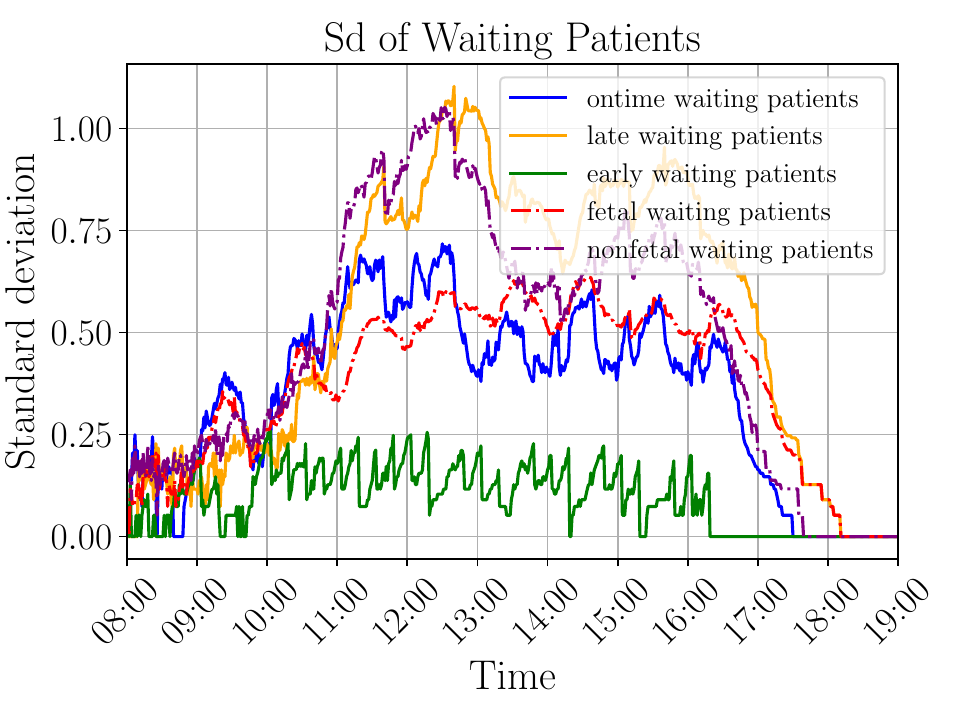}
\caption{Average results in terms of echo resources and waiting patients over 365 independent replications of a representative day for the stochastic hospital process enforcing policy 6 with $\alpha = 0\%$ and $\beta = 75\%$}\label{fig:res_Policy6 0 75}
\end{figure}
\begin{figure}[ht!]
\centering
\includegraphics[width=0.32\textwidth]{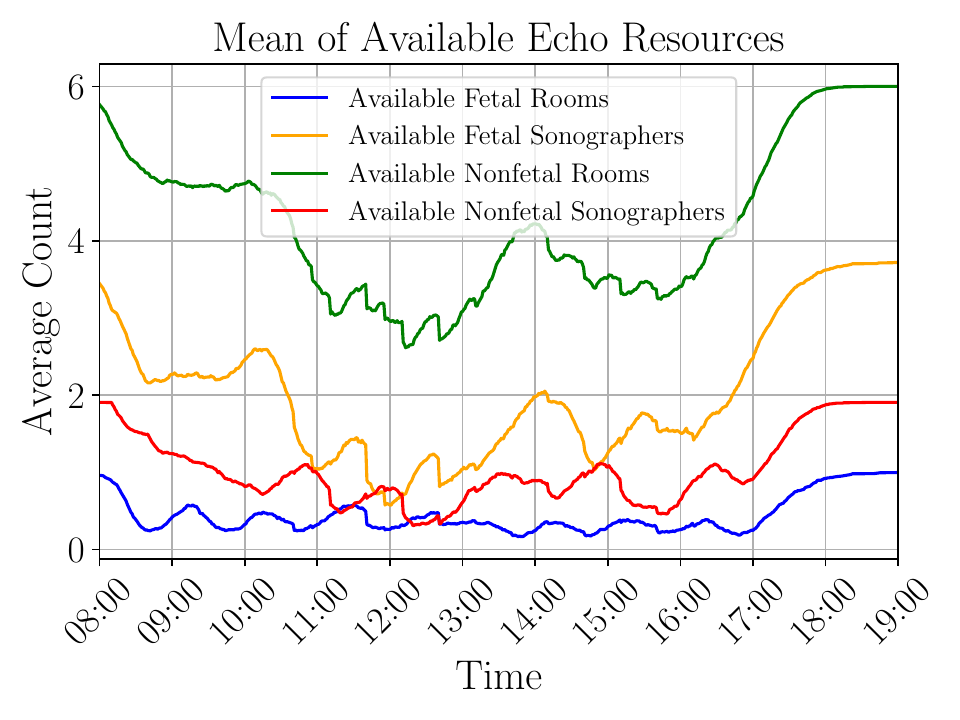}
\hfill
\includegraphics[width=0.32\textwidth]{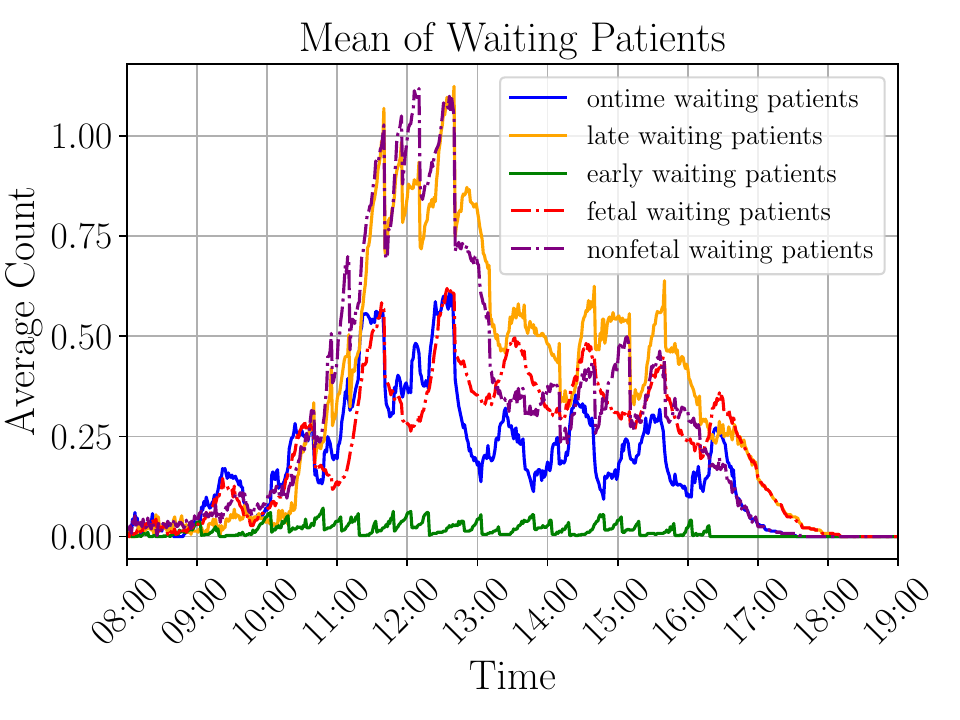}
\hfill
\includegraphics[width=0.32\textwidth]{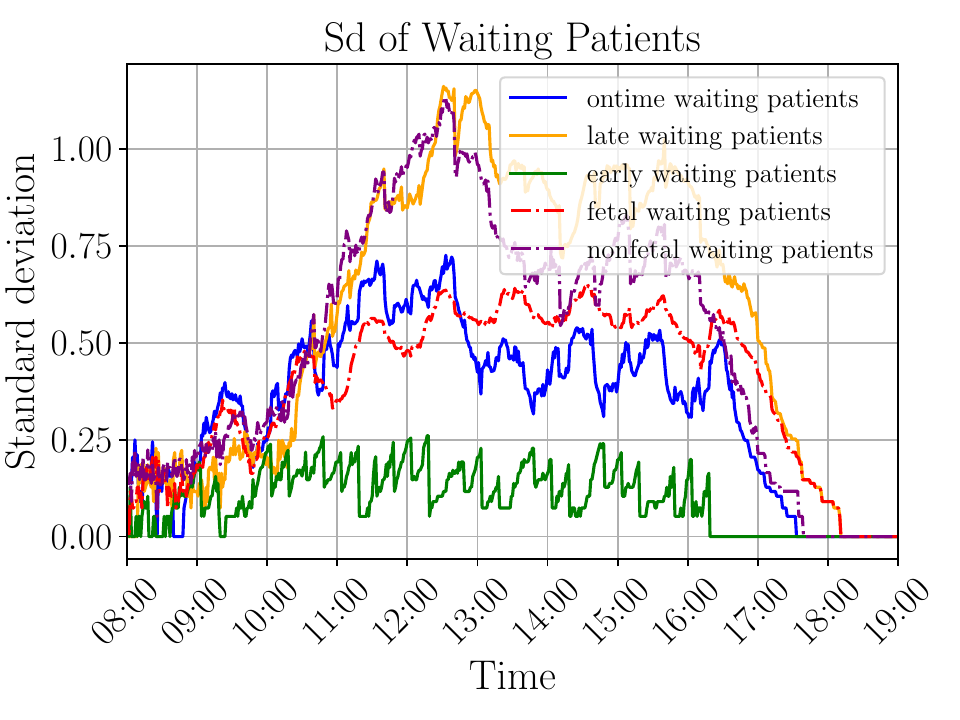}
\caption{Average results in terms of echo resources and waiting patients over 365 independent replications of a representative day for the stochastic hospital process enforcing policy 6 with $\alpha = 0\%$ and $\beta = 100\%$}\label{fig:res_Policy6 0 100}
\end{figure}
\begin{figure}[ht!]
\centering
\includegraphics[width=0.32\textwidth]{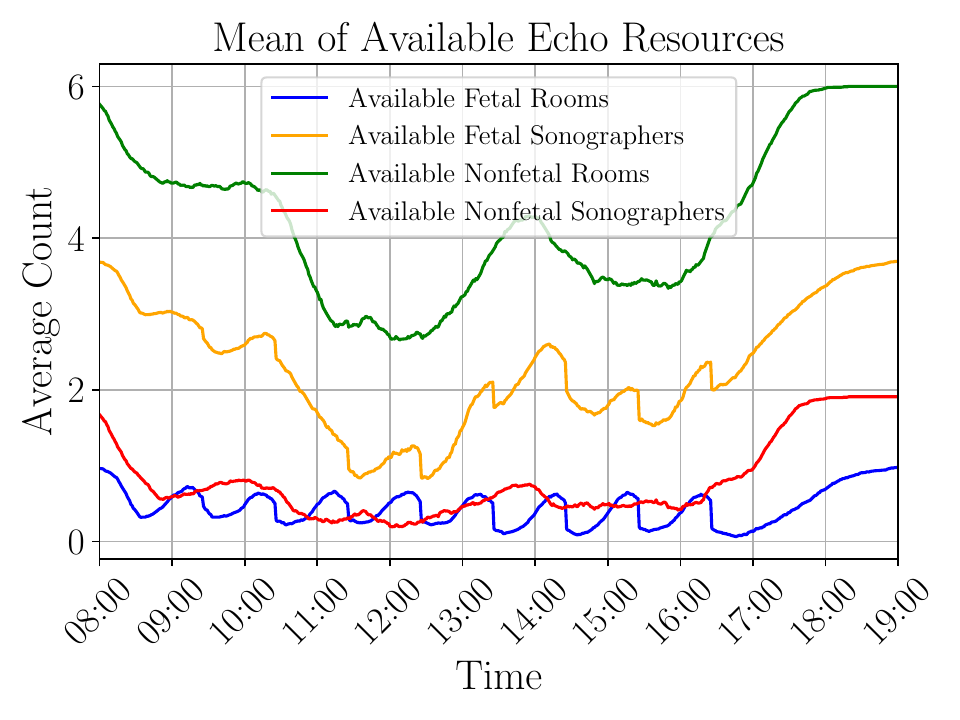}
\hfill
\includegraphics[width=0.32\textwidth]{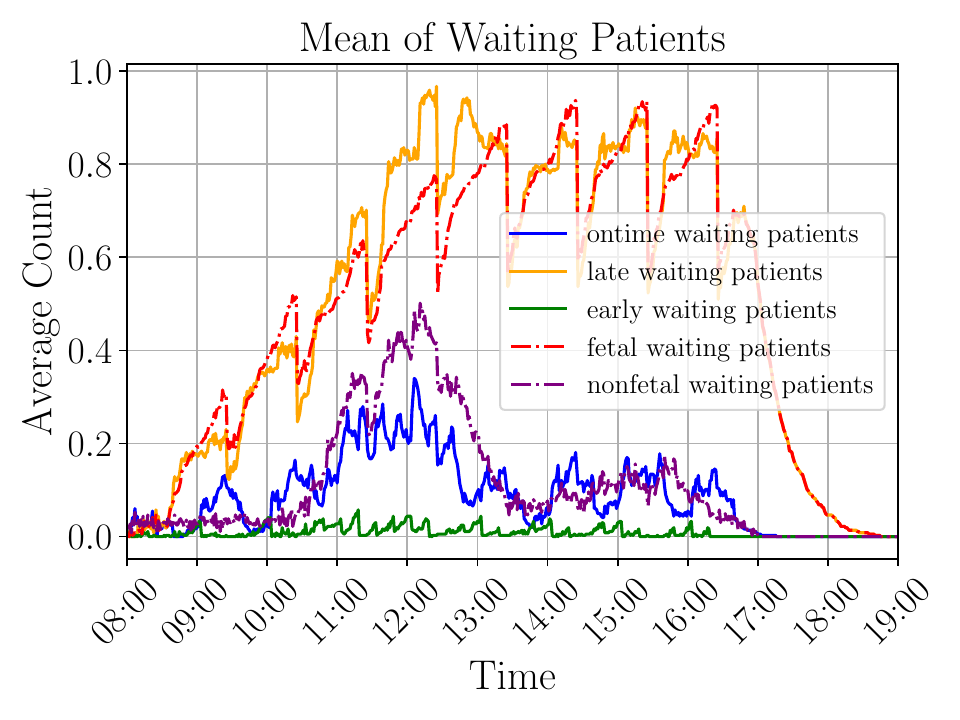}
\hfill
\includegraphics[width=0.32\textwidth]{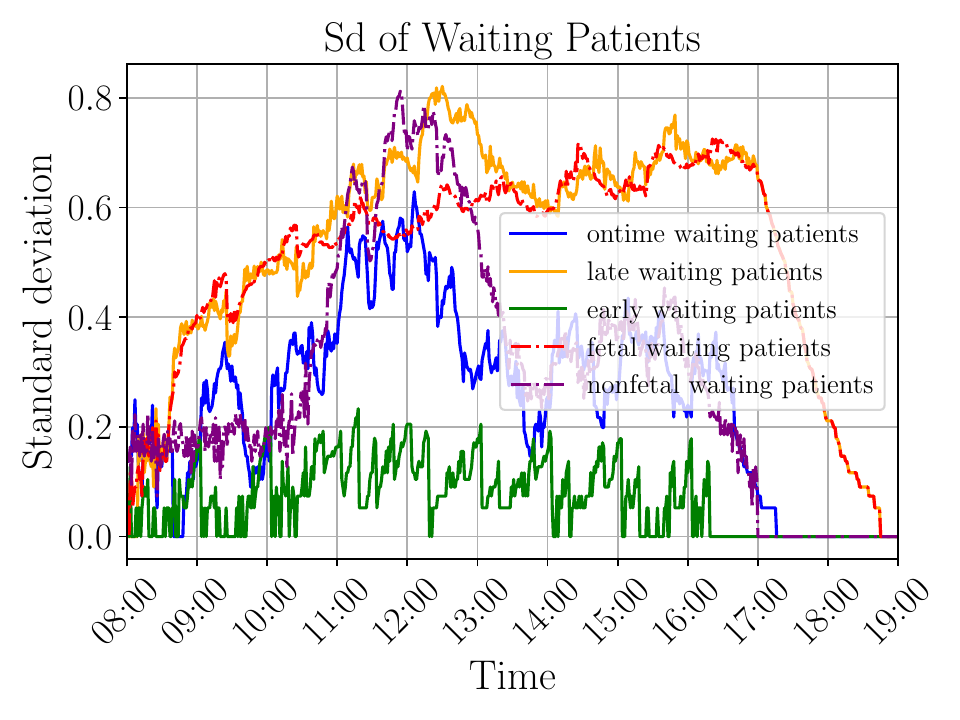}
\caption{Average results in terms of echo resources and waiting patients over 365 independent replications of a representative day for the stochastic hospital process enforcing policy 6 with $\alpha = 100\%$ and $\beta = 0\%$}\label{fig:res_Policy6 100 0}
\end{figure}
\begin{figure}[ht!]
\centering
\includegraphics[width=0.32\textwidth]{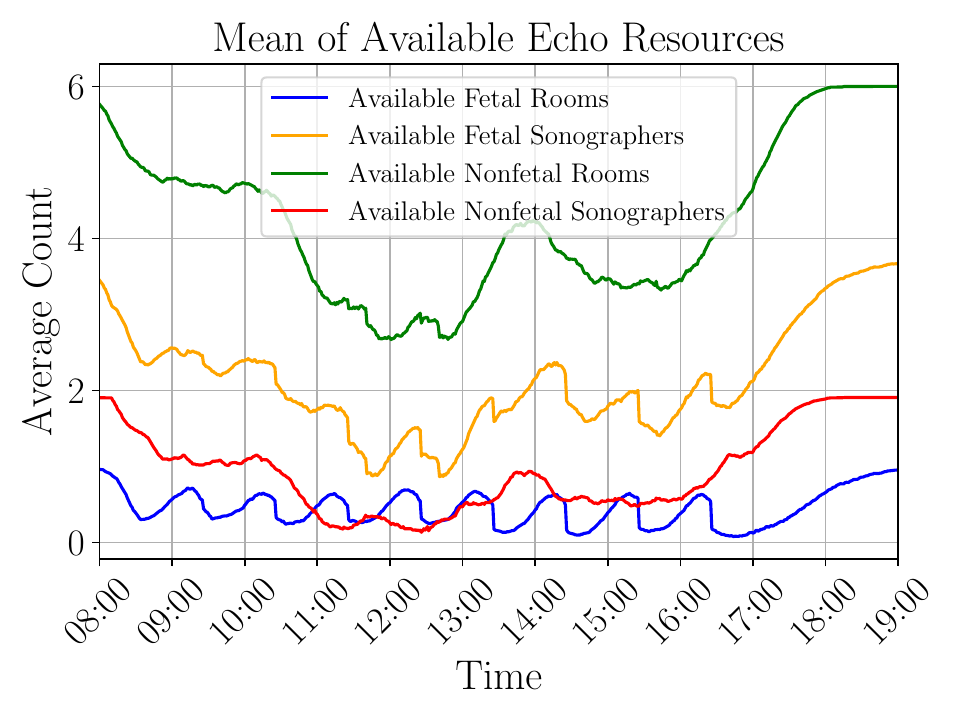}
\hfill
\includegraphics[width=0.32\textwidth]{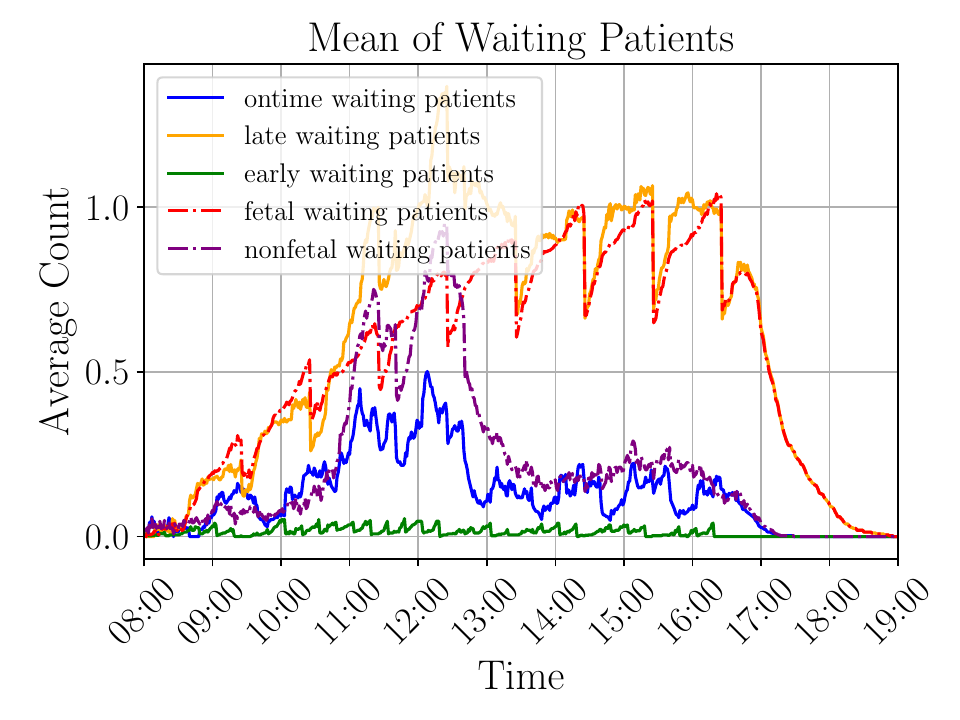}
\hfill
\includegraphics[width=0.32\textwidth]{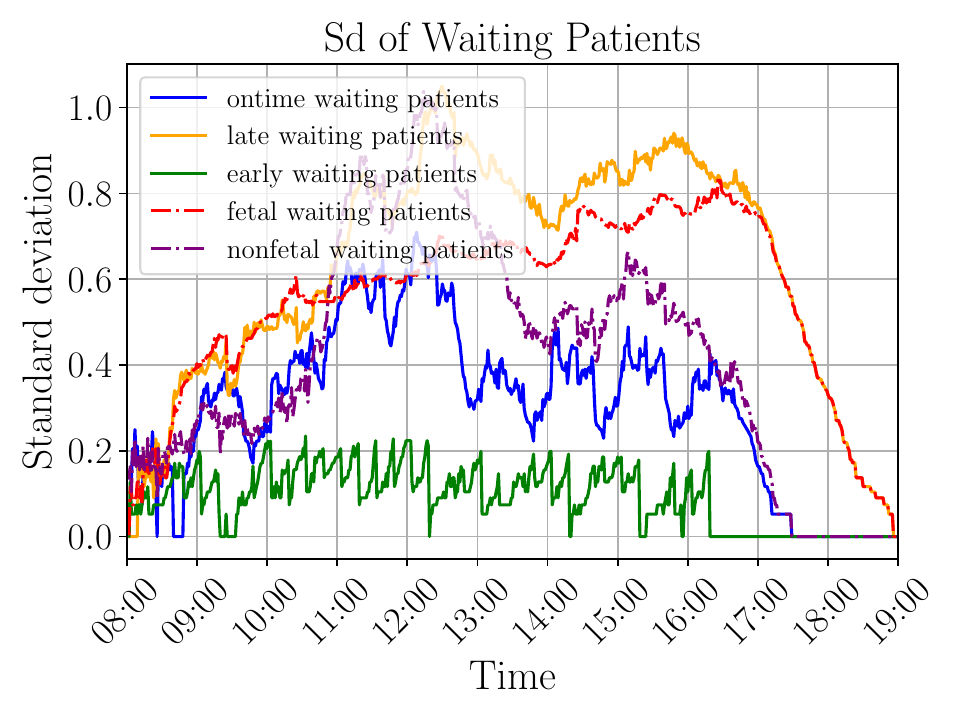}
\caption{Average results in terms of echo resources and waiting patients over 365 independent replications of a representative day for the stochastic hospital process enforcing policy 6 with $\alpha = 100\%$ and $\beta = 25\%$}\label{fig:res_Policy6 100 25}
\end{figure}

\begin{figure}[ht!]
\centering
\includegraphics[width=0.32\textwidth]{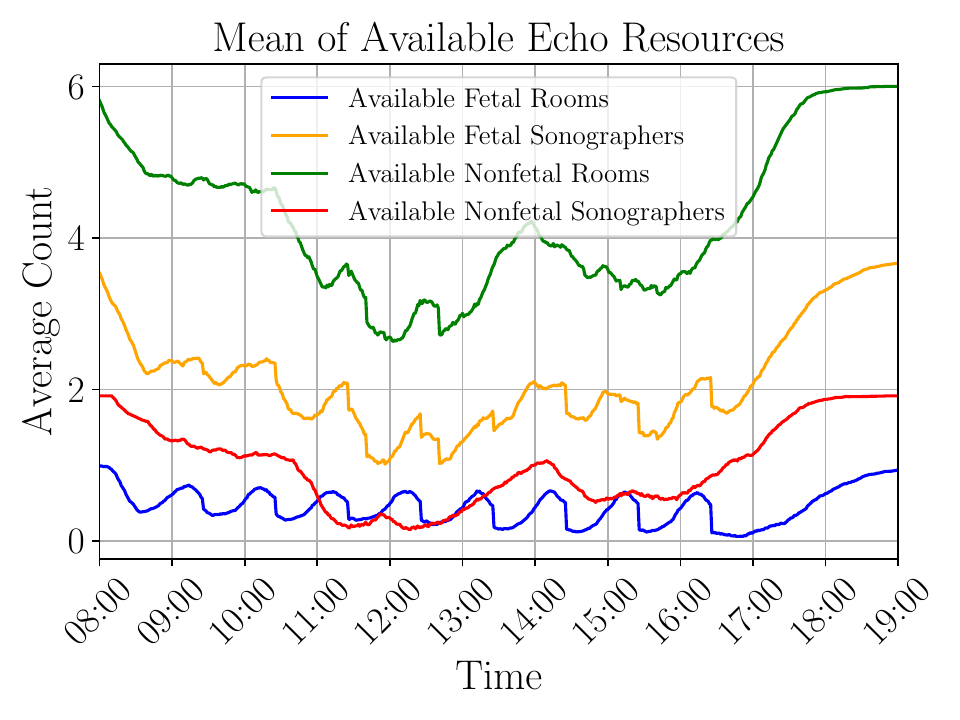}
\hfill
\includegraphics[width=0.32\textwidth]{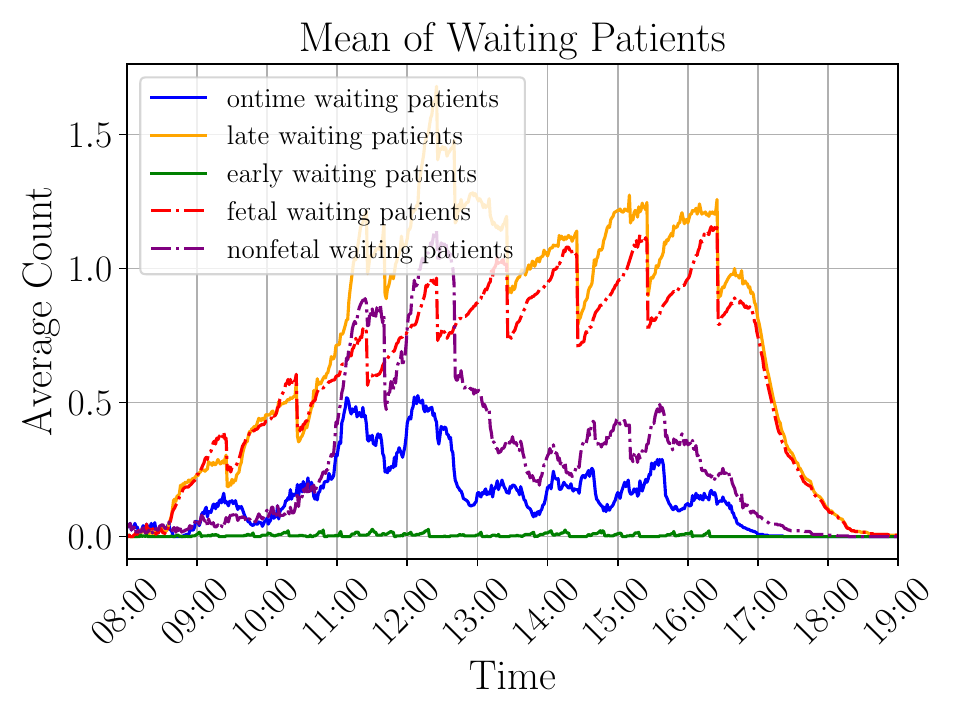}
\hfill
\includegraphics[width=0.32\textwidth]{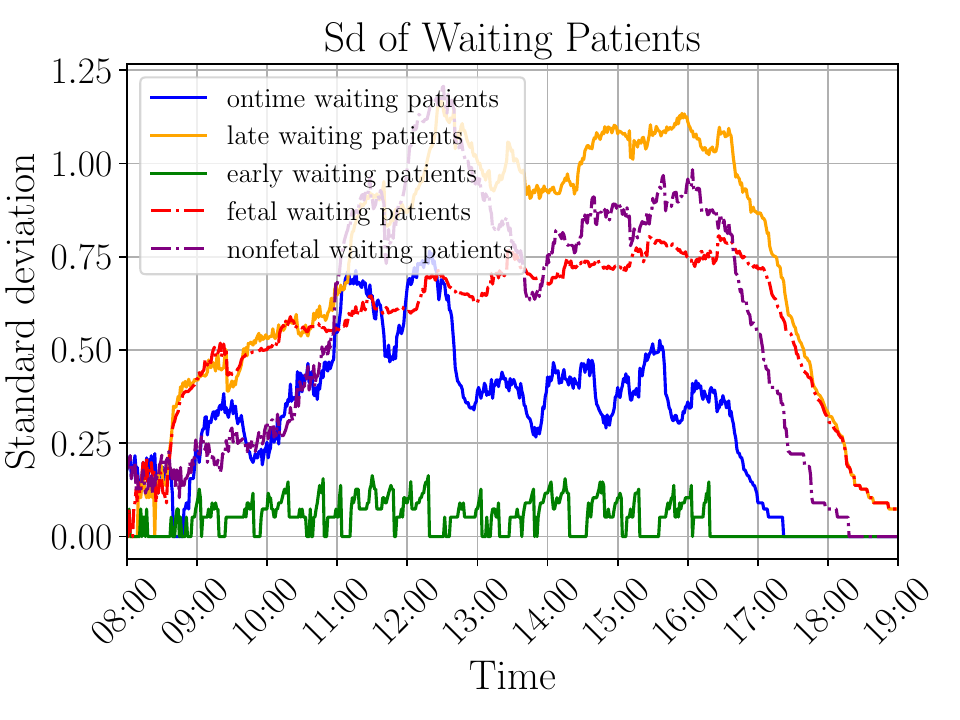}
\caption{Average results in terms of echo resources and waiting patients over 365 independent replications of a representative day for the stochastic hospital process enforcing policy 6 with $\alpha = 100\%$ and $\beta = 50\%$}\label{fig:res_Policy6 100 50}
\end{figure}

\begin{figure}[ht!]
\centering
\includegraphics[width=0.32\textwidth]{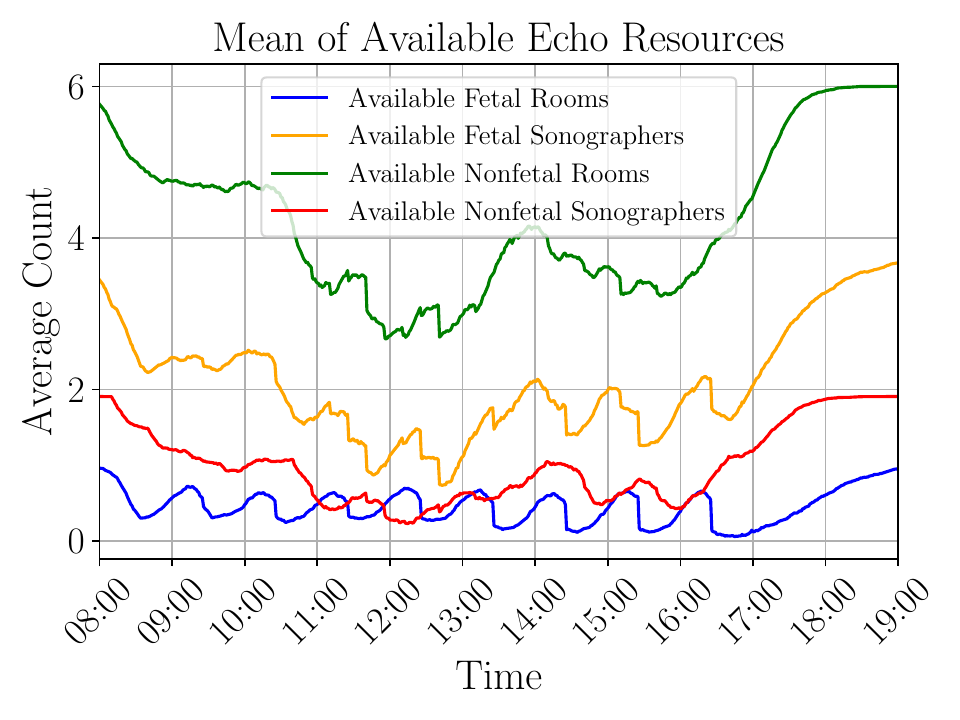}
\hfill
\includegraphics[width=0.32\textwidth]{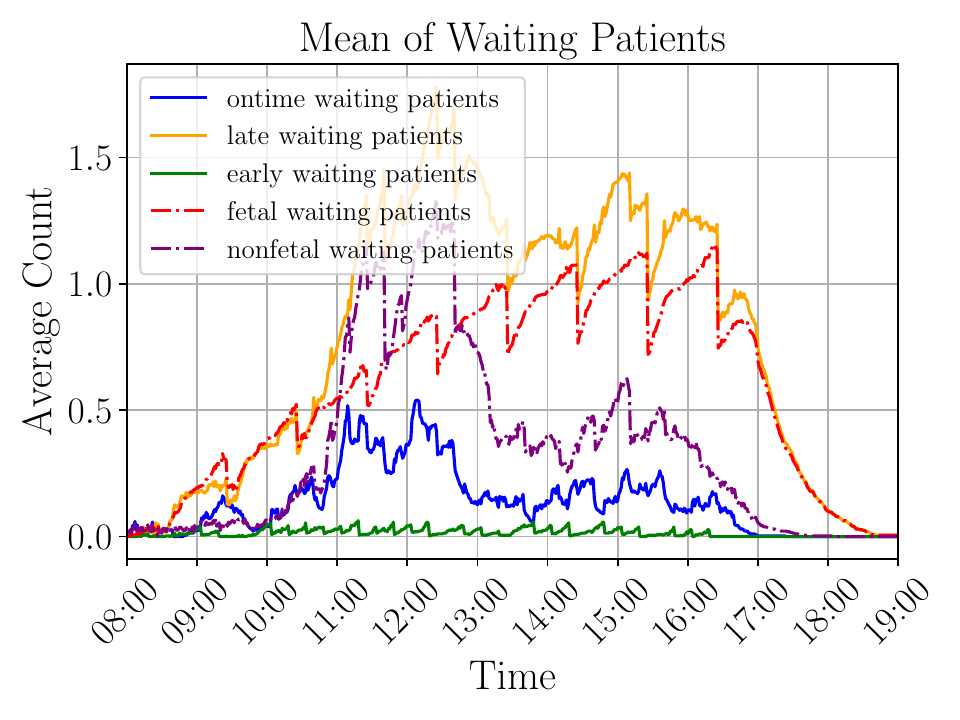}
\hfill
\includegraphics[width=0.32\textwidth]{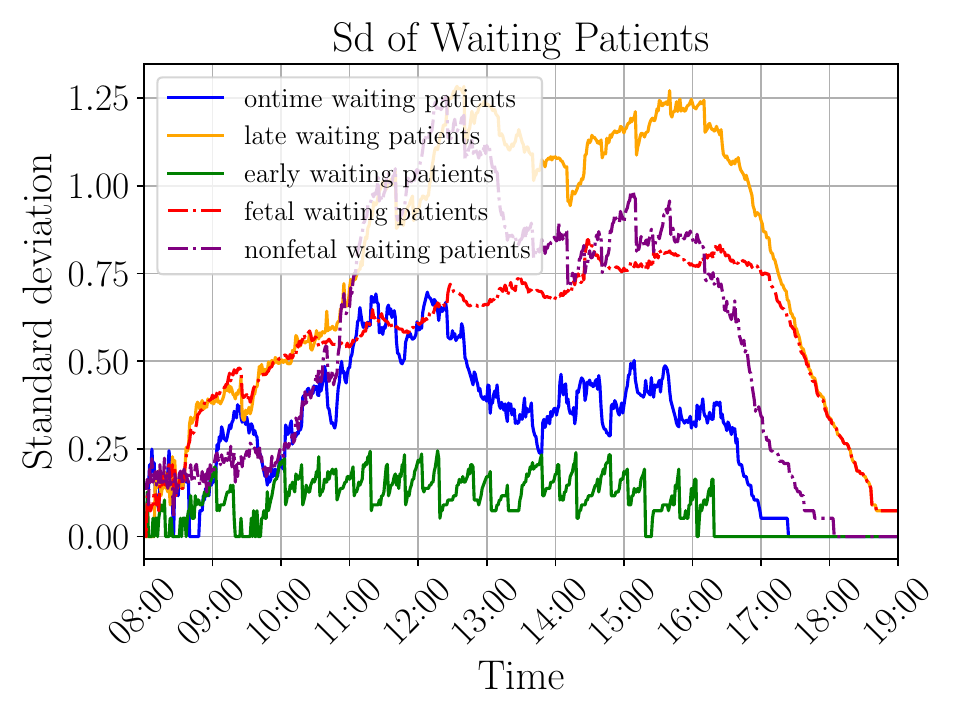}
\caption{Average results in terms of echo resources and waiting patients over 365 independent replications of a representative day for the stochastic hospital process enforcing policy 6 with $\alpha = 100\%$ and $\beta = 75\%$}\label{fig:res_Policy6 100 75}
\end{figure}

\begin{figure}[ht!]
\centering
\includegraphics[width=0.32\textwidth]{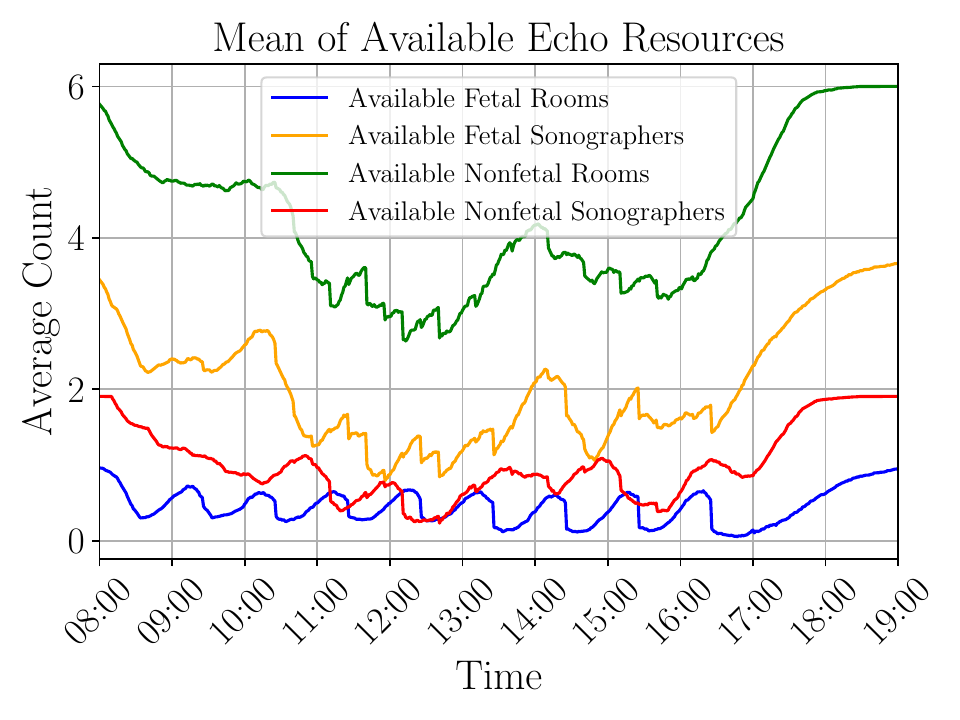}
\hfill
\includegraphics[width=0.32\textwidth]{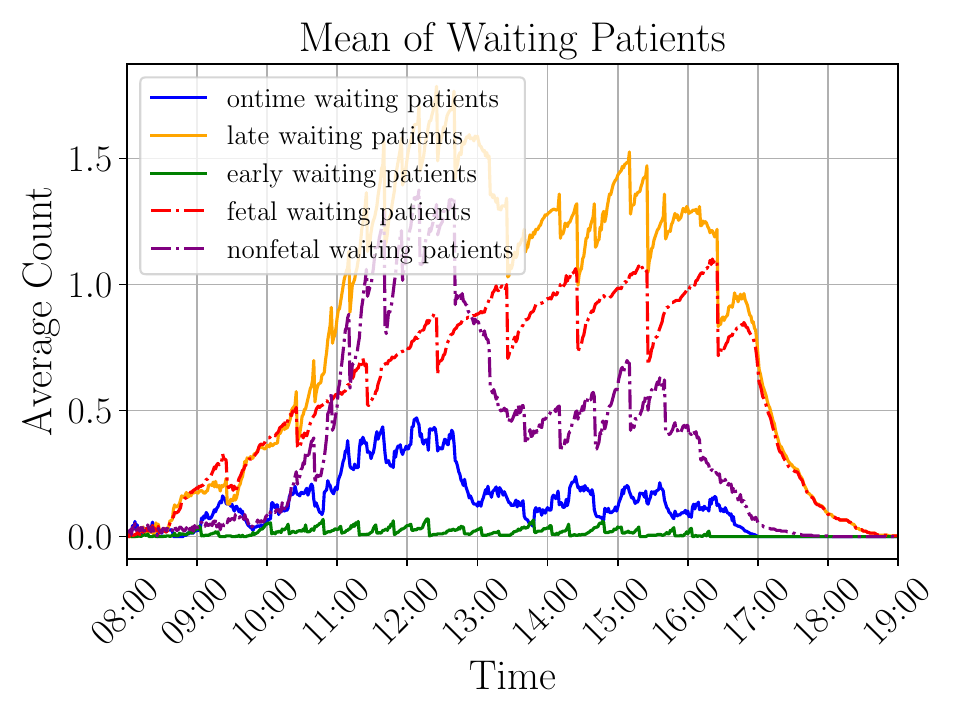}
\hfill
\includegraphics[width=0.32\textwidth]{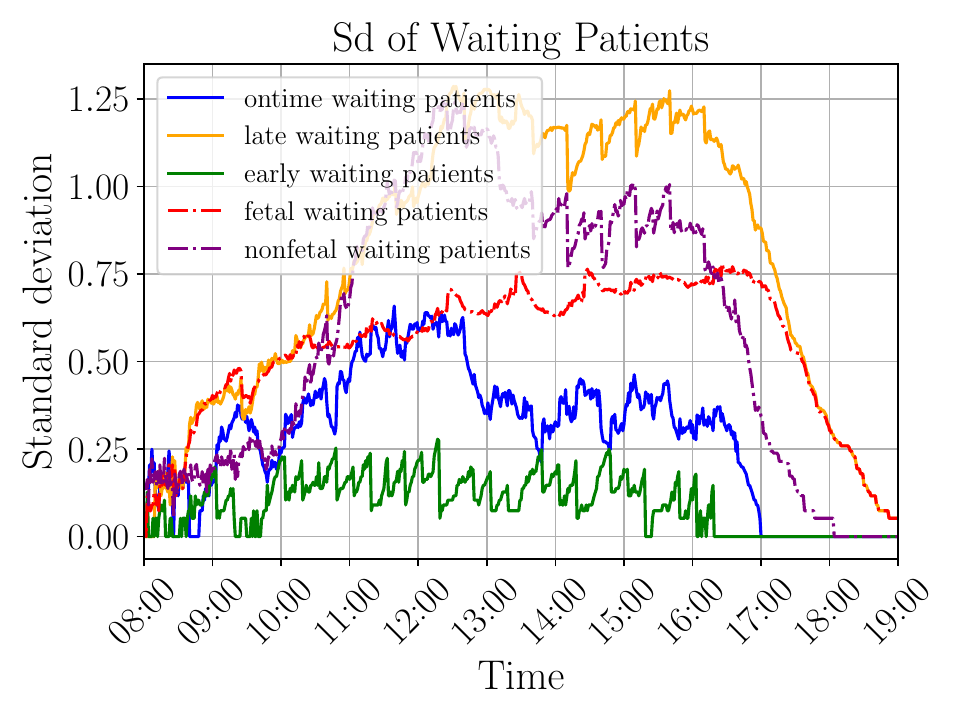}
\caption{Average results in terms of echo resources and waiting patients over 365 independent replications of a representative day for the stochastic hospital process enforcing policy 6 with $\alpha = 100\%$ and $\beta = 100\%$}\label{fig:res_Policy6 1 1}
\end{figure}


\subsection{Training strategy and hyperparameter selection}\label{sec:Policy-training}
Table~\ref{tab:hyperparams} summarizes the hyperparameters used to train the RL policy.
\begin{table}[ht]
\centering
\begin{tabular}{l p{10.5cm}}
\toprule
\textbf{Hyperparameter} & \textbf{Value} \\
\midrule
Network architecture & Dueling DQN; shared trunk $13\to128\to128$; value stream $128\to64\to1$; advantage stream $128\to64\to729$ \\
Normalization / activation & Layer Normalization; ReLU \\
Weight initialization & Kaiming (He et al.) \\
Optimizer & Adam (default PyTorch betas $(0.9,0.999)$, $\epsilon=10^{-8}$) \\
Learning rate & Initial $0.001$; $\times 0.95$ after 500 consecutive steps without loss improvement; floor $0.00001$ \\
Discount factor $\gamma$ & $0.99$ \\
Soft target-update rate $\tau$ & $0.005$ \\
Loss function & Smooth L1 (Huber) loss \\
Gradient-clipping threshold & $1.0$ (L2 norm) \\
Dropout rate & $0.2$ (shared feature extractor only) \\
Replay buffer capacity & $100{,}000$ transitions \\
Batch size & $64$ \\
Training start threshold & Buffer $> 640$ transitions (10 minibatches) \\
PER $\alpha_{\text{PER}}$ & $0.6$ \\
PER $\beta_{\text{PER}}$ & Initial $0.4$; annealed $+10^{-4}$ per step to $1.0$ \\
PER priority clamp & $[0.01,\,10.0]$ \\
Exploration ($\epsilon$-greedy) & $\epsilon_{\max}=1.0$, $\epsilon_{\min}=0.01$, decay $=0.9995$/episode \\
Training episodes & $30{,}000$ \\
Random seed & Fixed seed $42$ (Python/NumPy/PyTorch/CUDA); environment reset re-seeded per episode with the episode index \\
Hyperparameter search & None performed; values set manually \\
Hardware / wall-clock time & Notre Dame CRC cluster; exact wall-clock time not recorded \\
Software versions & Python, PyTorch, SimPy, Gymnasium, SciPy; exact version numbers not recorded \\
\bottomrule
\end{tabular}
\caption{Hyperparameters used to train the RL policy.}
\label{tab:hyperparams}
\end{table}
\subsubsection{Adaptive learning rate and weight initialization}

We adopt a scheduler that reduces the learning rate by $5\%$ every time a plateau is detected in the loss function.
The initial learning rate is set to $0.001$ and is bounded by a minimum learning rate of $0.00001$ (\texttt{LEARNING\_RATE\_MIN}) to prevent the learning rate from becoming excessively small.

During training, the \texttt{scheduler.step(loss)} function is called within the optimization process after each step. If the loss does not decrease for 500 consecutive steps, indicating a plateau in performance, the scheduler automatically reduces the current learning rate by multiplying it by $0.95$.
This adaptive approach allows the model to navigate different phases of training effectively.
Finally, we use Kaiming initialization for the neural network weights~\cite{he2015delving}.

\subsubsection{Batch learning using prioritized memory}

Experience replay with uniform sampling can be inefficient for learning because it treats all transitions equally, regardless of their potential learning value. This means rare but important experiences (such as discovering a new strategy or encountering a critical failure) have the same probability of being selected as common, well-understood transitions. 
In~\cite{schaul2015prioritized}, a replay-sampling scheme called \emph{Prioritized Experience Replay} (PER) addresses this limitation by sampling transitions with probability proportional to their temporal-difference (TD) error, which reflects how \emph{surprising} or \emph{informative} a transition is for the current model. 
When an agent encounters a transition with a large TD error, it suggests that the current value estimates are inaccurate for that situation, making it a valuable learning opportunity. By prioritizing these high-error transitions during sampling, PER ensures the agent spends more time learning from experiences where its predictions were poor, leading to faster learning and improved sample efficiency, especially in environments where critical experiences are rare.

Formally, our PER implementation assigns sampling probabilities according to transition priorities:
\begin{equation}
P(i) = \frac{(p_i + \epsilon)^{\alpha_{\text{PER}}}}{\sum_k (p_k + \epsilon)^{\alpha_{\text{PER}}}},
\end{equation}
where $p_i$ is the priority of transition $i$, $\epsilon = 10^{-6}$ is a small constant to prevent zero probabilities, and $\alpha_{\text{PER}} = 0.6$ controls the strength of prioritization (with $\alpha_{\text{PER}} = 0$ corresponding to uniform sampling). To prevent extreme priorities from dominating the sampling process, we clamp $p_i$ between \texttt{MIN\_PRIORITY} $= 0.01$ and \texttt{MAX\_PRIORITY} $= 10.0$. However, this non-uniform sampling introduces bias into the value updates because the replay distribution no longer matches the on-policy distribution. To correct for this bias, we use importance-sampling weights:
\begin{equation}
w_i = \frac{(N \cdot P(i))^{-\beta_{\text{PER}}}}{\max_j (N\cdot P(j))^{-\beta_{\text{PER}}}},
\end{equation}
normalized by the maximum weight in the batch so that updates remain bounded above, as in Schaul et al.~\cite{schaul2015prioritized}, where $N$ is the current number of transitions stored in the buffer (not its fixed capacity of $100{,}000$, particularly while the buffer is still filling) and $\beta_{\text{PER}}$ controls the degree of bias correction. We initialize $\beta_{\text{PER}} = 0.4$ and gradually anneal it towards $1.0$ over the course of training by increasing it at each sampling step with a small increment of $10^{-4}$. This schedule allows some bias early in training, when aggressive prioritization is most helpful, while progressively restoring unbiased updates as learning converges. The normalization above matches the actual implementation (weights are divided by their batch maximum, and $N$ is the current buffer size, not its fixed capacity). We also confirm the anneal pace: at an increment of $10^{-4}$ per step, $\beta_{\text{PER}}$ reaches 1.0 after 6{,}000 gradient steps, within roughly the first 9 of the 30{,}000 training episodes, after which bias correction remains full for the remainder of training.

The replay memory stores up to $100{,}000$ transitions, and mini-batches of size $64$ are sampled for each optimization step. Training of the online network begins once the buffer contains more than $10$ mini-batches of experience (i.e., more than $640$ transitions), ensuring that early updates are based on a sufficiently diverse set of trajectories. Thereafter, the online network is optimized at every environment step, and the temporal-difference errors from these updates are fed back into the memory to update the corresponding priorities. In parallel, we employ soft target-network updates, where the target network parameters are slowly moved towards those of the online network, ensuring stable targets while still allowing the value function to adapt to new information.

\subsubsection{Exploration-exploitation trade-off}

For exploration, we employ an $\epsilon$-greedy strategy in which the exploration rate $\epsilon$ is initialized to $1.0$ at the beginning of training and decays deterministically after each episode $t$ according to
\[
\epsilon(t)=\max\bigl(\epsilon_{\min},\; \epsilon_{\max}\times \text{decay}^{\,t}\bigr), \qquad \epsilon_{\max}=1.0,\ \epsilon_{\min}=0.01,\ \text{decay}=0.9995.
\]
With this schedule, $\epsilon$ reaches its floor of $0.01$ after approximately $9{,}200$ episodes, i.e., roughly the first third of a typical $30{,}000$-episode training run, after which the policy is effectively greedy aside from a residual $1\%$ random-action rate. The small lower bound ensures that a non-zero level of exploration is maintained throughout learning.

\subsubsection{Gradient clipping}\label{sec:gradient_clip}
Training deep neural networks can be challenging due to the exploding gradient problem, where gradients become extremely large during backpropagation, leading to unstable updates and poor convergence.
Gradient clipping was analyzed and popularized as an effective solution to this issue by~\cite{pascanu2013difficulty}, building on the earlier clipping heuristics used in recurrent-network training.
By imposing a maximum threshold on the gradient norm, clipping prevents excessive parameter updates while maintaining the direction of the gradient.
Specifically, when the L2 norm of the gradient vector exceeds our chosen threshold of 1.0, the gradient is rescaled to have a norm of 1.0 while preserving its direction.
This controlled update mechanism helps stabilize the training process by preventing the ``explosion'' of gradients that could otherwise lead to numerical instability or divergence.
The threshold value of 1.0 represents a conservative choice that effectively limits the magnitude of parameter updates without overly restricting the model's ability to learn.
While gradient clipping was initially proposed for recurrent neural networks, it has proven beneficial for training deep neural networks in general, particularly in scenarios involving complex loss landscapes or when using larger learning rates.

\subsubsection{Dropout layers}

To mitigate overfitting and improve generalization, we employ dropout regularization as introduced in~\cite{srivastava2014dropout}. During training, dropout randomly deactivates neurons with a probability of 0.2, effectively creating an ensemble of different subnetworks.
This technique prevents co-adaptation of neurons by ensuring that no single neuron can rely too heavily on specific features from the previous layer.
The relatively low dropout rate of 0.2 provides a mild form of regularization, helping to reduce overfitting while maintaining most of the network's capacity for learning.
We use the standard \emph{inverted}-dropout convention (as implemented by default in PyTorch and most modern frameworks): during training, the outputs of active neurons are scaled up by a factor of $1/(1-0.2)=1.25$ so that the expected activation magnitude is preserved, and at inference time all neurons are used with no rescaling.
\end{document}